%% file: main.tex
\documentclass{article}

\usepackage[preprint,nonatbib]{neurips_2023}
\usepackage{amsmath}
\usepackage{multirow}
\usepackage{hyperref}
\usepackage{xspace}
\usepackage{amsfonts}
\usepackage{wrapfig}
\usepackage{booktabs}
\usepackage{graphicx}
\usepackage{xcolor}
\usepackage{subfigure}
\usepackage{microtype}

\newcommand{\expo}[1]{\exp\left(#1\right)}
\usepackage{pifont}
\newcommand{\cmark}{\ding{51}}%
\newcommand{\xmark}{\ding{55}}%

\newcommand{\ours}{\texttt{D\"aRF}\xspace}
\newcommand{\norm}[1]{\left\lVert#1\right\rVert}
\newcommand{\mpage}[2]
{
\begin{minipage}{#1\linewidth}\centering
#2
\end{minipage}
}

\title{D\"aRF: Boosting Radiance Fields from Sparse Inputs with Monocular Depth Adaptation}

\author{%
  Jiuhn Song$^{}$\thanks{Equal contribution} ~\quad Seonghoon Park$^{*}$ ~\quad Honggyu An$^{*}$\\
  \enspace \, \textbf{Seokju Cho} \, ~\quad \textbf{Min-Seop Kwak} \, ~\quad \textbf{Sungjin Cho} \, ~\quad \textbf{Seungryong Kim}$^{}$\thanks{Corresponding author} \\\\
  Korea University \\
}

\begin{document}
\maketitle

\input{Writing/0_abstract}
\input{Writing/1_introduction}

\input{Writing/2_relworks}
\input{Writing/3_preliminary}
\input{Writing/4_methodology}
\input{Writing/5_experiments}
\input{Writing/6_conclusion}

\newpage
\bibliographystyle{plain}
\bibliography{egbib}

\newpage
\appendix
\include{Writing/7_suppl}

\end{document}

%% file: Writing/0_abstract.tex
\begin{abstract}
Neural radiance field (NeRF) shows powerful performance in novel view synthesis and 3D geometry reconstruction, but it suffers from critical performance degradation when the number of known viewpoints is drastically reduced. 
Existing works attempt to overcome this problem by employing external priors, but their success is limited to certain types of scenes or datasets. 
Employing monocular depth estimation (MDE) networks, pretrained on large-scale RGB-D datasets, with powerful generalization capability would be a key to solving this problem: however, using MDE in conjunction with NeRF comes with a new set of challenges due to various ambiguity problems exhibited by monocular depths. 
In this light, we propose a novel framework, dubbed \ours, that achieves robust NeRF reconstruction with a handful of real-world images by combining the strengths of NeRF and monocular depth estimation through online complementary training. 
Our framework imposes the MDE network's powerful geometry prior to NeRF representation at both seen and unseen viewpoints to enhance its robustness and coherence. In addition, we overcome the ambiguity problems of monocular depths through patch-wise scale-shift fitting and geometry distillation, which adapts the MDE network to produce depths aligned accurately with NeRF geometry. 
Experiments show our framework achieves state-of-the-art results both quantitatively and qualitatively, demonstrating consistent and reliable performance in both indoor and outdoor real-world datasets. Project page is available at \url{https://ku-cvlab.github.io/DaRF/}. 
\end{abstract}

%% file: Writing/1_introduction.tex
\section{Introduction}
Neural radiance field (NeRF)~\cite{mildenhall2020nerf} has gained significant attention for its powerful performance in reconstructing 3D scenes and synthesizing novel views. However, despite its impressive performance, NeRF often comes with a considerable limitation in that its performance highly relies on the presence of densely well-calibrated input images which are difficult to acquire. As the number of input images is reduced, 
NeRF's novel view synthesis quality drops significantly, displaying failure cases such as erroneous overfitting to the input images~\cite{jain2021putting, Niemeyer2021Regnerf}, artifacts clouding empty spaces~\cite{Niemeyer2021Regnerf}, or degenerate geometry that yields incomprehensible jumble when rendered at unseen viewpoints~\cite{Kim2021Infonerf}. These challenges derive from its under-constrained nature, causing it to have extreme difficulty mapping a pixel in input images to a correct 3D location. In addition, NeRF's volume rendering allows the model to map a pixel to multiple 3D locations~\cite{kangle2021dsnerf}, exacerbating this problem.

Previous \textit{few-shot} NeRF methods attempt to solve these issues by imposing geometric regularization~\cite{Niemeyer2021Regnerf, Kim2021Infonerf, kwak2023geconerf} or exploiting external 3D priors~\cite{kangle2021dsnerf, roessle2021dense} such as depth information extracted from input images by COLMAP~\cite{Schonberger_2016_CVPR}. However, these methods have weaknesses in that they use 3D priors extracted from a few input images only, which prevents such guidance from encompassing the entire scene. To effectively tackle all the issues mentioned above, pretrained monocular depth estimation (MDE) networks with strong generalization capability~\cite{ranftl2020towards, ranftl2021vision,bhat2023zoedepth} could be used to inject an additional 3D prior into NeRF that facilitates robust geometric reconstruction. Specifically, geometry prediction by MDE can constrain NeRF into recovering smooth and coherent geometry, while their bias towards predicting smooth geometry helps to filter out fine-grained artifacts that clutter the scene. More importantly, NeRF's capability to render any unseen viewpoints enables fully exploiting the capability of the MDE, as MDE could provide depth prior to the numerous renderings of unseen viewpoints as well as the original input viewpoints. This allows injecting additional 3D prior to effectively covering the entire scene instead of being constrained to a few input images.

However, applying MDE to few-shot NeRF is not trivial, as there are ambiguity problems that hinder the monocular depth from serving as a good 3D prior. Primarily, relative depths predicted by MDEs are not multiview-consistent~\cite{bhoi2019monocular,choi2022selftune}. Moreover, MDEs perform poorly in estimating depth differences between multiple objects: this prevents global scale-shift fitting~\cite{zhang2022hierarchical,miangoleh2021boosting} from being a viable solution, as alignment to one region of the scene inevitably leads to misalignment in many other regions. There also exists a convexity problem~\cite{miangoleh2021boosting}, in which the MDE has difficulty determining whether the surface is planar, convex, or concave, are also present. To overcome these challenges, we introduce a novel method to adapt MDE to NeRF's absolute scaling and multiview consistency as NeRF is regularized by MDE's powerful 3D priors, creating a complementary cycle.

In this paper, we propose \ours, short for Monocular \textbf{D}epth \textbf{A}daptation for boosting \textbf{R}adiance \textbf{F}ields from Sparse Input Views,  which achieves robust optimization of few-shot NeRF through MDE's geometric prior, as well as MDE adaptation for alignment with NeRF through complementary training (see Fig.~\ref{fig1:motivation}).
We exploit MDE for robust geometry reconstruction and artifact removal in both \textit{unseen} and \textit{seen} viewpoints. In addition, we leverage NeRF to adapt MDE toward multiview-consistent geometry prediction and introduce novel patch-wise scale-shift fitting to more accurately map local depths to NeRF geometry.
Combined with a confidence modeling technique for verifying accurate depth information, our method achieves state-of-the-art performance in few-shot NeRF optimization. We evaluate and compare our approach on real-world indoor and outdoor scene datasets, establishing new state-of-the-art results for the benchmarks.

\input{Figures/tex/fig1}

%% file: Figures/tex/fig1.tex
\begin{figure*}[t]
    \begin{center}
    \includegraphics[width=1\linewidth]{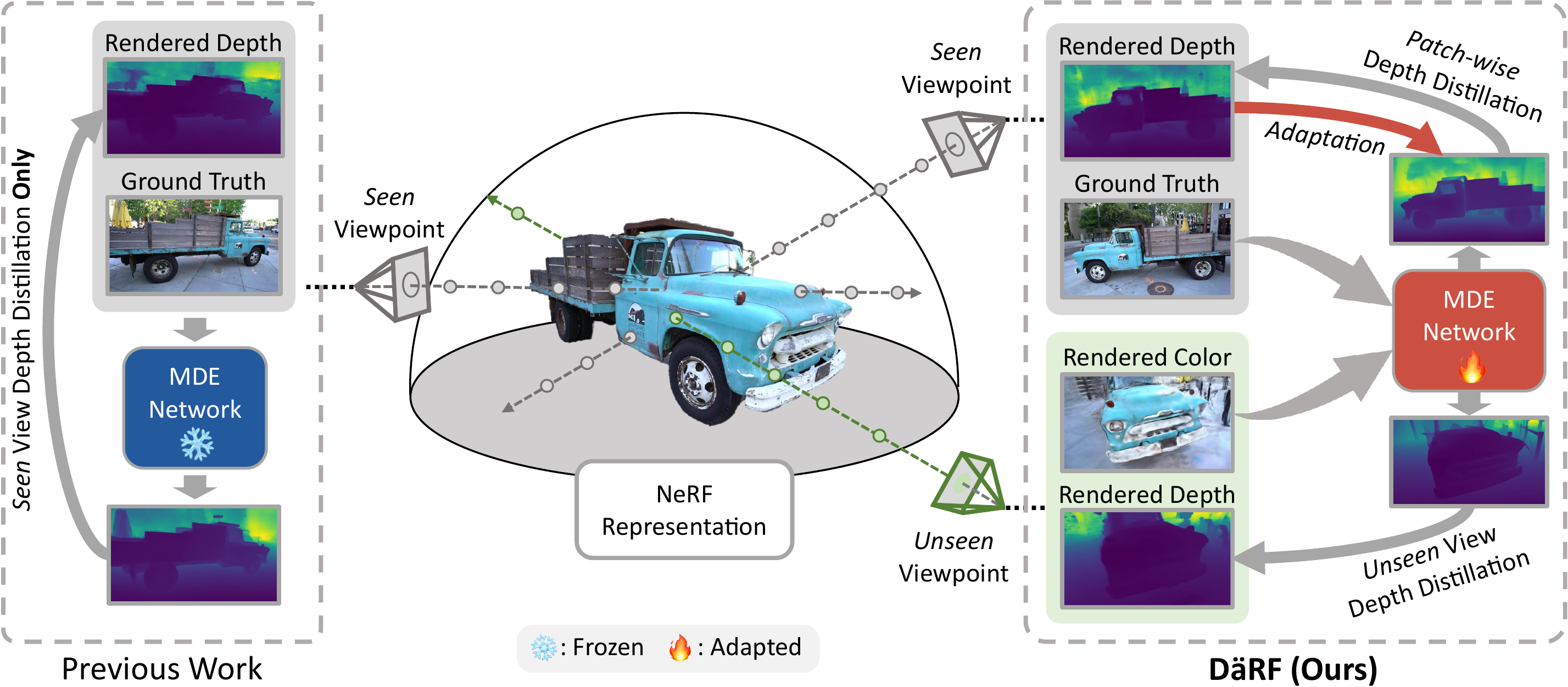}
    \end{center}
    \vspace{-5pt}
    \caption{\textbf{Overview.} \ours shows robust optimization of few-shot NeRF through MDE’s geometric prior, removing inherent ambiguity from MDE through novel patch-wise distillation loss and MDE adaptation.
    Unlike existing work~\cite{uy2023scade} that distills depths by applying pretrained MDE to NeRF at seen view only, our \ours fully exploits the ability of MDE by jointly optimizing NeRF and MDE at a specific scene, and distilling the monocular depth prior to NeRF at both seen and unseen views.}
\label{fig1:motivation}\vspace{-10pt}
\end{figure*}

%% file: Writing/2_relworks.tex
\section{Related Work}
\paragraph{Neural radiance field.} 
Neural radiance field (NeRF)~\cite{mildenhall2020nerf} represents photo-realistic 3D scenes with MLP. Owing to its remarkable performance, there has been a variety of follow-up studies~\cite{barron2021mipnerf,yu2021plenoctrees,liu2020neural}. These studies improve NeRF such as dynamic and deformable scenes~\cite{park2021nerfies, tretschk2021non,pumarola2021d,attal2021torf}, real-time rendering~\cite{yu2021plenoctrees, reiser2021kilonerf, muller2022instant}, unbounded scene~\cite{barron2022mip, tancik2022block, xiangli2022bungeenerf} and generative modeling~\cite{schwarz2020graf, niemeyer2021giraffe, xu2021generative, chan2022efficient}. However, these works still encounter challenges in synthesizing novel views with a limited number of images in a single scene, limiting their applicability in real-world scenarios.
\vspace{-5pt}

\paragraph{Few-shot NeRF.}
Numerous \textit{few-shot} NeRF works attempted to address few-shot 3D reconstruction problem through various techniques, such as pretraining external priors~\cite{yu2021pixelnerf, chibane2021stereo}, meta-learning~\cite{tancik2020meta}, regularization~\cite{jain2021putting, Niemeyer2021Regnerf, Kim2021Infonerf, yang2023freenerf, kwak2023geconerf} or off-the-shelf modules~\cite{jain2021putting, Niemeyer2021Regnerf}. Recent approaches~\cite{Niemeyer2021Regnerf, Kim2021Infonerf, yang2023freenerf, kwak2023geconerf} emphasize the importance of geometric consistency and apply geometric regularization at unknown viewpoints. However, these regularization methods show limitations due to their heavy reliance on geometry information recovered by NeRF. Other works such as DS-NeRF~\cite{kangle2021dsnerf}, DDP-NeRF~\cite{roessle2021dense} and SCADE~\cite{uy2023scade} exploit additional geometric information, such as COLMAP~\cite{Schonberger_2016_CVPR} 3D points or monocular depth estimation, for geometry supervision. However, these works have critical limitations of only being able to provide geometry information corresponding to existing input viewpoints. Unlike these works, our work demonstrates methods to provide geometric prior even at unknown viewpoints with MDE for more effective geometry reconstruction.
\vspace{-5pt}

\paragraph{Monocular depth estimation.}
Monocular depth estimation (MDE) is a task that aims to predict a dense depth map given a single image. Early works on MDE used handcrafted methods such as MRF for depth estimation~\cite{saxena2008make3d}. After the advent of deep learning, learning-based approaches~\cite{eigen2014depth, fu2018deep, kim2016unified, laina2016deeper} were introduced to the field. In this direction, the models were trained on ground-truth depth maps acquired by RGB-D cameras or LiDAR sensors to predict absolute depth values~\cite{li2015depth,lee2019big}. Other approaches trained the networks on large-scale diverse datasets~\cite{chen2017singleimage, lee2019monocular, ranftl2021vision, ranftl2020towards}, which demonstrates better generalization power. These approaches struggle with depth ambiguity caused by ill-posed problem, so the following works LeRes~\cite{yin2021learning} and ZoeDepth~\cite{bhat2023zoedepth} opt to recover absolute depths using additional parameters. 
\vspace{-5pt}

\paragraph{Incorporating MDE into 3D representation.}
As both NeRF and monocular depth estimation are closely related, there have been some works that utilize MDE models to enhance NeRF's performance. NeuralLift~\cite{xu2022neurallift}, MonoSDF~\cite{yu2022monosdf} and SCADE~\cite{uy2023scade} leverage depths predicted by pretrained MDE for depth ordering and detailed surface reconstruction, respectively. Other works optimize scene-specific parameters, such as depth predictor utilizing depth recovered by COLMAP~\cite{wei2021nerfingmvs} or learnable scale-shift values for reconstruction in noisy pose setting~\cite{bian2022nope}. However, these previous approaches were limited in that MDEs were used to provide prior to only the input viewpoints, which constrains their effectiveness when input views are reduced, e.g., in the few-shot setting.

As a concurrent work, SCADE~\cite{uy2023scade} utilizes MDE for sparse view inputs, by injecting uncertainty into MDE through additional pretraining so that canonical geometry can be estimated through probabilistic modeling between multiple modes of estimated depths. While the ultimate goal which is to overcome the ambiguity of MDE may be similar, our approach directly removes ambiguity present in MDE by finetuning with canonical geometry captured by NeRF,for effective suppression of artifacts and divergent behaviors of few-shot NeRF. 

%% file: Writing/3_preliminary.tex
\section{Preliminaries}
NeRF~\cite{mildenhall2020nerf} represents a scene as a continuous function $\mathcal{F}_\theta(\cdot)$ represented by a neural network with parameters $\theta$. During optimization, 3D points are sampled along rays represented by $\mathbf{r}$ coming from a set of input images $\mathcal{S}=\{I_i\}$, 
whose ground truth camera poses are given, for evaluation by the neural network. For each sampled point, $\mathcal{F}_\theta(\cdot)$ takes as input its coordinate $\mathbf{x}\in\mathbb{R}^3$ and viewing direction $\mathbf{d} \in \mathbb{R}^2$ with a positional encoding $\gamma(\cdot)$ that facilitates learning high-frequency details~\cite{tancik2020fourfeat}, and outputs a color $\mathbf{c}\in {\mathbb R^3}$ and a density  $\sigma\in {\mathbb R}$ such that $\{\mathbf{c},\sigma\} = \mathcal{F}_\theta\left(\gamma(\mathbf{x}), \gamma(\mathbf{d})\right)$. With a ray parameterized as $\mathbf{r}_\mathbf{p}(t) = \mathbf{o} + t\mathbf{d}_\mathbf{p}$, starting from camera center $\mathbf{o}$ along the direction $\mathbf{d}_\mathbf{p}$, color and depth value at the pixel $\mathbf{p}$ are rendered as follows:
\begin{equation} 
    \bar{I}(\mathbf{p}) =  \int_{t_n}^{t_f} T(t)\mathbf{\sigma}(\mathbf{r}_\mathbf{p}(t))\mathbf{c}(\mathbf{r}_\mathbf{p}(t))dt, \enspace \bar{D}(\mathbf{p}) = \int_{t_n}^{t_f} T(t)\sigma(\mathbf{r}_\mathbf{p}(t))t dt,
\end{equation}
where $\bar{I}(\mathbf{p})$ and $\bar{D}(\mathbf{p})$ are rendered color and depth values at the pixel $\mathbf{p}$ along the ray $\mathbf{r}_\mathbf{p}(t)$ from $t_n$ to $t_f$, and $T(t)$ denotes an accumulated transmittance along the ray from $\mathit{t_n}$ to $\mathit{t}$ as follows:
\begin{equation}
    \quad T(t) = \expo{-\int_{t_n}^{t}\sigma(\mathbf{r}_\mathbf{p}(s))ds}.
\end{equation}

Based on this volume rendering, $\mathcal{F}_\theta(\cdot)$ is optimized by the reconstruction loss $\mathcal{L}_\mathrm{recon}$ that compares rendered color $\bar{I}(\mathbf{p})$ to corresponding ground-truth $I(\mathbf{p})$, with $\mathcal{R}$ as a set of pixels for training rays: 
\begin{equation}   
\mathcal{L}_\mathrm{recon}=
\sum_{I_i\in\mathcal{S}} 
\sum_{\mathbf{p}\in \mathcal{R}}
\lVert I_i(\mathbf{p})-\bar{I}_i(\mathbf{p}) \rVert_2^2.
\end{equation}
Our work explores the setting of few-shot optimization with NeRF~\cite{Kim2021Infonerf, kwak2023geconerf}. Whereas the number of input viewpoints $|\mathcal{S}|$ is normally higher than one hundred in the standard NeRF setting~\cite{mildenhall2020nerf}, the task of few-shot NeRF considers scenarios when $|\mathcal{S}|$ is drastically reduced to a few viewpoints (e.g., $|\mathcal{S}|<20$). With such a small number of input viewpoints, NeRF shows high divergent behaviors such as geometry breakdown, overfitting to input viewpoints, and generation of artifacts that cloud the empty space between the camera and object, which causes its performance to drop sharply~\cite{jain2021putting,Kim2021Infonerf,Niemeyer2021Regnerf}. To overcome this problem, existing few-shot NeRF frameworks applied regularization techniques at unknown viewpoints to constrain NeRF with additional 3D priors~\cite{roessle2021dense, kangle2021dsnerf} and enhance the robustness of geometry, but they showed limited performance.

%% file: Writing/4_methodology.tex
\section{Methodology}
\input{Figures/tex/fig2}
\subsection{Motivation and Overview}
Our framework leverages the complementary benefits of few-shot NeRF and monocular depth estimation networks for the goal of robust 3D reconstruction. The benefits that pretrained MDE can provide to few-shot NeRF are clear and straightforward: because they predict dense geometry, they provide guidance for the NeRF to recover more smooth geometry. In cases where few-shot NeRF's geometry undergoes divergent behaviors, MDE provides strong constraints to prevent the global geometry from breaking down. 

However, there are difficult challenges that must be overcome if the depths estimated by MDE are to be used as 3D prior to NeRF. These challenges, which can be summarized as depth ambiguity problems~\cite{miangoleh2021boosting}, stem from the inherent ill-posed nature of the monocular depth estimation. Most importantly, MDE networks only predict relative depth information inferred from an image, meaning it is initially not aligned to NeRF's absolute geometry~\cite{bhat2023zoedepth}. Global scaling and shifting may seem to be the answer, but this approach leads us to another depth ambiguity problem, as predicted scales and spacings of each instance are inconsistent with one another, as demonstrated in (a) of Fig.~\ref{fig2:patch}. Additionally, MDE's weakness in predicting the convexity of a surface, whether it is flat, convex, or concave - also poses a problem in using this depth for NeRF guidance. 

In this light, we adapt a pre-trained monodepth network to a single NeRF scene so that its powerful 3D prior can be leveraged to its maximum capability in regularizing the few-shot NeRF. In the following, we first explain how to distill geometric prior from off-the-shelf MDE model~\cite{ranftl2020towards} from both seen and unseen viewpoints (Sec.~\ref{method:distillation}). We also provide a strategy for adapting the MDE model to handle ill-posed problems to a specific scene, while keeping its 3D prior knowledge (Sec.~\ref{method:adaptation}). Then, we demonstrate a method to handle inaccurate depths (Sec.~\ref{method:confidence}). Fig.~\ref{fig1:motivation} shows an overview of our method, compared to previous works using MDE prior~\cite{uy2023scade,yu2022monosdf}.

\subsection{Distilling Monocular Depth Prior into Neural Radiance Field} \label{method:distillation}
To prevent the degradation of reconstruction quality in few-shot NeRF, we propose to distill monocular depth prior to the neural radiance field during optimization. By exploiting pre-trained MDE networks~\cite{ranftl2021vision, ranftl2020towards}, which have high generalization power, we enforce a dense geometric constraint on both \textit{seen} and \textit{unseen} viewpoints by using estimated monocular depth maps as pseudo ground truth depth for training few-shot NeRF. We describe the details of this process below.
\vspace{-5pt}

\input{Figures/tex/fig3}
\paragraph{Monocular depth regularization on seen views.}
We leverage a pre-trained MDE model, denoted as $\mathcal{G}_\phi(\cdot)$ with parameters $\phi$, to predict pseudo depth map from given \textit{seen} view image $I_{i}$ as  ${D}^*_i = \mathcal{G}_\phi(I_i)$. Since ${D}^*_i$ is initially a relative depth map, it needs to be scaled and shifted into an absolute depth~\cite{zhang2022hierarchical} and aligned with NeRF's rendered depth $\bar{D}$ in order for it to be used as pseudo-depth $D^*$. However, the scale and shift parameters inferred from the global statistic may undermine local statistic~\cite{zhang2022hierarchical}. For example, as shown in Fig.~\ref{fig2:patch}~(a), global scale fitting tends to favor dominant objects in the image, leading to ill-fitted depths in less dominant sections of the scene due to inconsistencies in predicted depth differences between the objects. Na\"ively employing such inaccurately estimated depths for distillation can adversely impact the overall geometry of the NeRF. 

To alleviate this issue, we propose a patch-wise adjustment of scale and shift parameters, reducing the impact of erroneous depth differences, as illustrated in Fig.~\ref{fig2:patch}~(b). The depth consistency loss is defined as follows:
\begin{equation}
\mathcal{L}_\text{seen} = \sum_{I_i\in\mathcal{S}} \sum_{\mathbf{p} \in \mathcal{P}} {\|  (w_i \texttt{sg}\left(D^*_{i}(\mathbf{p})\right) + q_i) - \bar{D}_i(\mathbf{p}) \|}, \label{eq:depth_prior}
\end{equation}
where $w_i$ and $q_i$ denote the scale and shift parameters obtained by least square~\cite{ranftl2020towards} between $D^*_i$ and $\bar{D}_{i}$, $\mathcal{P}$ denotes a set of pixels within a patch, and $\texttt{sg}(\cdot)$ denotes stop-gradient operation~\cite{chen2021exploring}. Thus patch-based approach also helps to overcome the computational bottleneck of full image rendering. 
\vspace{-5pt}

\paragraph{Monocular depth regularization on unseen views.}
We further propose to give supervision even at \textit{unseen} viewpoints. As NeRF has the ability to render any unseen viewpoint of the scene, we render color $\bar{I}_l$ and depth $\bar{D}_l$ from a sampled patch of $l$-th novel viewpoint. Sequentially, we extract a monocular depth map from the rendered image as $\bar{D}^*_l = \mathcal{G}_\phi(\bar{I}_l)$. Then, we enforce consistency between our rendered depth $\bar{D}_l$ and the monocular depth $\bar{D}^*_l$ of $l$-th novel viewpoint as follows: 
\begin{equation}
\mathcal{L}_\text{unseen} =  \sum_{I_l\in\mathcal{U}} \sum_{\mathbf{p} \in \mathcal{P}} {\| (w_l\texttt{sg}\left(\bar{D}^*_{l}(\mathbf{p})\right) +q_l) - \bar{D}_l(\mathbf{p}) \|}, \label{eq:nvl}
\end{equation}
where $\mathcal{U}$ denotes a set of unseen view images, $w_l$ and $q_l$ denotes the scale and shift parameters used to align $\bar{D}^*_{l}$ towards $\bar{D}_{l}$, and $\mathcal{P}$ denotes randomly sampled patch. 

A valid concern regarding this approach is that monocular depth obtained from noisy NeRF rendering may be affected by fine-grained rendering artifacts that frequently appear in unseen viewpoints of few-shot NeRF, resulting in noisy and erroneous pseudo-depths. However, we demonstrate in Fig.~\ref{fig3:MDE} that a strong geometric prior within the MDE model exhibits robustness against such artifacts, effectively filtering out the artifacts and thereby providing reliable supervision for the unseen views.

It should be noted that our strategy differs from previous methods~\cite{kangle2021dsnerf, roessle2021dense, yu2022monosdf, uy2023scade} that exploit monocular depth estimation~\cite{ranftl2021vision} and external depth priors such as COLMAP~\cite{Schonberger_2016_CVPR}. These methods only impose depth priors upon the input viewpoints, and thus their priors only influence the scene partially due to self-occlusions and sparsity of known views. Our method, on the other hand, enables external depth priors to be applied to any arbitrary viewpoint and thus allows guidance signals to thoroughly reach every location of the scene, leading to more robust and coherent NeRF optimization.

\subsection{Adaptation of MDE via Neural Radiance Field} \label{method:adaptation}
Although the patch-wise distillation of monocular depth provides invariance to depth difference inconsistency in MDE, the ill-posed nature of monocular depth estimation often introduces additional ambiguities, such as the inability to distinguish whether the surface is concavity, convexity, or planar or difficulty in determining the orientation of flat surfaces~\cite{miangoleh2021boosting}. We argue that these ambiguities arise due to the MDE lacking awareness of the scene-specific absolute depth priors and multiview consistency. To address this issue, we propose providing the scene priors optimized NeRF to MDE, whose knowledge of canonical space and absolute geometry helps eliminate the ambiguities present within MDE. Therefore, we propose to adapt the MDE to the absolute scene geometry, formally written as:
\begin{equation}
\mathcal{L}_\text{MDE} =  \sum_{I_i\in\mathcal{S}} \sum_{\mathbf{p} \in \mathcal{P}} 
\left\{
{\| \texttt{sg}\left(\bar{D}_i(\mathbf{p})\right) - D^*_i(\mathbf{p}) \|} + {\| (w_i\texttt{sg}\left(\bar{D}_{i}(\mathbf{p})\right)+q_i) - D^*_i(\mathbf{p}) \|} \right\}. \label{eq:depth_adaptation}
\end{equation}
In addition to the patch-wise loss in Eq.~\ref{eq:depth_prior}, we add an $l$-1 loss without scale-shift adjustment to adapt the MDE with absolute depth prior. We also introduce a regularization term to preserve the local smoothness of MDE, given by:
\begin{equation}
\mathcal{L}_\text{reg} = \sum_{I_i\in\mathcal{S}}\sum_{\mathbf{p} \in \mathcal{P}} {\| (w_{i}\texttt{sg}\left({D}^{*,\mathrm{init}}_{i}(\mathbf{p})\right) +q_{i}) - {D}^*_{i}(\mathbf{p}) \|},
\end{equation}
where ${D}^{*,\mathrm{init}}_{i}$ denotes monocular depth map of $I_i$ extracted from MDE with initial pre-trained weight.

\subsection{Confidence Modeling} \label{method:confidence}

Our framework must take into account the errors present in both few-shot NeRF and estimated monocular depths, which will propagate~\cite{sohn2020fixmatch} and intensify during the distillation process if left unchecked. To prevent this, we adopt confidence modeling~\cite{kwak2023geconerf, sohn2020fixmatch} inspired by self-training approaches~\cite{sohn2020fixmatch, arazo2020pseudo}, to verify the accuracy and reliability of each ray before the distillation process.

The homogeneous coordinates  of a pixel $\mathbf{p}$ in the seen viewpoint are transformed to $\mathbf{p}'$ at the target viewpoint using the viewpoint difference $R_{i\rightarrow l}$ and the camera intrinsic parameter $K$, as follows:
\begin{equation}
\mathbf{p}' \sim KR_{i\rightarrow l}{D}_i(\mathbf{p})K^{-1}\mathbf{p}.
\end{equation}
We generate the confidence map $M_i$ by measuring the distance between rendered depth of the unseen viewpoint and MDE output of seen viewpoint such that
\begin{equation}
    M_i(\mathbf{p}) = \big[\|(w_i{D}^*_i(\mathbf{p}) + q_i) - \bar{D}_l(\mathbf{p}') \| < \tau\big],
\end{equation}
where $\tau$ denotes threshold parameter, $[\cdot]$ is Iverson bracket, and ${D}_l(\mathbf{p}')$ refers to depth value of the corresponding pixel at $l$-th unseen viewpoint for reprojected target pixel $\mathbf{p}$ of $i$-th seen viewpoint. We fit $D^*_i$ to absolute scale, where scale and shift parameters, $w_i$ and $q_i$, are obtained by least square~\cite{ranftl2020towards} between $D^*_i$ and $\bar{D}_{i}$.

\subsection{Overall Training}

With the incorporation of confidence modeling, the loss functions for both the radiance field and MDE can redefined.
$\mathcal{L}_\mathrm{seen}$ and $\mathcal{L}_\mathrm{unseen}$ can be redefined as:
\begin{align}
    &\mathcal{L}_\text{seen} = \sum_{I_i\in\mathcal{S}} \sum_{\mathbf{p} \in \mathcal{P}} M_i(\mathbf{p}) \norm{ (w_i \texttt{sg}\left(D^*_{i}(\mathbf{p})\right) + q_i)-\bar{D}_i(\mathbf{p})}, \\
    &\mathcal{L}_\text{unseen} = \sum_{I_l\in\mathcal{U}} \sum_{\mathbf{p} \in \mathcal{P}} M_l(\mathbf{p})\norm{(w_l\texttt{sg}\left(\bar{D}^*_{l}(\mathbf{p})\right) +q_l)-\bar{D}_l(\mathbf{p})}.
\end{align}
In addition, the loss for the adaptation of the MDE module can be redefined considering $M$:
\begin{equation}
\mathcal{L}_\text{MDE} =  \sum_{I_i\in\mathcal{S}} \sum_{\mathbf{p} \in \mathcal{P}} 
M_i(\mathbf{p})\left(
{ {\lVert \texttt{sg}\left(\bar{D}_{i}(\mathbf{p})\right)-\bar{D}^*_{i}(\mathbf{p}) \rVert} + {\| (w_i\texttt{sg}\left(\bar{D}_{i}(\mathbf{p})\right) +q_i)-\bar{D}^*_{i}(\mathbf{p}) \|}}
\right).
\end{equation}

With these losses, we train both NeRF and MDE simultaneously, enhancing both models by complementing each other. MDE provides a strong geometric prior to NeRF while having the inherent limitation of obliviousness to the scene-specific prior, whereas NeRF provides it with its absolute geometry.

%% file: Figures/tex/fig2.tex
\begin{figure}[t]
\centering 
    \renewcommand{\thesubfigure}{}
     \subfigure[(a)]{
    \includegraphics[width=0.32\linewidth]{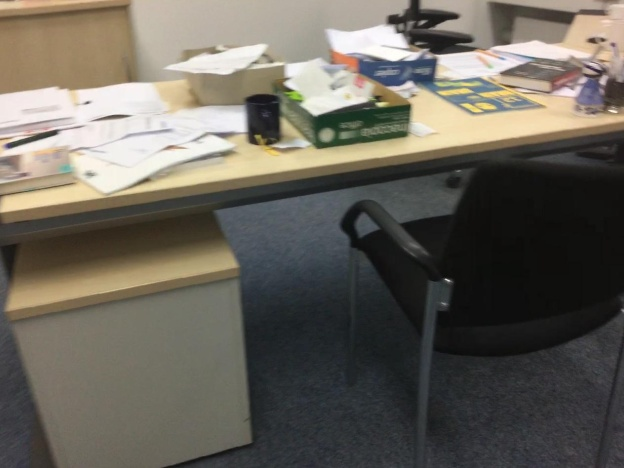}}
     \subfigure[(b)]{
    \includegraphics[width=0.32\linewidth]{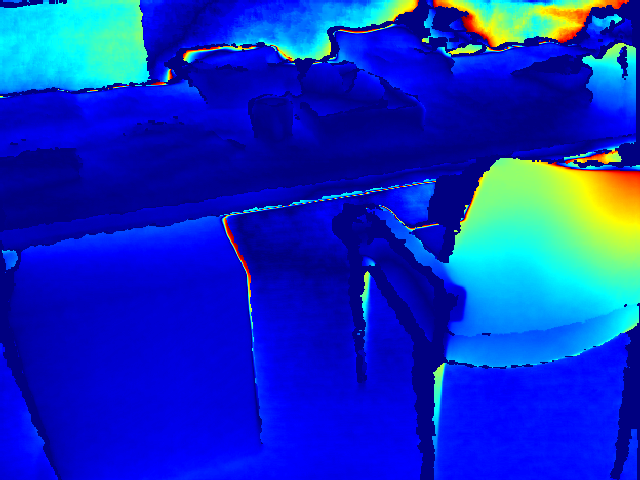}}
     \subfigure[(c)]{
    \includegraphics[width=0.32\linewidth]{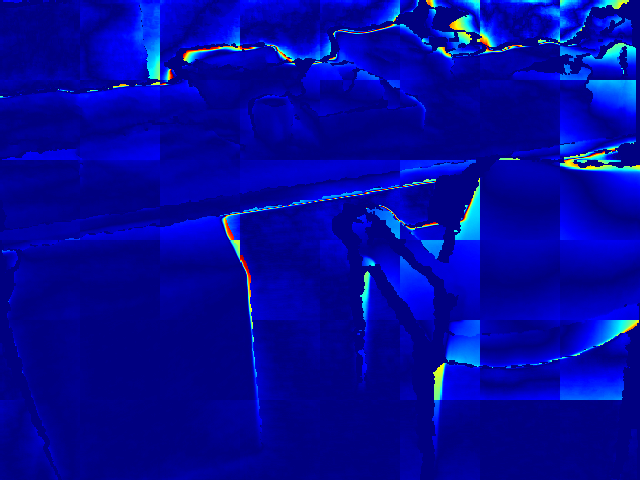}}\hfill\\
    \vspace{-5pt}
    \caption{\textbf{Effectiveness of patch-wise scale and shift adjustment:} (a) input image, (b) monocular depth with image-level adjustment, and (c) monocular depth with patch-level adjustment. We visualize the error of adjusted monocular depth from input image compared to GT depth value. The proposed patch-level adjustment helps to minimize the errors caused by inconsistency in depth differences among objects.}
    \label{fig2:patch}\vspace{-10pt}
\end{figure}

%% file: Figures/tex/fig3.tex
\begin{figure}[t]
  \centering
  \renewcommand{\thesubfigure}{}
     \subfigure[]
    {\includegraphics[width=0.246\linewidth]{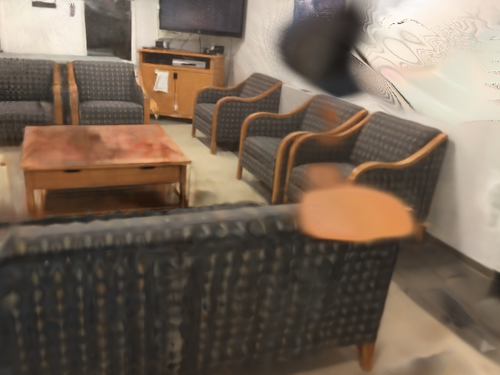}}\hfill
     \subfigure[]
    {\includegraphics[width=0.246\linewidth]{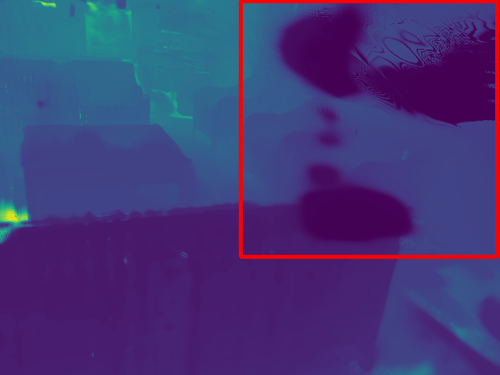}}\hfill
     \subfigure[]
    {\includegraphics[width=0.246\linewidth]{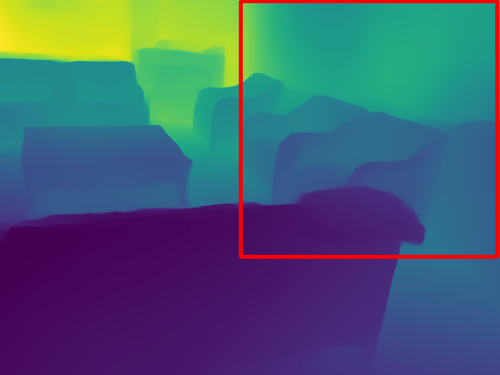}}\hfill
     \subfigure[]
    {\includegraphics[width=0.246\linewidth]{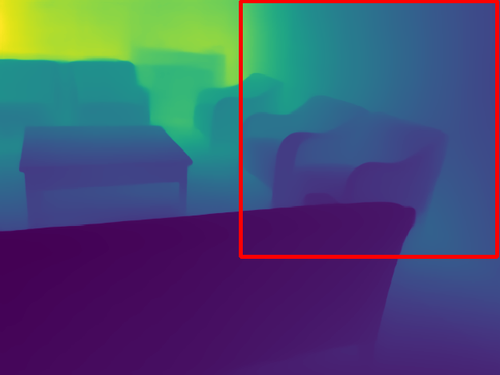}}\hfill\\
    \vspace{-20pt}
    
     \subfigure[(a) Rendered color $\bar{I}$]
    {\includegraphics[width=0.246\linewidth]{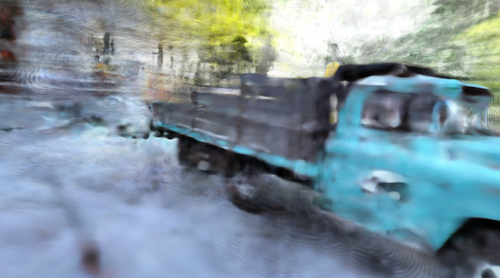}}\hfill
     \subfigure[(b) Rendered depth $\bar{D}$]
    {\includegraphics[width=0.246\linewidth]{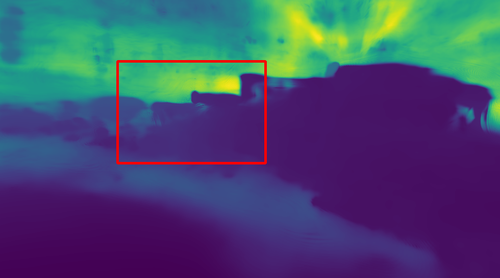}}\hfill
     \subfigure[(c) Mono-depth from $\bar{I}$]
    {\includegraphics[width=0.246\linewidth]{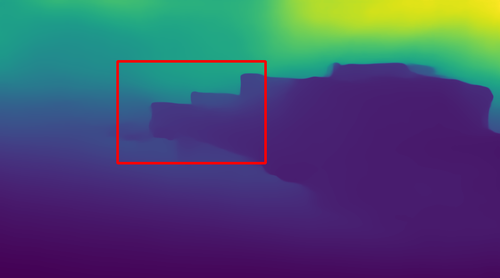}}\hfill
     \subfigure[(d) Mono-depth from $I$]
    {\includegraphics[width=0.246\linewidth]{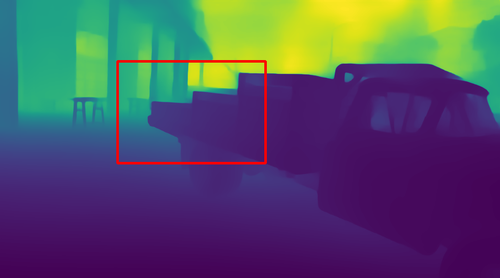}}\hfill\\
    
\vspace{-5pt}
    \caption{\textbf{Robustness of MDE model for multi-view scale ambiguity and artifacts:} (a-b) color and depth of NeRF rendered in the early stage of the training, (c-d) monocular depths estimated from rendered image $\bar{I}$ and input image $I$. The results show that MDE model ignores the artifacts of rendered images by NeRF, enabling reliable supervision for seen and unseen viewpoint.}
    \label{fig3:MDE}\vspace{-10pt}
\end{figure}

%% file: Writing/5_experiments.tex
\begin{table*}[t]
\begin{center}
    \caption{\textbf{Quantitative comparison on ScanNet~\cite{Dai_2017_CVPR} and Tanks and Temples~\cite{knapitsch2017tanks}.} The best results are highlighted in bold, while the second best results are underlined.}\label{tab:quan}
    % \vspace{-5pt}
    \resizebox{0.99\linewidth}{!}{
    \begin{tabular}{l|c|ccc|ccc|ccc}
    \toprule
       \multirow{3}{*}{Methods} &  \multirow{3}{*}{Depth prior} &  \multicolumn{6}{c|}{ScanNet~\cite{Dai_2017_CVPR}}  & \multicolumn{3}{c}{Tanks and Temples~\cite{knapitsch2017tanks}}\\
       \cline{3-11}
        &  &\multicolumn{3}{c|}{9 - 10 views} &\multicolumn{3}{c|}{18 - 20 views} & \multicolumn{3}{c}{10 views}\\
        & & PSNR $\uparrow$ & SSIM $\uparrow$ & LPIPS $\downarrow$ & PSNR $\uparrow$ & SSIM $\uparrow$ & LPIPS $\downarrow$  & PSNR $\uparrow$ & SSIM $\uparrow$ & LPIPS $\downarrow$  \\ \midrule\midrule %
        NerfingMVS~\cite{wei2021nerfingmvs} & \cmark  & N/A & N/A & N/A & 16.29 & 0.626 & 0.502  & N/A & N/A & N/A  \\ %
        $K$-planes~\cite{fridovich2023k} & \xmark  & 16.01 & 0.618 & 0.494 & 18.70 & 0.708 & 0.400 & 12.57 & 0.453 & 0.607 \\  \midrule 
        RegNeRF~\cite{Niemeyer2021Regnerf} & \xmark   & \underline{16.38} & \underline{0.624} & \underline{0.493} & 18.93 & 0.676 & 0.450 & \underline{14.12} & \underline{0.469} & \textbf{0.580}\\ %
        DS-NeRF~\cite{kangle2021dsnerf} & \cmark & N/A & N/A & N/A & 20.85 & 0.713 & 0.344 & N/A & N/A & N/A \\ %
        DDP-NeRF~\cite{roessle2021dense} & \cmark  & N/A & N/A & N/A & 19.29 & 0.695 & 0.368 & N/A & N/A & N/A \\  %
        SCADE~\cite{uy2023scade} & \cmark & - & - & - & \underline{21.54} & \underline{0.732} & \textbf{0.292} & - & - & - \\ \midrule 
        \ours (Ours) & \cmark  & \textbf{18.29} & \textbf{0.690} & \textbf{0.412} & \textbf{21.58} &  \textbf{0.765} & \underline{0.325} & \textbf{15.70} & \textbf{0.514} & \underline{0.583} \\
        \bottomrule
    \end{tabular}}
    \label{quan:main}
\vspace{-10pt} 
\end{center}
\end{table*}
\begin{table}[t]
	\centering
    \vspace{-2pt}
    \caption{%
    \textbf{Evaluation of depth quality:}
    (a) quantitative evaluation of the adapted MDE, compared with other monocular depth estimation models and (b) visualization of depth distributions. The adapted MDE by our method shows a similar distribution to that of the ground truth. }
	\setlength{\tabcolsep}{2pt}
    \begin{tabular}{cc}
    	\resizebox{0.55\textwidth}{!}{
    	    \begin{tabular}{l|cccc}
    \toprule
       Methods & AbsRel $\downarrow$ & SqRel $\downarrow$ & RMSE $\downarrow$ & RMSE log $\downarrow$ \\ \midrule\midrule 
        LeRes~\cite{yin2021learning}  & 0.391  & 0.472 	& 0.999 & 0.661     \\ 
        MiDaS~\cite{ranftl2020towards} & \underline  {0.152}  & 0.095  & 0.452  & 0.183   \\ 
        DPT~\cite{ranftl2021vision} & 0.191  & 0.135  & 0.563  & 0.220     \\ \midrule 
        \ours (9 - 10 views) & 0.154 & \underline{0.074}  & \underline    {0.361}   & \underline{0.171}   \\
        \ours (18 - 20 views) & \textbf{0.151} & \textbf{0.071}  & \textbf{0.356}   & \textbf{0.168}  \\
        \bottomrule
    \end{tabular}
    	}
    	&
    	\begin{minipage}{0.44\textwidth}
    		\includegraphics[width=0.9\textwidth]{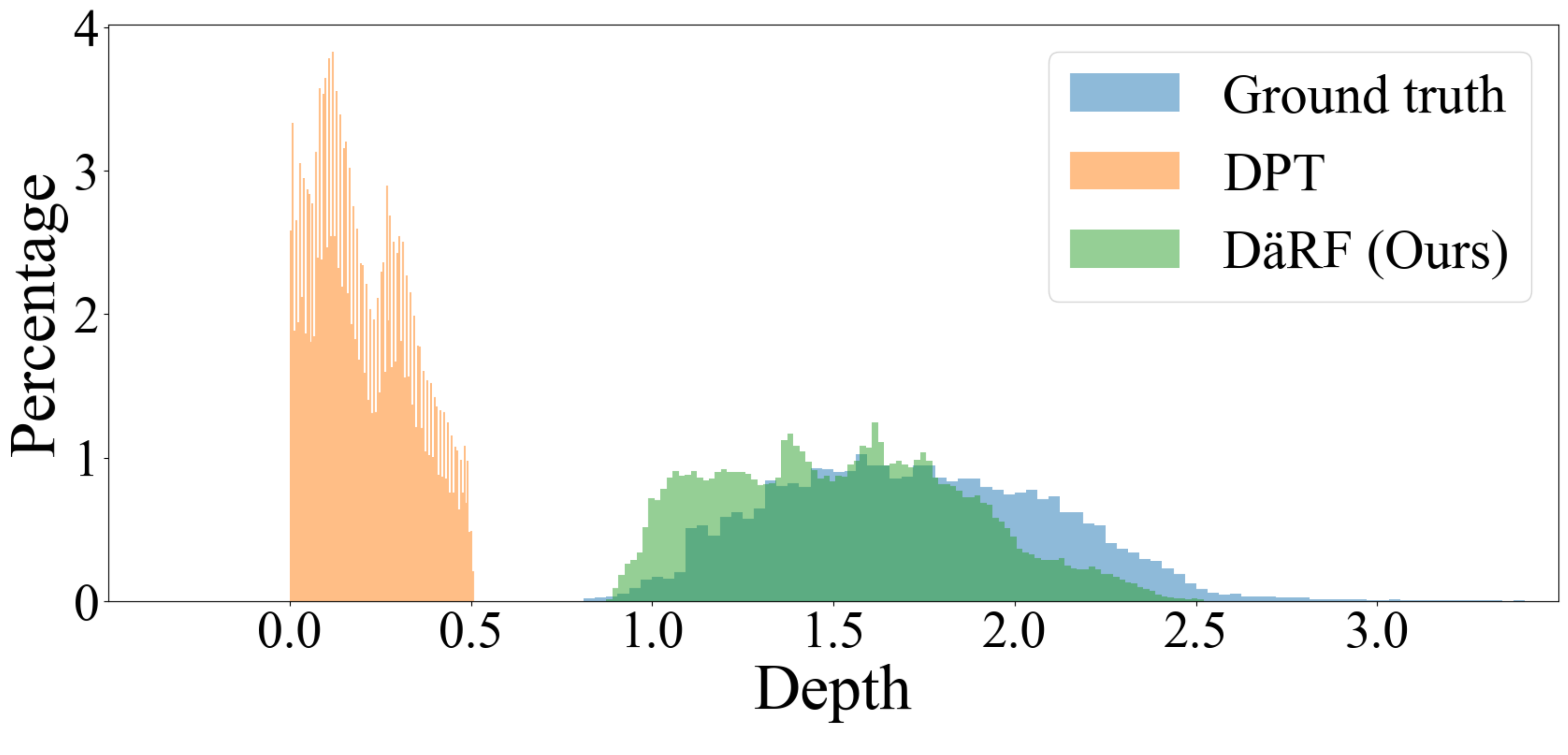}
    	\end{minipage}\\
    (a) \small Quantitative comparison & \small (b) Depth distribution comparison \\
    \end{tabular}
    \label{tab:depth}
    \vspace{-15pt}
\end{table}
\section{Experiments}
\subsection{Experimental Settings}
\paragraph{Implementation details.} 
\ours is implemented based on $K$-planes~\cite{paszke2019pytorch} as NeRF. We use DPT-hybrid~\cite{ranftl2021vision} as MDE model. We use Adam~\cite{kingma2015adam} as an optimizer, with a learning rate of $1\cdot10^{-2}$ for NeRF and $1\cdot10^{-5}$ for the MDE, along with a cosine warmup learning rate scheduling. See supplementary material for more details. The code and pre-trained weights will be made publicly available. \vspace{-5pt}
\paragraph{Datasets.}
We evaluate our method in real-world scenes captured at both indoor and outdoor locations. Following previous works~\cite{roessle2021dense, uy2023scade}, we use a subset of sparse-view ScanNet data~\cite{Dai_2017_CVPR} comprised with three indoor scenes, each consisting of 18 to 20 training images and 8 test images. We also conduct evaluations on more challenging setting with 9 to 10 train images. 
For outdoor reconstruction, we further test on 5 challenging scenes from the Tanks and Temples dataset~\cite{knapitsch2017tanks}. The scenes are real-world outdoor dataset, with a wide variety of scene scales and lighting conditions. Note that these setups are extremely sparse compared to full image setups, where we use approximately 0.5 to 5 percent of the whole training inputs.\vspace{-5pt}

\paragraph{Baselines.}
We adopt the following six recently proposed methods as baselines: 
standard neural radiance field method: $K$-planes~\cite{fridovich2023k},
few-shot NeRF method: RegNeRF~\cite{Niemeyer2021Regnerf},
 and depth prior based methods: NerfingMVS~\cite{wei2021nerfingmvs}, DS-NeRF~\cite{kangle2021dsnerf}, DDP-NeRF~\cite{roessle2021dense}, and SCADE~\cite{uy2023scade}. For methods whose code has not open-sourced, we leave the result as blank. \vspace{-5pt}

\paragraph{Evaluation metrics.}
For quantitative comparison, we follow the NeRF~\cite{mildenhall2020nerf} and report the PSNR, SSIM~\cite{ssim}, LPIPS~\cite{lpips}. We report standard evaluation metrics for depth estimation~\cite{eigen2014depth}, absolute relative error (Abs Rel), squared relative error (SqRel), root mean squared error (RMSE), root mean squared log error (RMSE log). To evaluate view consistency, we utilize a single scaling factor $s$ for each scene, which is the median scaling~\cite{zhang2021consistent} value averaged across all test views.

\begin{wrapfigure}{r}{0.36\textwidth}
\vspace{-15pt}
\centering{
\renewcommand{\thesubfigure}{}
\subfigure[]
{\includegraphics[width=0.49\linewidth]{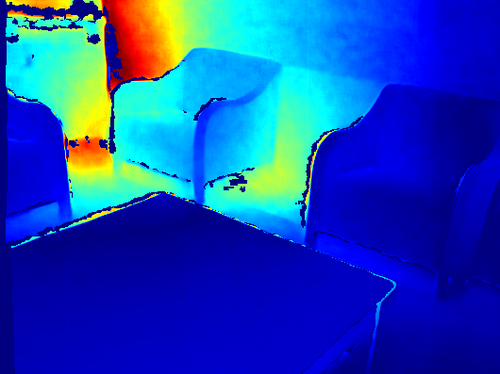}}\hfill
 \subfigure[]
{\includegraphics[width=0.49\linewidth]{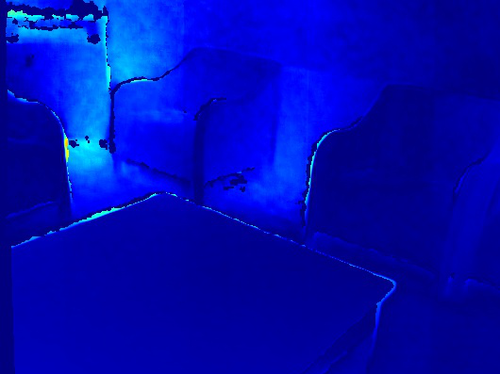}}\hfill\\
\vspace{-20pt}
    
 \subfigure[(a) Frozen MDE]
{\includegraphics[width=0.49\linewidth]{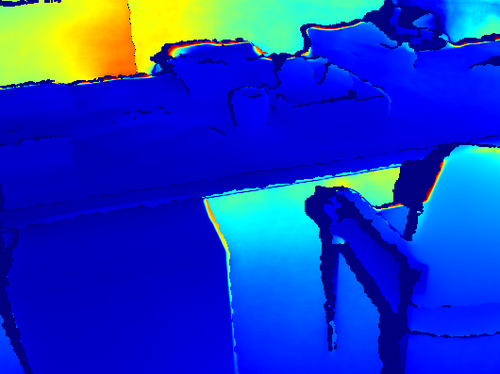}}\hfill
 \subfigure[(b) Adapted MDE]
{\includegraphics[width=0.49\linewidth]{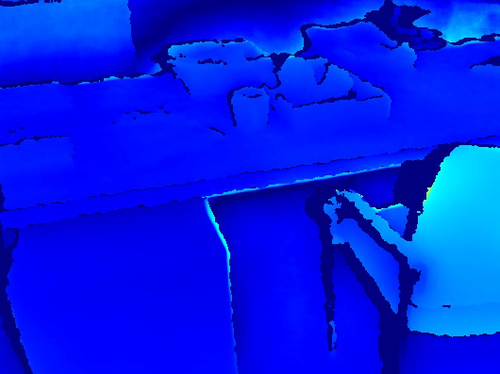}}\\\hfill
\vspace{-10pt}
\caption{\textbf{Error map visualization.}  MDE adaptation results in a reduction of errors.}
\vspace{-10pt}
\label{qual:depth_error}
}
\end{wrapfigure}

\input{Figures/tex/fig5}

\input{Figures/tex/fig6}
\subsection{Comparisons}
\paragraph{Indoor scene reconstruction.}
We conducted experiments in two settings: (1) a standard few-shot setup as described in literature~\cite{roessle2021dense, uy2023scade}, and (2) an extreme few-shot setup with approximately 0.5 percent of the full images. As shown in Tab.~\ref{tab:quan}, our approach outperforms the baseline methods in both settings in most of the metrics. Additionally, we provide quantitative results of the adapted MDE model in ScanNet dataset in Tab.~\ref{tab:depth}, and qualitative results in Fig.~\ref{qual:depth_error}. 
As shown in Fig.~\ref{qual:scan20} for the setting of standard few-shot, DS-NeRF~\cite{kangle2021dsnerf} and DDP-NeRF~\cite{roessle2021dense} still show floating artifacts in the novel view and show limitation in capturing details in the chair, smoothing into nearby object. Our method shows better qualitative results compared to other baselines, showing better geometry understanding and detailed view synthesis in the small objects near the chair.
In the extreme few-shot setup, we conducted a visual comparison between our method and a baseline~\cite{fridovich2023k} in Fig.~\ref{qual:scan10}. This is a more complex setting than standard, but our method outperforms the baseline, showing better geometric understanding. More qualitative images are included in the supplementary material. \vspace{-5pt}

\begin{figure}[t]
\begin{center}
    \renewcommand{\thesubfigure}{}
    \subfigure[]
    {\includegraphics[width=0.245\textwidth]{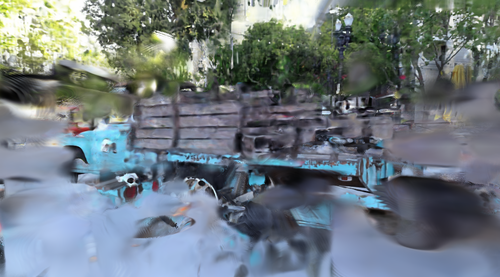}}
    \subfigure[]
    {\includegraphics[width=0.245\textwidth]{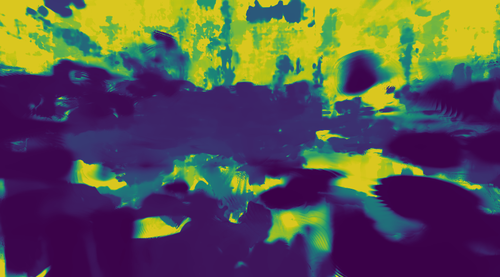}}
    \subfigure[]
    {\includegraphics[width=0.245\textwidth]{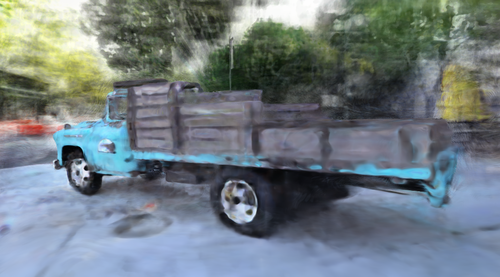}}
    \subfigure[]
    {\includegraphics[width=0.245\textwidth]{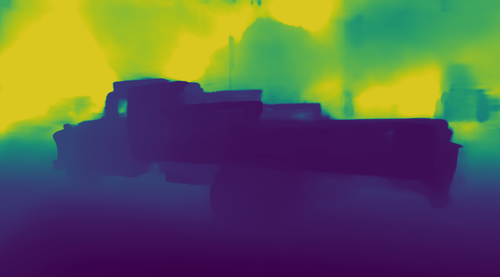}}
    \hfill\\
    \vspace{-20.5pt}
    \subfigure[(a) Baseline~\cite{fridovich2023k}]
    {\includegraphics[width=0.245\textwidth]{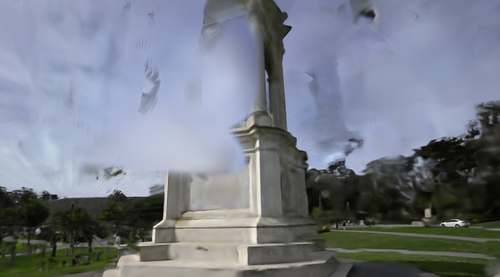}}
    \subfigure[(b) Baseline- Depth]
    {\includegraphics[width=0.245\textwidth]{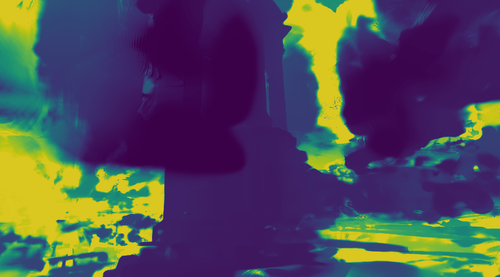}}
    \subfigure[(c) \ours]
    {\includegraphics[width=0.245\textwidth]{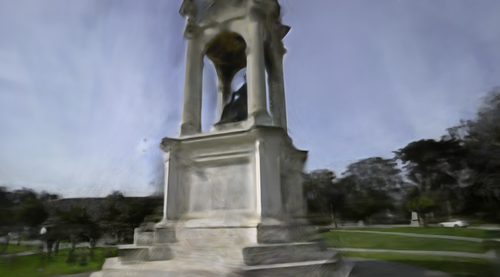}}
    \subfigure[(d) \ours- Depth]
    {\includegraphics[width=0.245\textwidth]{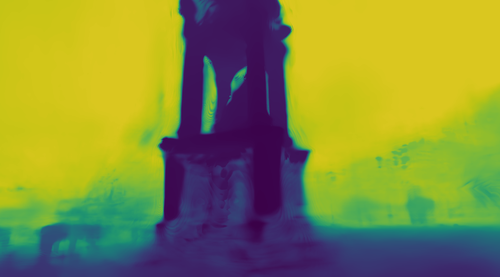}}\\
    \vspace{-5pt}
    \caption{\textbf{Qualitative results on Tanks and Temples~\cite{knapitsch2017tanks}}.}
    \label{qual:tnt}
    \vspace{-15pt}
\end{center}
\end{figure}

\paragraph{Outdoor scene reconstruction.}
We conduct the qualitative and quantitative comparisons on the Tanks and Temples dataset in Tab.~\ref{tab:quan} and Fig.~\ref{qual:tnt}. Since COLMAP~\cite{Schonberger_2016_CVPR} with sparse images is not available, we provide comparisons with baselines without explicit depth prior. The quantitative results show that our approach outperforms the baseline methods on this complex outdoor dataset in all metrics. As shown in Fig.~\ref{qual:tnt}, our baseline shows limited performance, despite its feasible results of view synthesis in novel viewpoint, its depth results show that the network totally fails to understand 3D geometry. Our method shows rich 3D understanding, even in this real-world outdoor setting which is more complicated than other scenes. More qualitative images are included in the supplementary material. \vspace{-5pt}

\input{Figures/tex/fig8}
\subsection{Ablation Study}
\paragraph{Ablation on core components.}

\begin{wraptable}{r}{0.48\linewidth}
    \centering
    \vspace{-20pt}
    \caption{\textbf{Ablation study.}}
    % \vspace{-5pt}
    \resizebox{\linewidth}{!}{
    \begin{tabular}{ll|cccc}
    \toprule
    &Components & PSNR$\uparrow$ & SSIM$\uparrow$ & LPIPS$\downarrow$\\
    \midrule\midrule
    \textbf{(a)}& Baseline~\cite{fridovich2023k} & 18.70 & 0.708 & 0.400 \\
    \textbf{(b)}& \textbf{(a)} + $\mathcal{L}_\mathrm{seen}$ & 19.71 & 0.730 & 0.380 \\
    \textbf{(c)}& \textbf{(b)} + $\mathcal{L}_\mathrm{unseen}$ & 21.21 & 0.758 & 0.333 \\ 
    \textbf{(d)}& \textbf{(c)} + MDE Adapt. ($\mathcal{L}_\text{MDE}$) & 21.39 & 0.763 & 0.327 \\ 
    \textbf{(e)}& \textbf{(d)} + Conf. Modeling (\ours) & \textbf{21.58} &  \textbf{0.765} & \textbf{0.325}  \\
    \bottomrule
    \end{tabular}}
    \vspace{-5pt}
    \label{tab:main_abl}
\end{wraptable}

In Tab.~\ref{tab:main_abl} and Fig.~\ref{qual_nonoverlapping_scannet}, we evaluate the effect of each proposed component. The quantitative results show effectiveness of each component. For qualitative results, we found out that $\mathcal{L}_\mathrm{unseen}$ suppresses the artifacts in novel viewpoint, compared to when only $\mathcal{L}_\mathrm{seen}$ is given. With adaptation of the MDE network to this scene, red basket in the background shows more accurate results and artifacts near the table are removed. In our model, with confidence modeling, view synthesis results show to be more structurally confident in the overall scene. \vspace{-5pt}

\paragraph{Analysis of local fitting.}
\begin{wraptable}{r}{0.39\linewidth}
    \vspace{-17pt}
    \centering
    \caption{\textbf{Local fitting ablation.}}
    % \vspace{-5pt}
    \resizebox{\linewidth}{!}{
    \begin{tabular}{l|cccc}
    \toprule
      Components & PSNR$\uparrow$ & SSIM$\uparrow$ & LPIPS$\downarrow$\\
      \midrule\midrule
Baseline~\cite{fridovich2023k} & 18.65 & 0.706 & 0.502 \\
        w/ global fitting   & 19.05 & 0.698 & 0.399  \\
        w/ local fitting (\ours) & \textbf{19.71} & \textbf{0.730} & \textbf{0.380}   \\
        \bottomrule
    \end{tabular}}
    \vspace{-5pt}
    \label{tab:local_analysis}
\end{wraptable}

In Tab.~\ref{tab:local_analysis}, we further investigate the effectiveness of local scale-shift fitting and global scale-fitting. For global scale-shift fitting, we give learnable scale and shift parameters for each input image and convert MDE's output to the absolute value in a global manner. For a fair comparison, we compare our model only with MDE distillation on seen viewpoints. The results show that our method local scale-shift fitting is more effective on giving accurate depth supervision.
\vspace{-5pt}

%% file: Figures/tex/fig5.tex
\begin{figure*}[t]
\begin{center}
    \renewcommand{\thesubfigure}{}
    \subfigure[]
    {\includegraphics[width=0.195\textwidth]{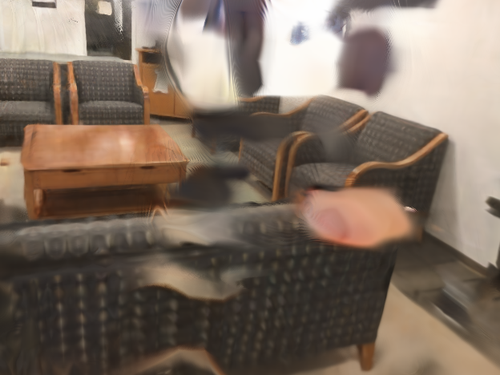}}
    \subfigure[]
    {\includegraphics[width=0.195\textwidth]{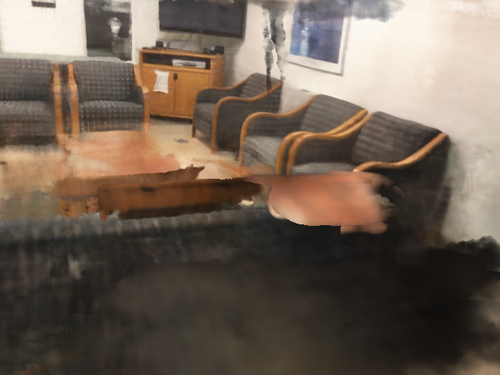}}
    \subfigure[]
    {\includegraphics[width=0.195\textwidth]{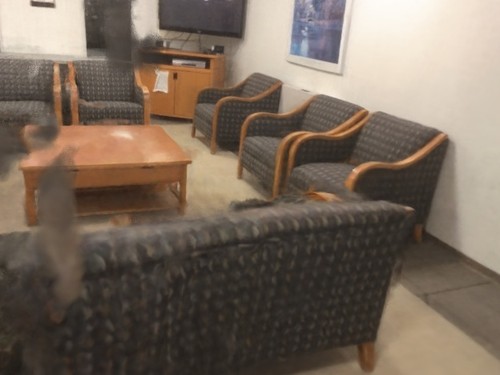}}
    \subfigure[]
    {\includegraphics[width=0.195\textwidth]{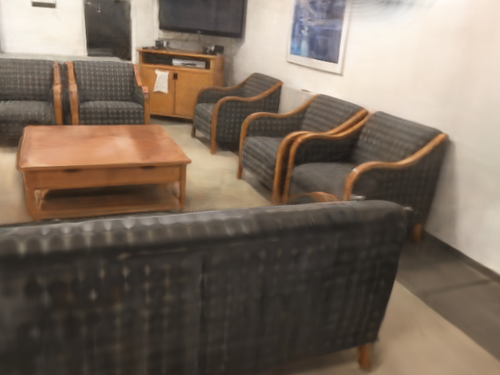}}
    \subfigure[]
    {\includegraphics[width=0.195\textwidth]{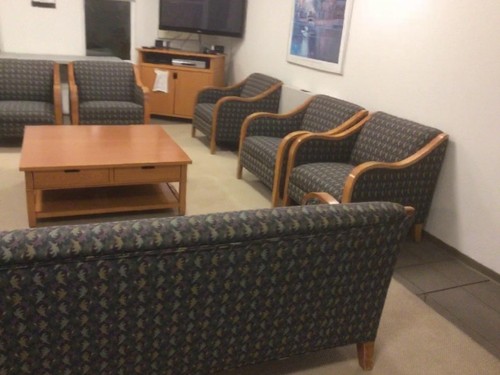}}
    \hfill\\\vspace{-20.5pt}
    \subfigure[(a) $K$-planes~\cite{fridovich2023k}]
    {\includegraphics[width=0.195\textwidth]{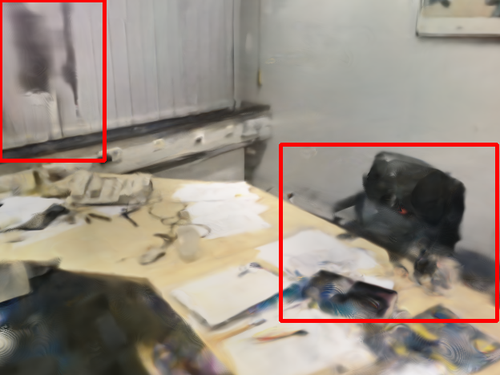}}
    \subfigure[(b) DS-NeRF~\cite{kangle2021dsnerf}]
    {\includegraphics[width=0.195\textwidth]{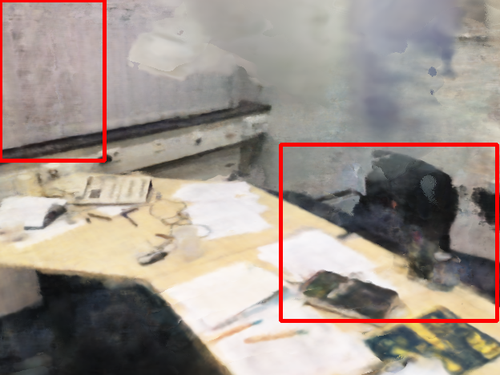}}
    \subfigure[(c) DDP-NeRF~\cite{roessle2021dense}]
    {\includegraphics[width=0.195\textwidth]{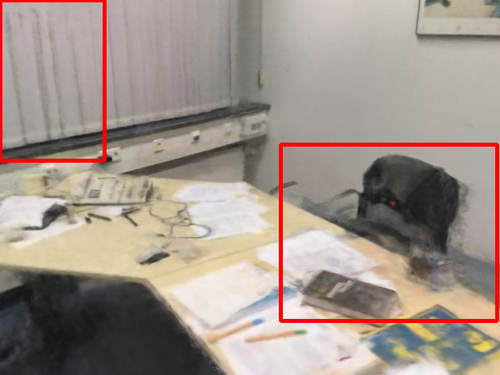}}
    \subfigure[(d) \ours]
    {\includegraphics[width=0.195\textwidth]{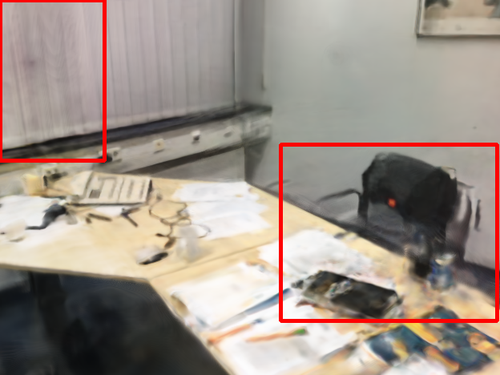}}
    \subfigure[(e) Ground truth]
    {\includegraphics[width=0.195\textwidth]{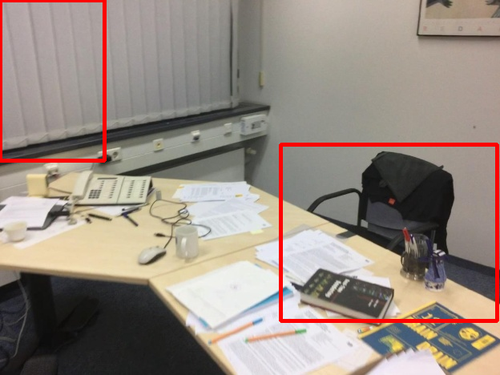}}\\
    \vspace{-5pt}
    \caption{\textbf{Qualitative results of on ScanNet~\cite{Dai_2017_CVPR} with 18 - 20 input views}.}
    \vspace{-15pt}
    \label{qual:scan20}
\end{center}
\end{figure*}

%% file: Figures/tex/fig6.tex
\begin{figure}
\begin{center}
    \renewcommand{\thesubfigure}{}
    \subfigure[]
    {\includegraphics[width=0.195\textwidth]{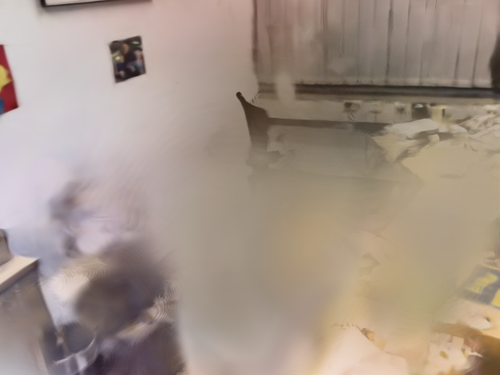}}
    \subfigure[]
    {\includegraphics[width=0.195\textwidth]{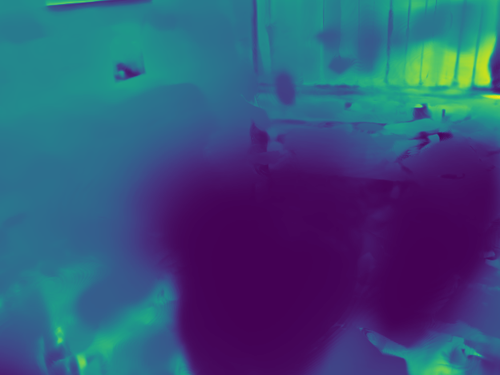}}
    \subfigure[]
    {\includegraphics[width=0.195\textwidth]{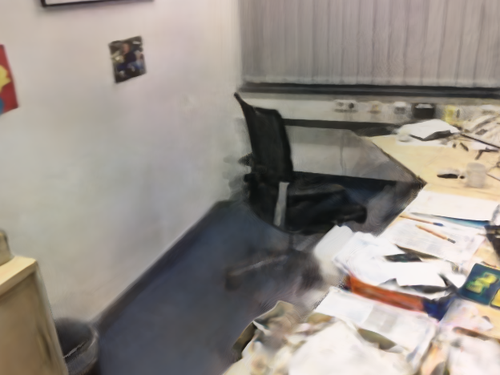}}
    \subfigure[]
    {\includegraphics[width=0.195\textwidth]{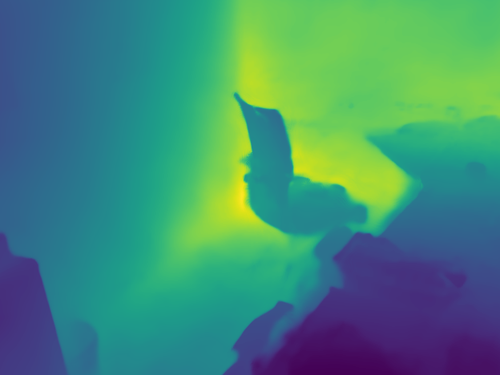}}
    \subfigure[]
    {\includegraphics[width=0.195\textwidth]{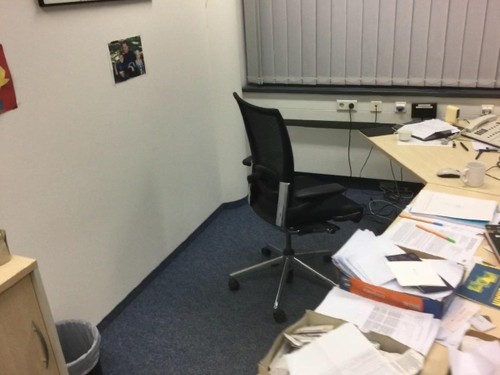}}
    \hfill\\\vspace{-20.5pt}
    \subfigure[(a) Baseline~\cite{fridovich2023k}]
    {\includegraphics[width=0.195\textwidth]{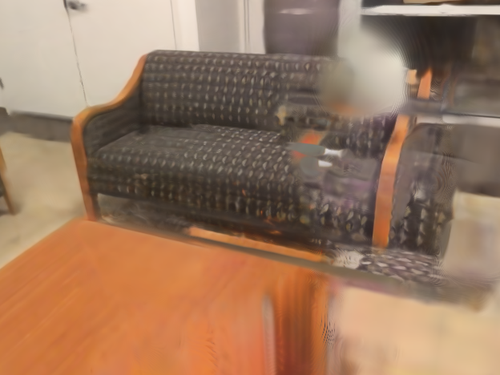}}
    \subfigure[(b) Baseline- Depth]
    {\includegraphics[width=0.195\textwidth]{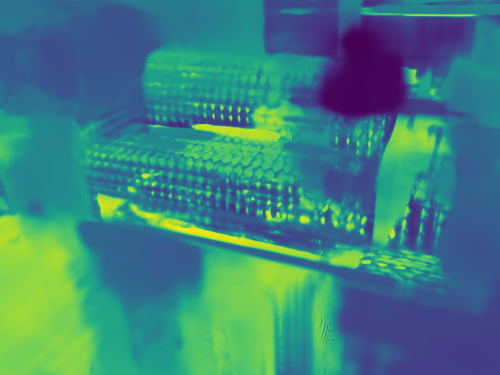}}
    \subfigure[(c) \ours]
    {\includegraphics[width=0.195\textwidth]{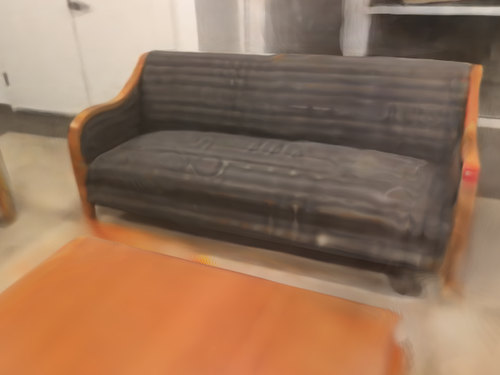}}
    \subfigure[(d) \ours- Depth]
    {\includegraphics[width=0.195\textwidth]{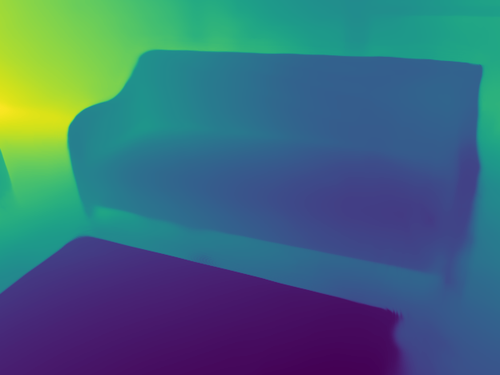}}
    \subfigure[(e) Ground truth]
    {\includegraphics[width=0.195\textwidth]{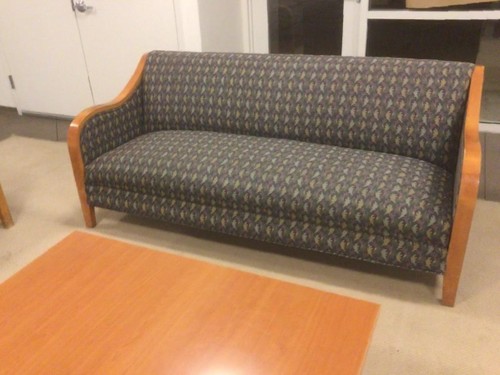}}
    \vspace{-5pt}
    \caption{\textbf{Qualitative results on ScanNet~\cite{Dai_2017_CVPR} with 9 - 10 input views}.} 
    \vspace{-19pt}
    \label{qual:scan10}
\end{center}
\end{figure}

%% file: Figures/tex/fig8.tex
\begin{figure*}[t]
\begin{center}
    \renewcommand{\thesubfigure}{}
    \subfigure[(a) Baseline~\cite{fridovich2023k}]
    {\includegraphics[width=0.195\textwidth]{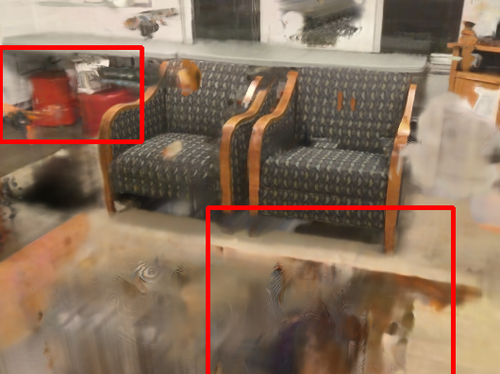}}
    \subfigure[(b) (a) + $\mathcal{L}_\text{seen}$]
    {\includegraphics[width=0.195\textwidth]{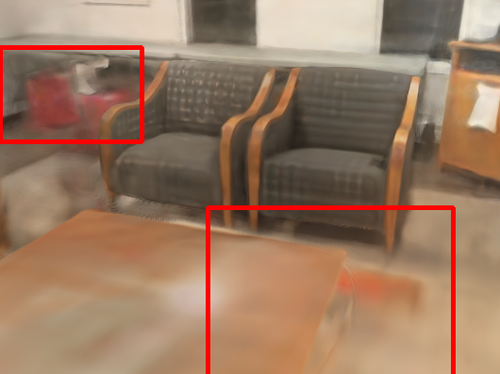}}
    \subfigure[(c) (b) + $\mathcal{L}_\text{unseen}$]
    {\includegraphics[width=0.195\textwidth]{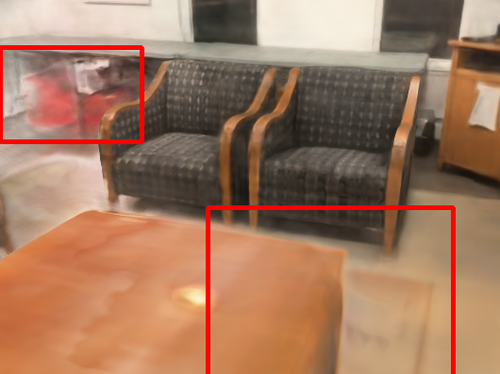}}
    \subfigure[(d) (c) + $\mathcal{L}_\text{MDE}$]
    {\includegraphics[width=0.195\textwidth]{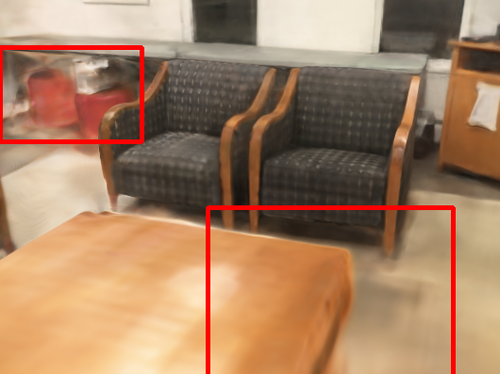}}
    \subfigure[(e) (d) + $M$ (\ours)]
    {\includegraphics[width=0.195\textwidth]{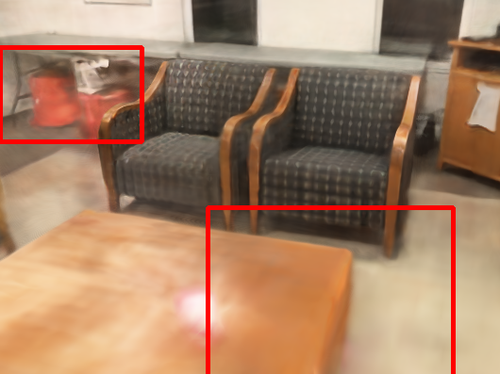}}
    
    \vspace{-5pt}
    \caption{\textbf{Visualization of ablation studies on ScanNet~\cite{Dai_2017_CVPR}.}}
    \vspace{-15pt}
    \label{qual_nonoverlapping_scannet}
\end{center}
\end{figure*}

%% file: Writing/6_conclusion.tex
\section{Conclusion}
\vspace{-5pt}
We propose \ours, a novel method that addresses the limitations of NeRF in few-shot settings by fully leveraging the ability of monocular depth estimation networks. By integrating MDE's geometric priors, \ours achieves robust optimization of few-shot NeRF, improving geometry reconstruction and artifact removal in both unseen and seen viewpoints. We further introduce patch-wise scale-shift fitting for accurate mapping of local depths to 3D space, and adapt MDE to NeRF's absolute scaling and multiview consistency, by distilling NeRF's absolute geometry to monocular depth estimation. Through complementary training, \ours establishes a strong synergy between MDE and NeRF, leading to a state-of-the-art performance in few-shot NeRF. Extensive evaluations on real-world scene datasets demonstrate the effectiveness of \ours.

%% file: Writing/7_suppl.tex
\begin{center}
	\textbf{\Large Appendix}
\end{center}

\input{Suppl_Writing/0_summary}
\input{Suppl_Writing/1_detail}
\input{Suppl_Writing/2_more_details}

\newpage
\input{Suppl_Writing/3_analysis}
\newpage
\input{Suppl_Writing/4_qual}
\newpage
\input{Suppl_Writing/5_others}

%% file: Suppl_Writing/0_summary.tex
\setcounter{figure}{0}
\setcounter{equation}{0}
\setcounter{table}{0}

%% file: Suppl_Writing/1_detail.tex
\section{Implementation Details}\label{supp:details}
\subsection{Architecture}\label{supp:details:arch}

We implement \ours with $K$-planes~\cite{fridovich2023k} as the base model. It represents a radiance field using tri-planes with three multi-resolutions for each plane: 128, 256, and 512 in both height and width, and 32 in feature depth. This approach also incorporates small MLP decoders and a two-stage proposal sampler.
It should be noted that our framework is not restricted to the $K$-planes baseline, but can be incorporated into any NeRF backbone models~\cite{mildenhall2020nerf, muller2022instant,li2022nerfacc}. In our experiments, we implemented our framework on top of the $K$-planes hybrid version codebase due to its quality, reasonable optimization speed, and model size. For the monocular depth estimation (MDE) module, we choose the pre-trained DPT~\cite{ranftl2021vision} as our base MDE model due to its powerful generalization ability in a zero-shot setting. Trained on very large datasets, DPT demonstrates impressive prediction quality and generalizes well to novel scenes. However, any MDE model can be utilized within our framework~\cite{yin2021learning, ranftl2020towards, ranftl2021vision}.

\subsection{Training details}\label{supp:details:train}
We use the Adam optimizer~\cite{kingma2015adam} and a cosine annealing with warm-up scheduler for NeRF optimization. The learning rate is set to $1\cdot10^{-2}$, and we perform 512 warm-up steps. For MDE adaptation, we also employ the Adam optimizer~\cite{kingma2015adam} with a learning rate of $1\cdot10^{-5}$. NeRF optimization is performed with a pixel batch size of 4,096, totaling 20K iterations. For $\mathcal{L}_\text{seen}$, we render a $64\times64$ patch, while for $\mathcal{L}_\text{unseen}$, we render a $128\times128$ patch with a stride of 3.

For the loss functions, we set the coefficients of $\mathcal{L}_\text{seen}$, $\mathcal{L}_\text{MDE}$, and $\mathcal{L}_\text{reg}$ as 0.01, 0.01, and 0.1, respectively. During the warm-up stage of 5,000 steps, the coefficient of $\mathcal{L}_\text{unseen}$ is initially set to 0 and then increased to 0.01 after 5,000 warm-up steps. For the first 1,000 steps, we employ the ranking loss~\cite{yin2021learning} with a coefficient of 0.1, in addition to $\mathcal{L}_\text{seen}$. All experiments were conducted using a single NVIDIA GeForce RTX 3090. The training process takes approximately 3 hours.

\subsection{Training loss details}
In the following, we describe a least-square alignment~\cite{wu2021toward} used in loss functions for MDE prior distillation in detail. As described in the main paper, we use a scale-shift invariant loss~\cite{ranftl2020towards} with patch-wise adjustment for depth consistency as follows:
\begin{equation}
\mathcal{L} = \sum_{I_i\in\mathcal{S}} \sum_{\mathbf{p} \in \mathcal{P}} {\|  (w_i D^*_{i}(\mathbf{p}) + q_i) - \bar{D}_i(\mathbf{p}) \|}, \label{supp:eq:depth_prior}
\end{equation}
where $w_i$ and $q_i$ are scale and shift values that align $D^*_{i}(\mathbf{p})$ to the absolute locations of $\bar{D}_i(\mathbf{p})$. In this loss function, to calculate $w_i$ and $q_i$, we following least-squares criterion~\cite{ranftl2020towards}:
\begin{equation}
(w_i,q_i) = \arg \min_{w_i,q_i} \sum_{\mathbf{p} \in \mathcal{P}} {\|  (w_iD^*_{i}(\mathbf{p}) + q_i) - \bar{D}_i(\mathbf{p}) \|}
\end{equation}
In other words, we can rewrite the above scheme as a closed problem. 
Let $\mathbf{h}_i=[w_i,q_i]^T$ and $\Vec{D}_i(\mathbf{p}) = [D^*_{i}(\mathbf{p}),1]^T$, then we can modify our problem as
\begin{equation}
\mathbf{h}_i^\text{opt} = \arg \min_{\mathbf{h}_i}  \sum_{\mathbf{p} \in \mathcal{P}} (\Vec{D}_i(\mathbf{p})^T\mathbf{h}_i-\bar{D}_i(\mathbf{p}))^2,
\end{equation}
which can be solved as follows:
\begin{equation}
\mathbf{h}_i^\text{opt} = (\sum_{\mathbf{p} \in \mathcal{P}}\Vec{D}_i(\mathbf{p})\Vec{D}_i(\mathbf{p})^T)^{-1}(\sum_{\mathbf{p} \in \mathcal{P}}\Vec{D}_i(\mathbf{p})\bar{D}_i(\mathbf{p}))
\end{equation}

\subsection{Baseline implementations}\label{supp:details:basline}
We directly use quantitative results reported in prior literature~\cite{uy2023scade} for the comparison of NerfingMVS~\cite{wei2021nerfingmvs}, DS-NeRF~\cite{kangle2021dsnerf} and DDP-NeRF~\cite{roessle2021dense}. As the setting~\cite{uy2023scade} requires out-of-domain priors, it should be noted that the results for DDP-NeRF are with out-of-domain priors. The results of DDP-NeRF with in-domain priors are 20.96, 0.737, and 0.236 for PSNR, SSIM, and LPIPS, respectively. However, we were unable to evaluate DDP-NeRF in the extreme settings of ScanNet and Tanks and Temples, as reliable COLMAP 3D points could not be obtained.

We utilized the authors' provided official implementations of RegNeRF~\cite{Niemeyer2021Regnerf} and $K$-planes~\cite{fridovich2023k}, training one model for each scene using two different scenarios on the ScanNet~\cite{Dai_2017_CVPR} and Tanks and Temples~\cite{knapitsch2017tanks} datasets. However, since there is no official code available for SCADE~\cite{uy2023scade}, we are unable to provide performance comparisons for this method.

%% file: Suppl_Writing/2_more_details.tex
\section{Datasets and Metrics}\label{supp:dnm}
\subsection{Datasets}\label{supp:dnm:datasets}
\paragraph{ScanNet~\cite{Dai_2017_CVPR}.} We adhere to the few-shot protocol provided by DDP-NeRF~\cite{roessle2021dense} in our experimental setup. 
We noticed that the split contained major overlaps across the train and test sets, which makes the task easier compared to realistic few-shot settings where images exhibit minimal overlap. For this reason, we construct an extreme few-shot scenario, using only half of the training images while maintaining the same test set.
\vspace{-5pt}

\paragraph{Tanks and Temples~\cite{knapitsch2017tanks}.}
To test the robustness of our method in challenging real-world outdoor environments, we conduct further experiments on Tanks and Temples dataset, an real-world outdoor dataset acquired under drastic lighting effects and reflectances. As no existing protocols exist for a few-shot scenario for this dataset, we introduce a new split for the few-shot setting.
We carefully selected 5 object-centric scenes —truck, francis, family, lighthouse, and ignatius— with inward-facing cameras. From each scene, we sample 10 training images that capture the overall geometry of the whole scene. For testing, we use one-eighth of the dataset as a test set, consisting every 8$^{th}$ repeating image from the entire image set.
We run COLMAP~\cite{Schonberger_2016_CVPR} on all images to obtain camera poses for NeRF training. 
However, for the lighthouse scene, which exhibits highly sensitive lighting and specular effects dependent on view pose, we manually preprocess the parts that contain these effects.
\subsection{Evaluation metrics}\label{supp:dnm:metrrics}
To evaluate the quality of novel view synthesis, following previous works~\cite{mildenhall2020nerf}, we measure PSNR, SSIM, and LPIPS. It is mentioned in $K$-planes that an implementation of SSIM from mip-NeRF~\cite{barron2022mip} results in lower values than standard scikit-image implementation. For a fair comparison per dataset, we use the latter scikit-image SSIM implementation following the relevant prior work. 

For the evaluation of the MDE module, we use 4 depth estimation metrics as follows:
\begin{itemize}
    \item AbsRel: $\frac{1}{|\mathcal{I}|} \sum_{\mathbf{p} \in \mathcal{I}} \|\bar{D}(\mathbf{p}) - D^{\mathrm{GT}}(\mathbf{p})\| / D^{\mathrm{GT}}(\mathbf{p})$;
    \item SqRel: $\frac{1}{|\mathcal{I}|} \sum_{\mathbf{p} \in \mathcal{I}} \|\bar{D}(\mathbf{p}) - D^{\mathrm{GT}}(\mathbf{p})\|^2 / D^{\mathrm{GT}}(\mathbf{p})$;
    \item RMSE: $\sqrt{\frac{1}{|\mathcal{I}|} \sum_{\mathbf{p}\in \mathcal{I}} \|\bar{D}(\mathbf{p}) - D^{\mathrm{GT}}(\mathbf{p})\|^2}$;
    \item RMSE log: $\sqrt{\frac{1}{|\mathcal{I}|} \sum_{\mathbf{p}\in \mathcal{I}} \|\log{\bar{D}(\mathbf{p})} - \log{D^{\mathrm{GT}}}\|^2}$;
\end{itemize}
where $\mathbf{p}$ is a pixel in the image $\mathcal{I}$ and $D^{\mathrm{GT}}$ is ground truth depth map. In addition, following~\cite{zhang2021consistent}, we use single scaling factor $s$ for each scene which is obtained by 
\begin{equation}
    s = {\frac{1}{N}}{\sum_{I_i\in\mathcal{S}}}(\mathrm{median}( D^{\mathrm{GT}}_{i}/ \bar{D_{i}})),
\end{equation}
rather than fit each frame to ground truth, to evaluate view consistency of MDE models. Here, $\mathcal{S}$ denotes set of images from single scene.

%% file: Suppl_Writing/3_analysis.tex
\section{Additional Analysis}\label{supp:analysis}
\subsection{Comparison of patch- and image-level scale-shift adjustment}\label{supp:analysis:local}
We provide additional analysis and visualization results regarding the patch-wise scale and shift adjustment. In Fig.\ref{supp:error:local} and Fig.\ref{supp:error:local3d}, we present error maps showing the discrepancies between the ground truth sensor depth and the predicted depth. Additionally, in Fig.\ref{supp:qual:local}, we present qualitative results of rendered color and depth using each fitting method. It is important to note that in the image-level fitting scheme, a single set of scale and shift values is computed for an entire depth map. Conversely, in our patch-level fitting method, scale and shift values are calculated individually for each $80\times80$ patch within the depth map. The error map clearly demonstrates the significant reduction in misalignment errors achieved by our patch-level fitting method compared to the image-level fitting approach.

For the comparison of image-level and patch-level fitting provided in the Fig.~\ref{fig2:patch} and Tab.~\ref{tab:local_analysis} of the main paper, we set the scale and shift as learnable parameters per image for image-level fitting and conduct patch-wise scale-shift invariant loss for patch-level fitting. This comparison is conducted only with $\mathcal{L}_\text{seen}$ given and results with patch-level fitting show better performance compared to image-level fitting. The difference between the two methods is especially distinguished in rendered depth maps of these two settings, in that patch-level fitting lets NeRF learn depth more accurately. \vspace{5pt}

\input{Suppl_Figures/_tex/local_fitting/MDE_fitting}
\input{Suppl_Figures/_tex/local_fitting/MDE_fitting_3dvis}
\newpage
\input{Suppl_Figures/_tex/local_fitting/NeRF_fitting}

\subsection{Confidence modeling}\label{supp:analysis:confidence}
\input{Suppl_Figures/_tex/confidence/confidence}
In Fig.~\ref{supp:fig:conf}, we demonstrate the effectiveness of our confidence modeling which effectively eliminates inaccurate information present in depth maps from both NeRF and the MDE network through leveraging multi-view consistency of NeRF. MDE depth from the input image contains errors, which can be filtered out by verifying consistency with depth from NeRF's other viewpoint. Likewise, the error of MDE depth from unseen viewpoint can be filtered through consistency check with MDE depth from the seen viewpoint. 

\newpage
\subsection{Ablation of MDE baselines}\label{supp:analysis:mde_basline}
\begin{table*}[!h]
    \centering
    \caption{\textbf{Ablation study on MDE baseline.}}
    \resizebox{0.5\linewidth}{!}{
    \begin{tabular}{l|ccc}
    \toprule
    Components & PSNR$\uparrow$ & SSIM$\uparrow$ & LPIPS$\downarrow$ \\
    \midrule\midrule
    \ours with LeReS~\cite{yin2021learning}   & 21.31 & 0.757 & 0.343 \\
    \ours with MiDaS~\cite{ranftl2020towards} & 21.48 & 0.758 & 0.337 \\
    \ours with DPT~\cite{ranftl2021vision} & 21.58 & 0.765 & 0.325 \\
    \bottomrule
    \end{tabular}}
    \label{supp:tab:mde_abl}
\end{table*}

We conduct an ablation on the Monocular Depth Estimation (MDE) network to assess its impact on our methodology.
Considering the recent advancements~\cite{ranftl2020towards, ranftl2021vision} in MDE models that shows strong generalization power for depth estimation in unseen images, we replace our MDE network with state-of-the-art models such as LeReS, MiDaS, and DPT. 
The results in Tab.~\ref{supp:tab:mde_abl} show that our method shows consistent performance across different baselines.

\subsection{Analysis of MDE Adaptation Loss.}
\begin{table*}[!h]
    \centering
    \caption{\textbf{Ablation study on MDE Adaptation Loss.}}
    \resizebox{0.9\linewidth}{!}{
    \begin{tabular}{l|ccc|cccc}
    \toprule
    Components & PSNR$\uparrow$ & SSIM$\uparrow$ & LPIPS$\downarrow$ & AbsRel $\downarrow$ & SqRel $\downarrow$ & RMSE $\downarrow$ & RMSE log $\downarrow$ \\
    \midrule\midrule
    scale-shift loss & 21.31 & 0.757 & 0.343 & 0.182 & 0.109 & 0.484 & 0.205\\
    $l$1 loss & 21.48 & 0.758 & 0.337 & 0.157 & 0.079 & 0.386 & 0.176\\
    \ours (Ours) & \textbf{21.58} & \textbf{0.765} & \textbf{0.325} & \textbf{0.151} &\textbf{ 0.071} & \textbf{0.356} & \textbf{0.168}\\
    \bottomrule
    \end{tabular}}
    \label{supp:tab:adapt_abl}
\end{table*}

In Tab.~\ref{supp:tab:adapt_abl}, we further investigate the effectiveness of scale-shift loss and $l$1 loss at MDE adaptation. Equation ~\ref{eq:depth_adaptation} revolves around the idea of adapting MDE toward predicting a scene-specific absolute geometry, which is achieved by the first addend term: this first term forces itself MDE to adapt towards multiview consistency so that its ill-posed nature is reduced and its initial global depth prediction grows to be more in accordance with the absolute geometry captured by NeRF. In contrast, the second addend term, which takes into account patch-wise scale-shift fitting, is designed to aid the modeling of fine, detailed, local geometry which the model has difficulty modeling without such local fitting. As shown in the results, when only one term is used for optimization (scale-shift or $l$1), it performs worse in every metric than in both are used in conjunction (ours). This justifies our strategy of using both losses as effective.

\newpage
\section{Camera Visualization of proposed Few-shot setting}\label{supp:results}
\vspace{45pt}
\input{Suppl_Figures/_tex/cam_visualization/cam_visualization}

%% file: Suppl_Figures/_tex/local_fitting/MDE_fitting.tex
\begin{figure}[h]
\centering
  \rotatebox[origin=C]{90}{\parbox{20mm}{\centering \small Image-Level}} 
  \hspace{-3.85pt}
  \mpage{0.23}{\includegraphics[width=\linewidth]{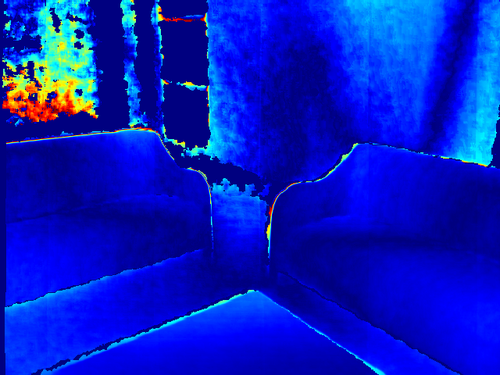}}
  \mpage{0.23}{\includegraphics[width=\linewidth]{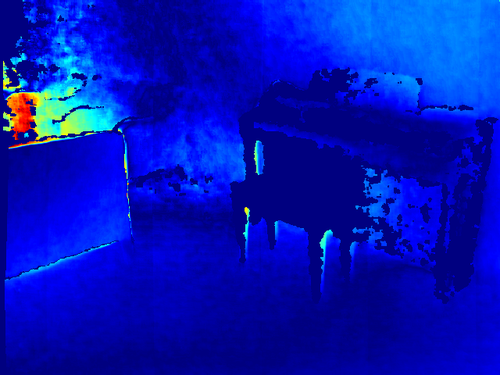}}
  \mpage{0.23}{\includegraphics[width=\linewidth]{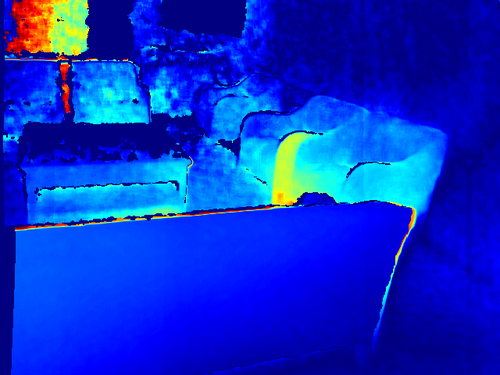}}
  \mpage{0.23}{\includegraphics[width=\linewidth]{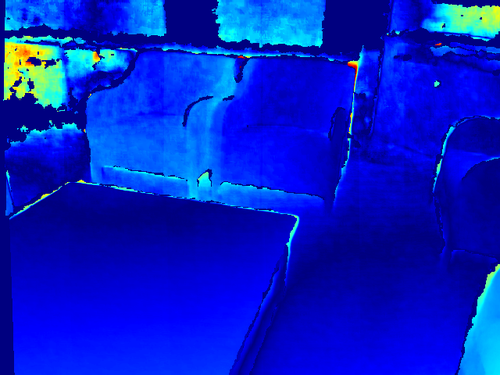}}
  \\
  \rotatebox[origin=C]{90}{\parbox{20mm}{\centering \small Patch-Level}} 
  \mpage{0.23}{\includegraphics[width=\linewidth]{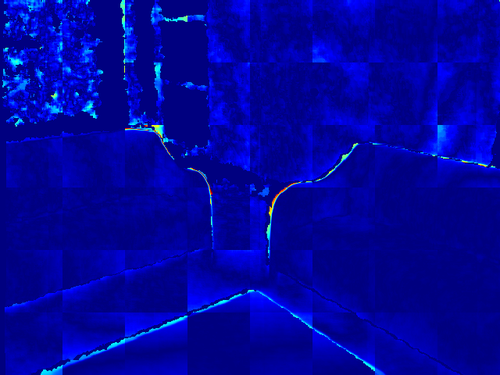}}
  \mpage{0.23}{\includegraphics[width=\linewidth]{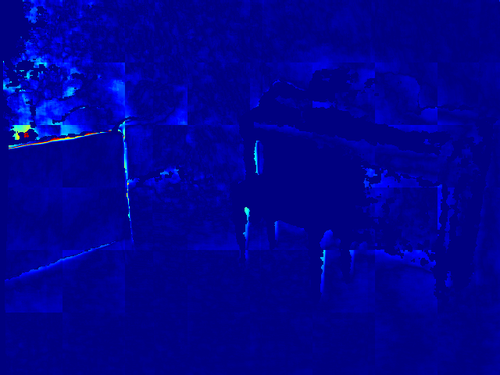}}
  \mpage{0.23}{\includegraphics[width=\linewidth]{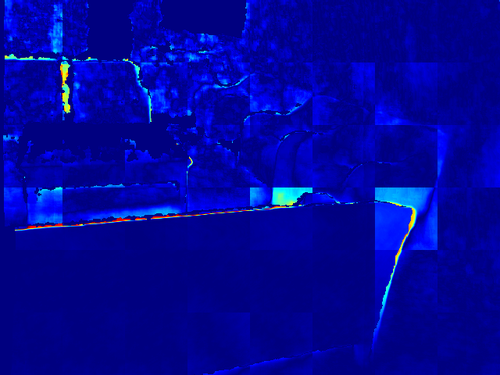}}
  \mpage{0.23}{\includegraphics[width=\linewidth]{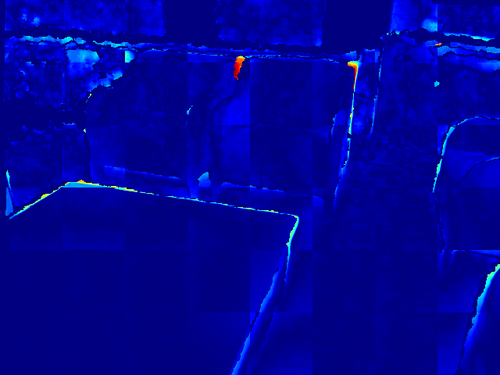}}
  \\
  \caption{\textbf{Error map visualization of image-level and patch-wise scale and shift adjustment:} relative depth map in various viewpoints is fitted in two ways, image-level fitting (first row) and patch-level fitting (second row).}
  \label{supp:error:local}
\end{figure}

%% file: Suppl_Figures/_tex/local_fitting/MDE_fitting_3dvis.tex
\begin{figure}[h]
\centering 
    \renewcommand{\thesubfigure}{}
     \subfigure[(a)]{
    \includegraphics[width=0.32\linewidth]{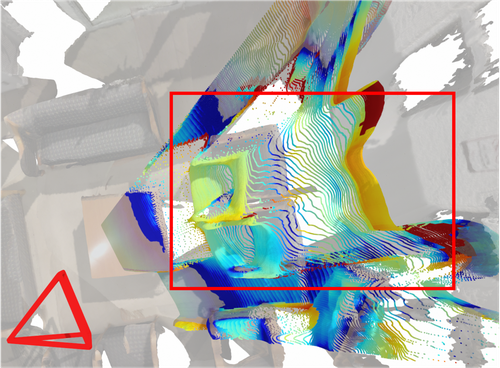}}
     \subfigure[(b)]{
    \includegraphics[width=0.32\linewidth]{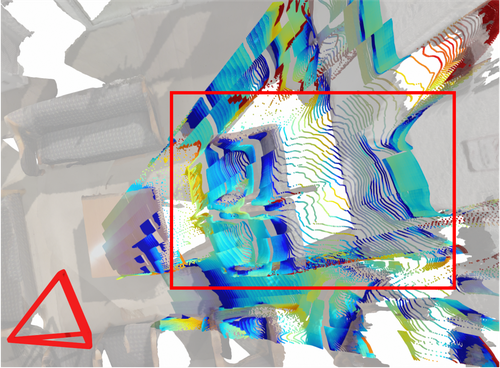}}
     \subfigure[(c)]{
    \includegraphics[width=0.32\linewidth]{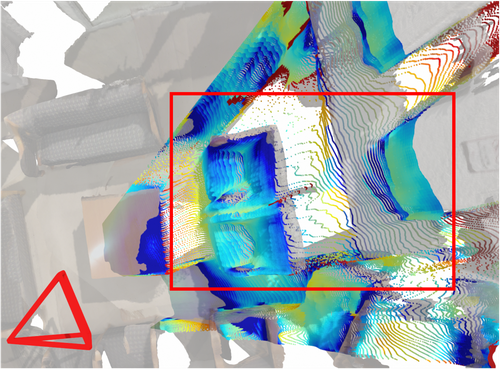}}\hfill\\
    \vspace{-5pt}
    \caption{\textbf{3D Visualization of the error map of MDE and NeRF:} (a) monocular depth with image-level adjustment, (b) monocular depth with patch-level adjustment, and (c) rendered depth by NeRF trained with patch-level adjustment. Depth from the input image of the viewpoint stated as red camera is adjusted in each ways and unprojected into 3D space.
    Error of each point cloud of a room is visualized from the bird’s eye view. This is done with jet color coding, so that red color means large error and blue color means small error. The proposed patch-wise adjustment helps to minimize the errors caused by inconsistency in depth differences among objects. }
    \label{supp:error:local3d}
\end{figure}

%% file: Suppl_Figures/_tex/local_fitting/NeRF_fitting.tex
\begin{figure}[h]
\centering
\begin{center}
    \renewcommand{\thesubfigure}{}
    \subfigure[]
    {\includegraphics[width=0.195\textwidth]{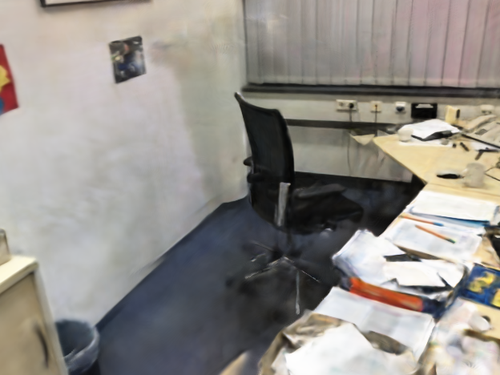}}
    \subfigure[]
    {\includegraphics[width=0.195\textwidth]{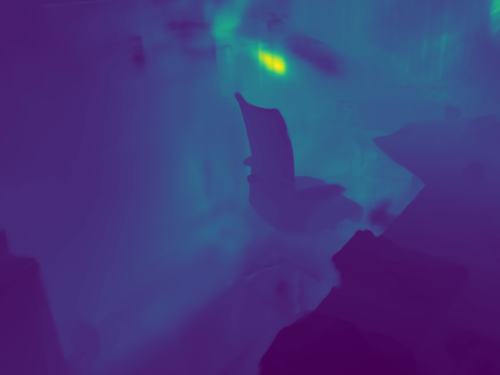}}
    \subfigure[]
    {\includegraphics[width=0.195\textwidth]{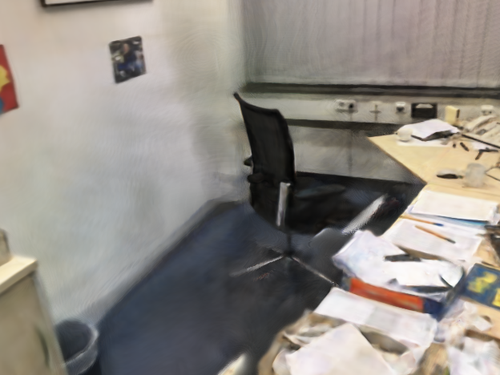}}
    \subfigure[]
    {\includegraphics[width=0.195\textwidth]{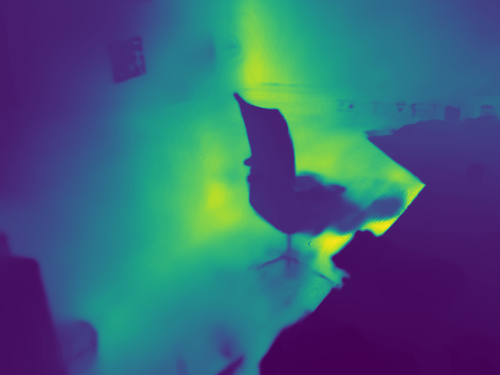}}
    \subfigure[]
    {\includegraphics[width=0.195\textwidth]{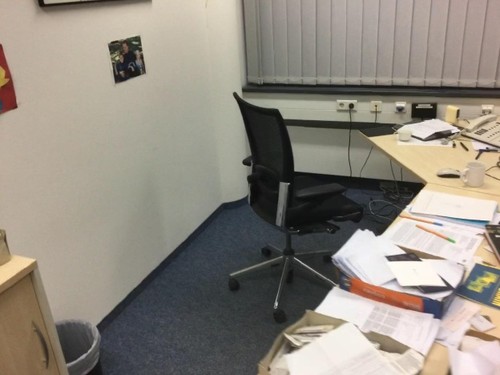}}
    \hfill\\\vspace{-20.5pt} 
    \subfigure[(a) Image-Level $\bar{I}$]
    {\includegraphics[width=0.195\textwidth]{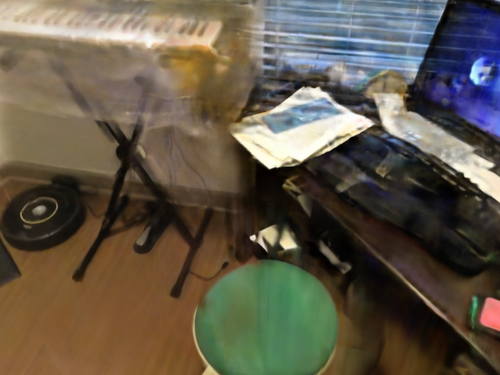}}
    \subfigure[(b) Image-Level $\bar{D}$]
    {\includegraphics[width=0.195\textwidth]{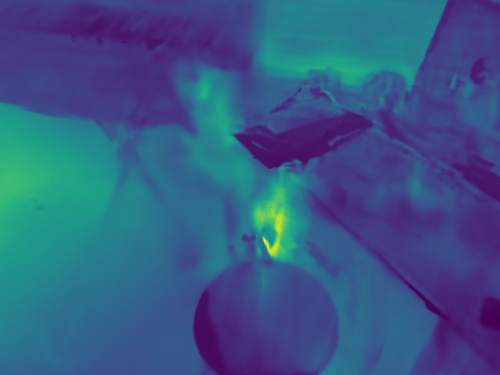}}
    \subfigure[(c) Patch-Level $\bar{I}$]
    {\includegraphics[width=0.195\textwidth]{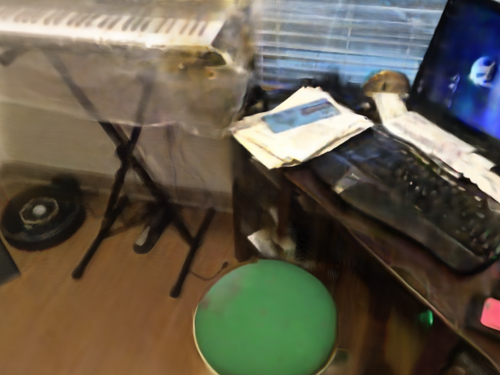}}
    \subfigure[(d) Patch-Level $\bar{D}$]
    {\includegraphics[width=0.195\textwidth]{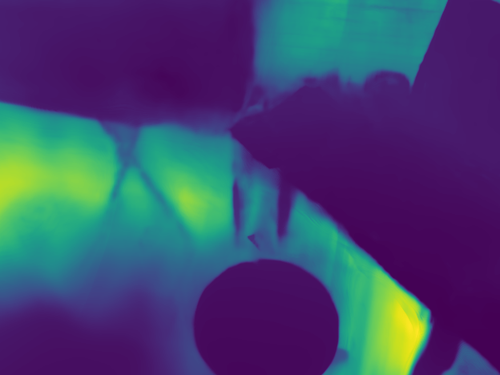}}
    \subfigure[(e) Ground-truth]
    {\includegraphics[width=0.195\textwidth]{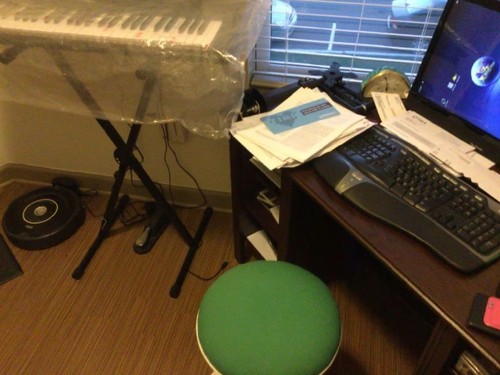}}\\
    \vspace{-5pt}
    \caption{\textbf{Comparison of patch- and image-level scale-shift adjustment.} Rendered color and depth from NeRF with (a-b) image-level scale and shift adjustment and (c-d) patch-level scale and shift adjustment.}
    \label{supp:qual:local}
\end{center}
\end{figure}

%% file: Suppl_Figures/_tex/confidence/confidence.tex
\begin{figure*}[h]
\centering
\begin{center}
    \renewcommand{\thesubfigure}{}
    \subfigure[(a) Input Image]
    {\includegraphics[width=0.23\textwidth]{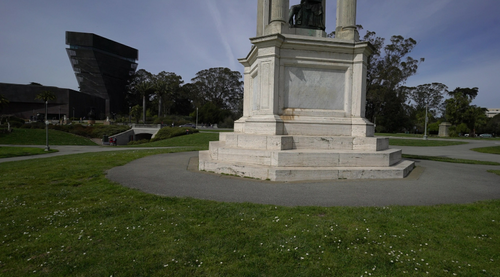}}
    \subfigure[(b) Initial MDE]
    {\includegraphics[width=0.23\textwidth]{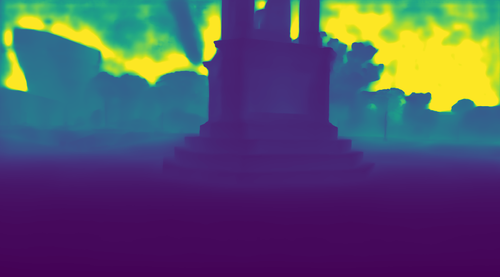}}
    \subfigure[(c) Conf. Mask]
    {\includegraphics[width=0.23\textwidth]{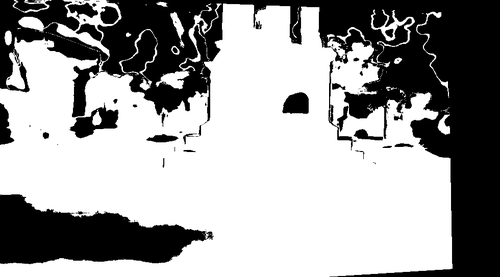}}
    \subfigure[(d) Masked MDE]
    {\includegraphics[width=0.23\textwidth]{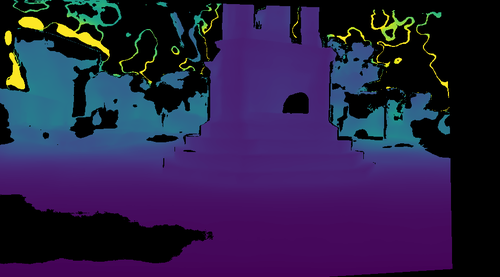}}\\
    \vspace{-5pt}\caption{\textbf{Comparions on MDE depth map with and without confidence masking.} the initial MDE depth map predicted is filtered through mask from our confidence modeling.}
    \label{supp:fig:conf}
\end{center}
\end{figure*}

%% file: Suppl_Figures/_tex/cam_visualization/cam_visualization.tex
\begin{figure*}[h]
\vspace*{\fill}%
\centering
  \rotatebox[origin=C]{90}{\parbox{20mm}{\centering \small Francis}} 
\hspace{-3.85pt}
  \mpage{0.47}{\includegraphics[width=0.75\linewidth]{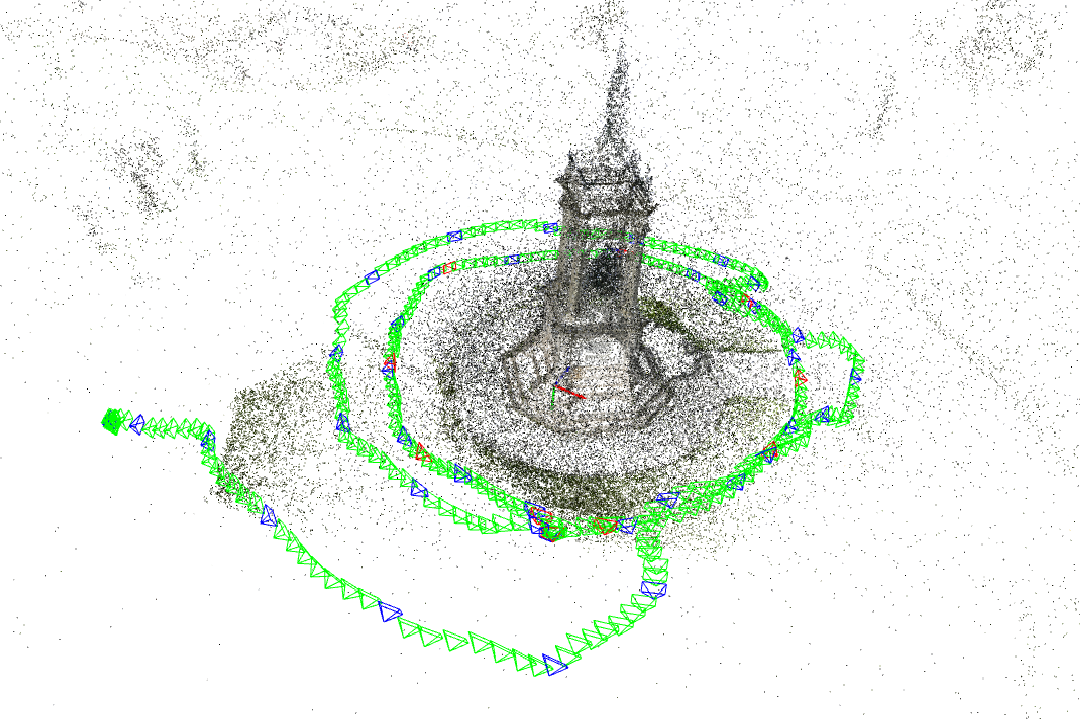}}
  \mpage{0.47}{\includegraphics[width=0.75\linewidth]{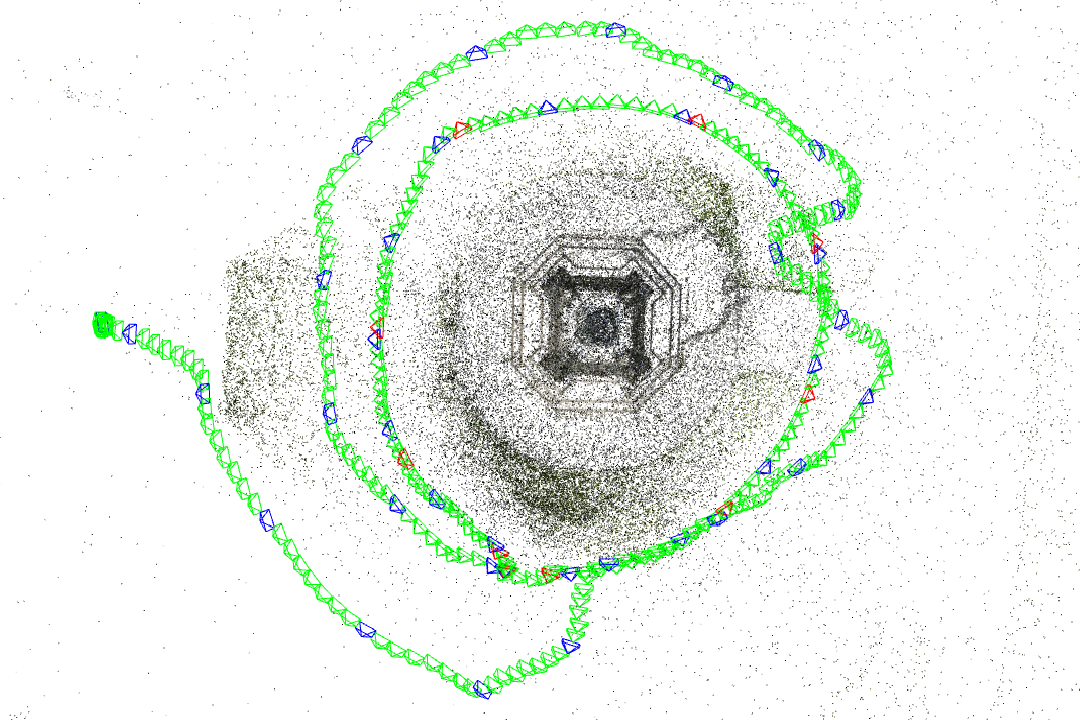}}
  \\
  \rotatebox[origin=C]{90}{\parbox{20mm}{\centering \small Family}} 
  \mpage{0.47}{\includegraphics[width=0.75\linewidth]{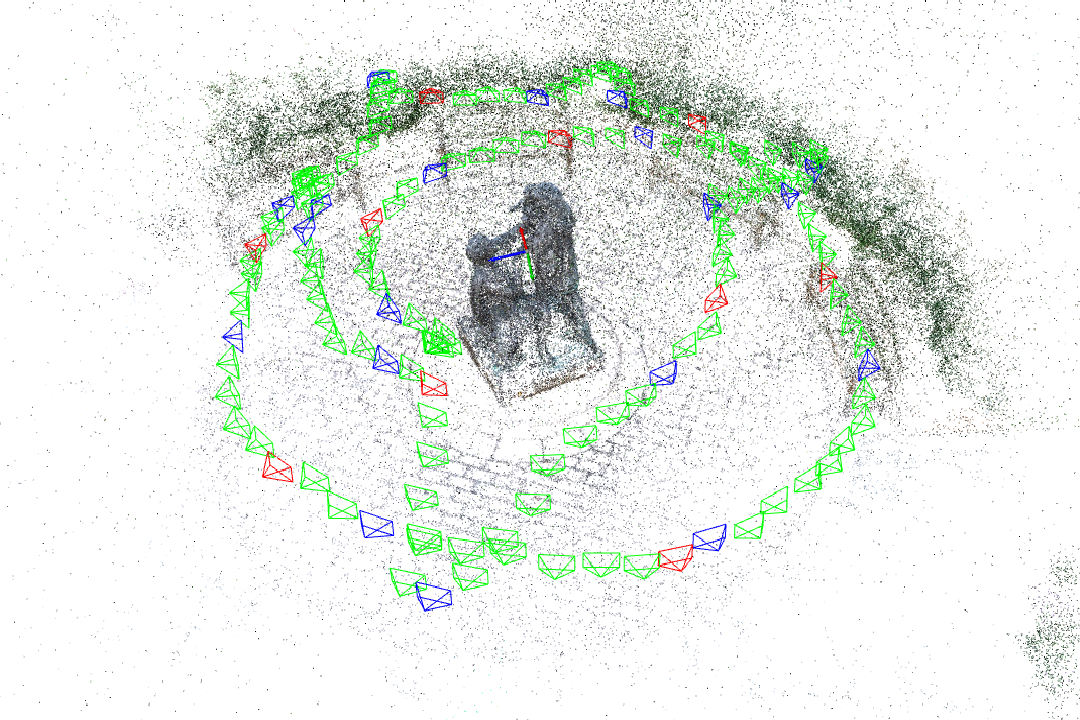}}
  \mpage{0.47}{\includegraphics[width=0.75\linewidth]{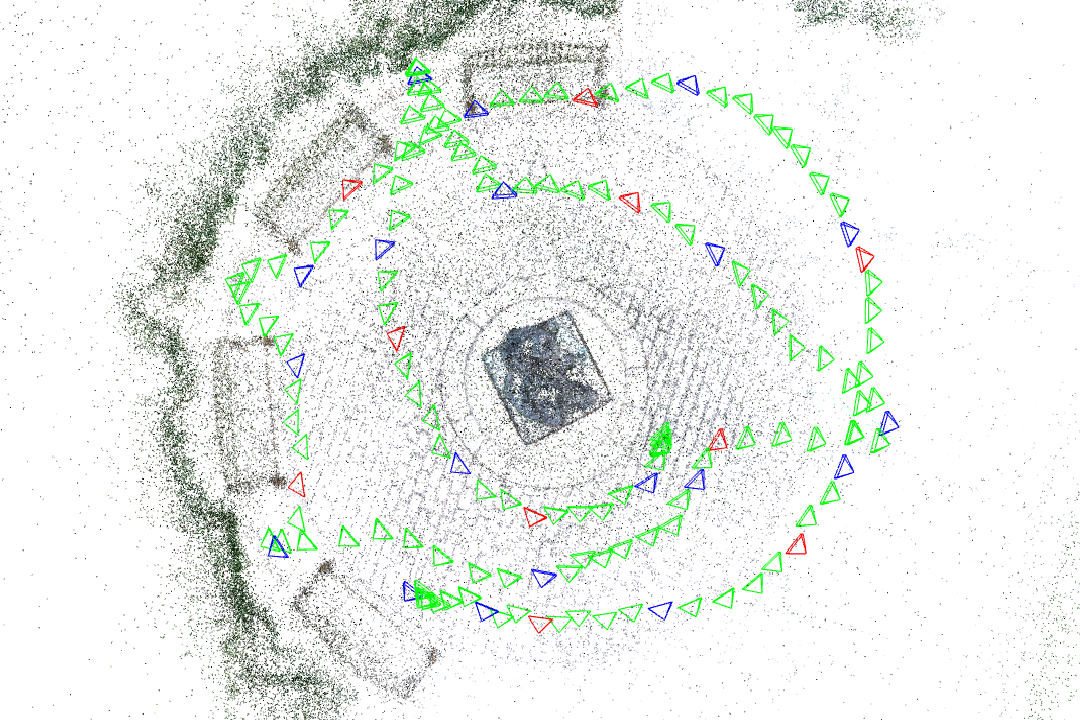}}
  \\
  \rotatebox[origin=C]{90}{\parbox{20mm}{\centering \small Ignatius}} 
  \mpage{0.47}{\includegraphics[width=0.75\linewidth]{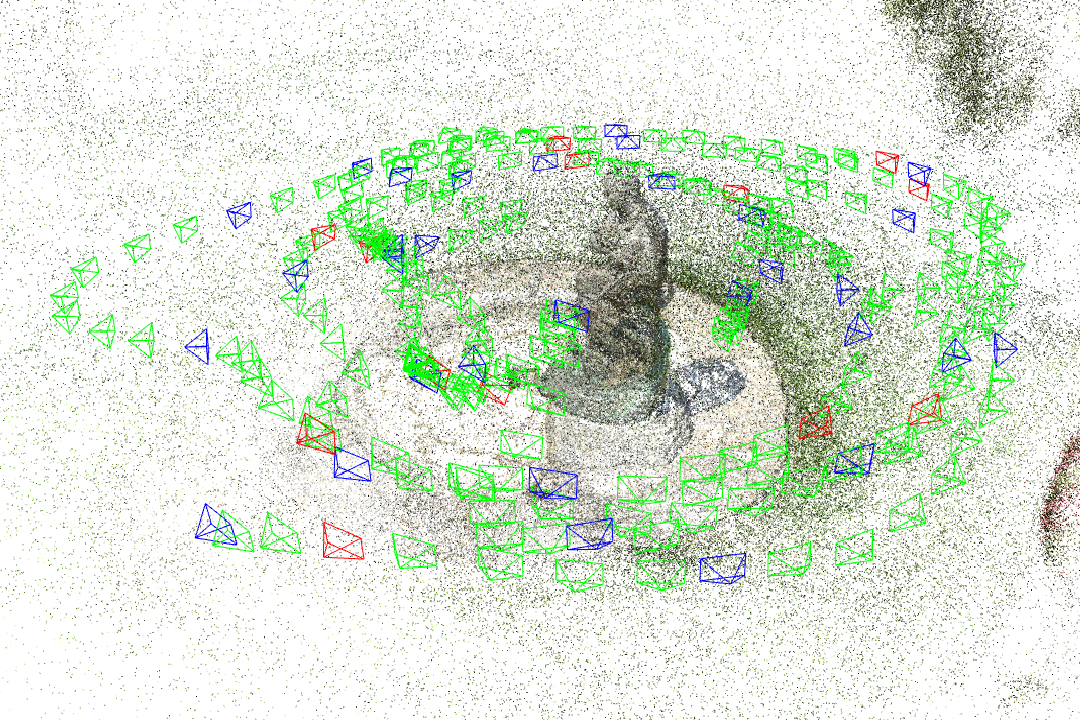}}
  \mpage{0.47}{\includegraphics[width=0.75\linewidth]{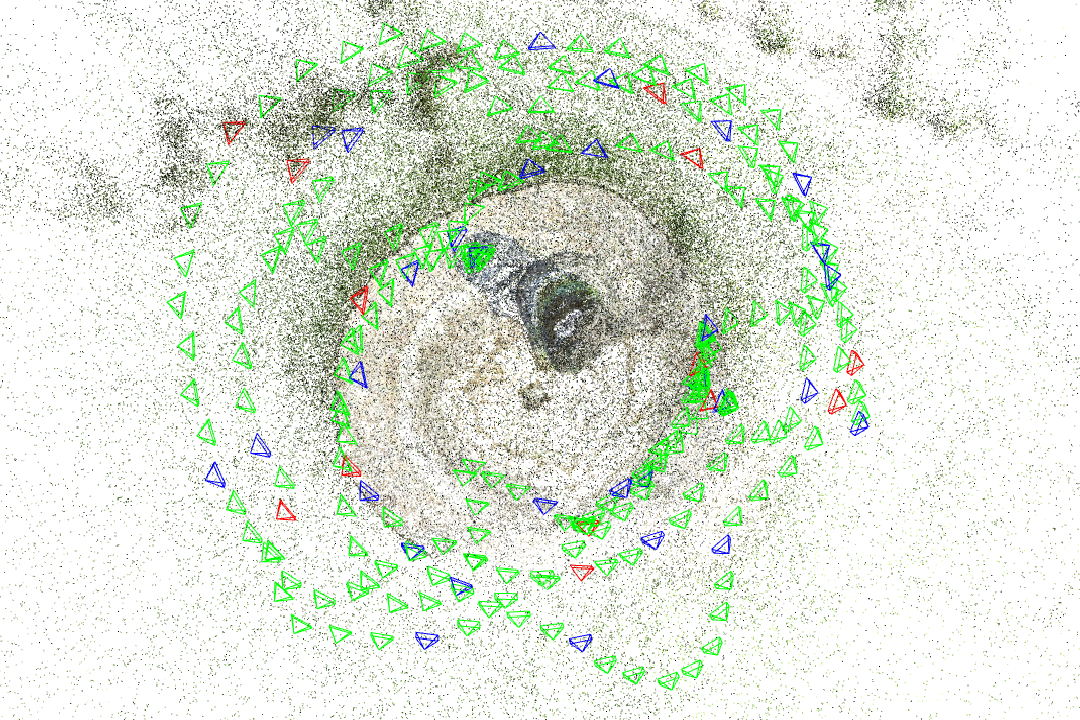}}
  \\
  \rotatebox[origin=C]{90}{\parbox{20mm}{\centering \small Lighthouse}} 
  \mpage{0.47}{\includegraphics[width=0.75\linewidth]{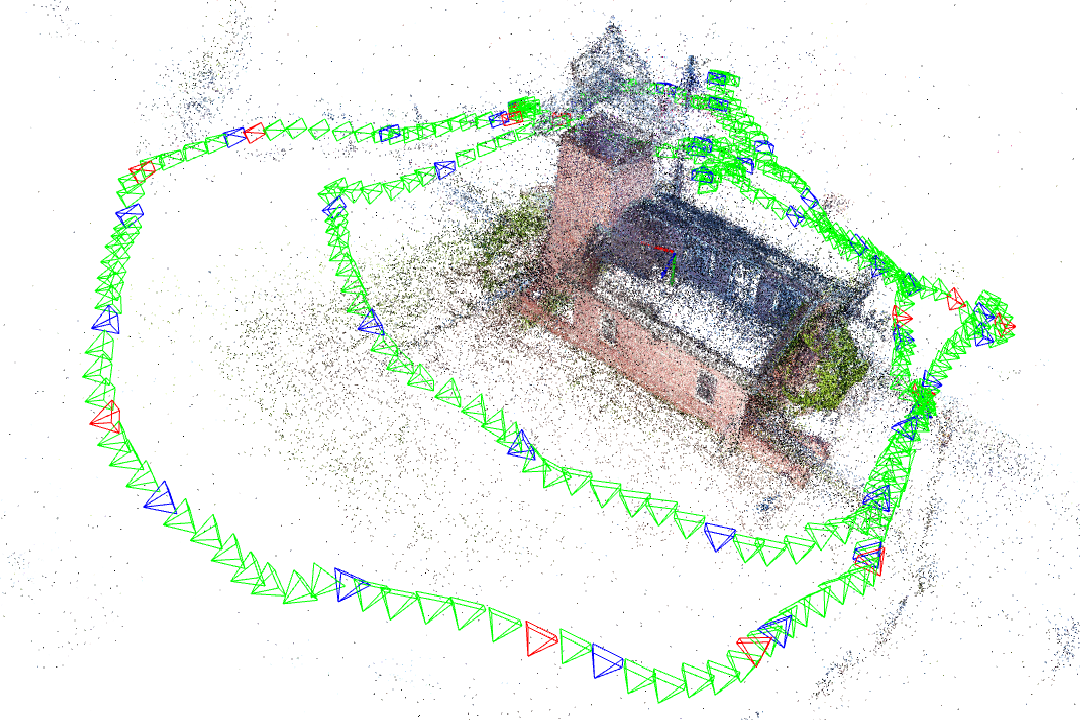}}
  \mpage{0.47}{\includegraphics[width=0.75\linewidth]{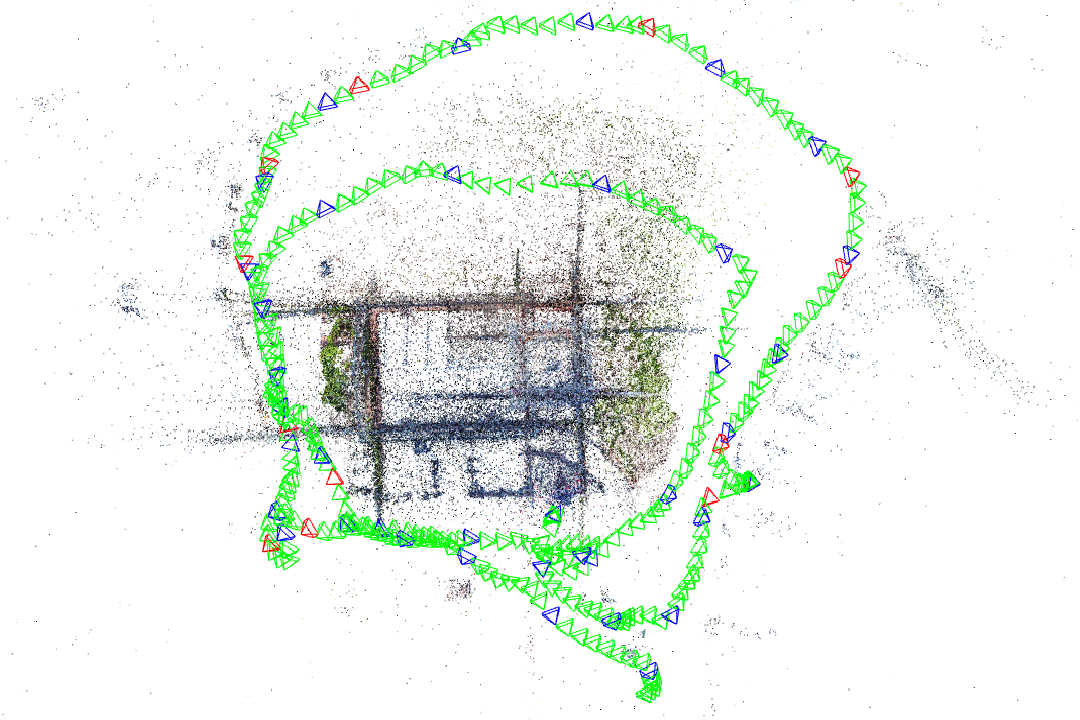}}
  \\
  \rotatebox[origin=C]{90}{\parbox{20mm}{\centering \small Truck}} 
  \mpage{0.47}{\includegraphics[width=0.75\linewidth]{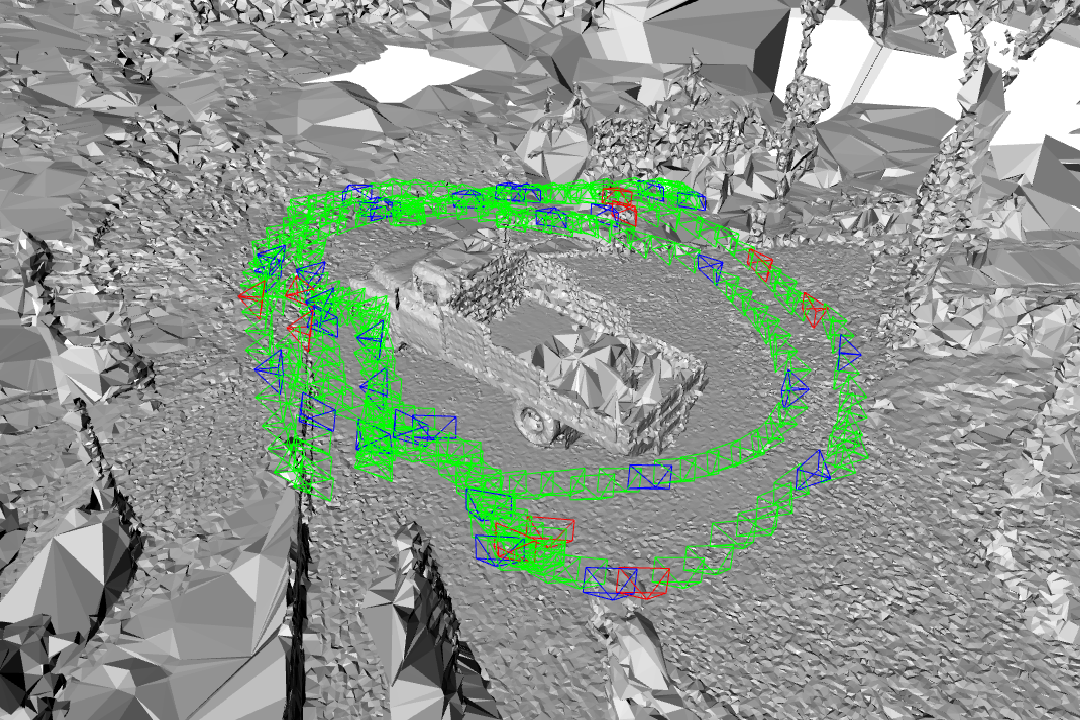}}
  \mpage{0.47}{\includegraphics[width=0.75\linewidth]{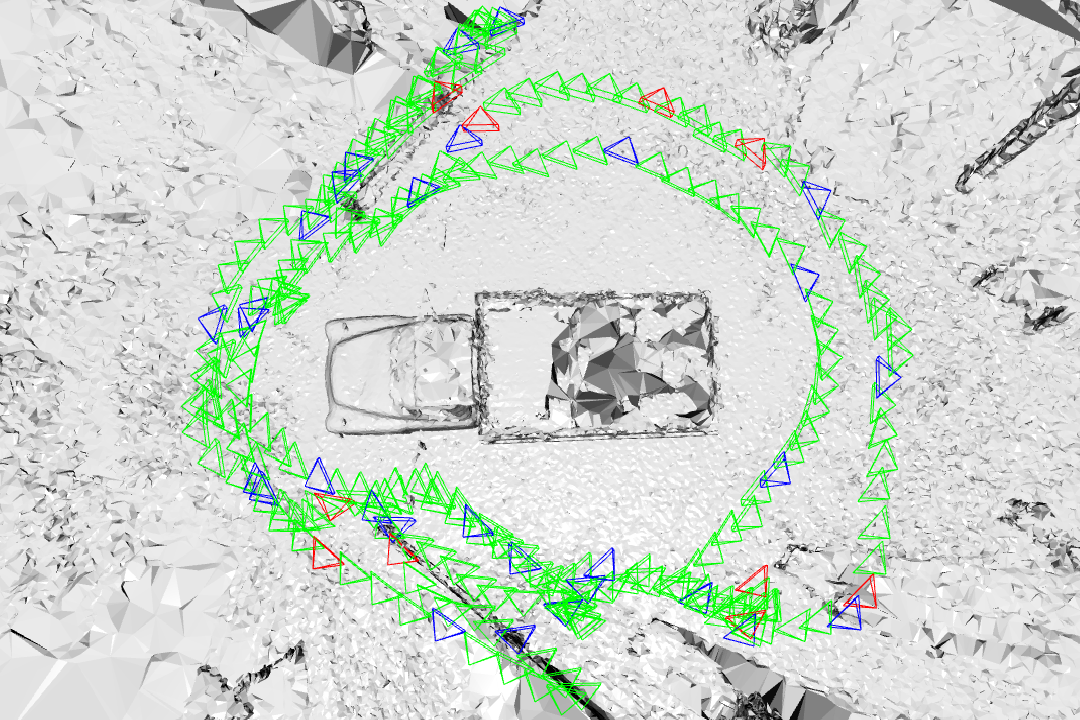}}
  \\
  \rotatebox[origin=C]{90}{\parbox{20mm}{\centering \small}} 
  \mpage{0.47}{Side view}
  \mpage{0.47}{Top view}
  \\
  
  \caption{\textbf{Camera visualization of Tanks and Temples dataset.} \textcolor{green}{Green} cameras mean \textcolor{green}{all} sets of images provided, and \textcolor{red}{red} and \textcolor{blue}{blue} cameras mean \textcolor{red}{train} and \textcolor{blue}{test} sets. As shown here, \textcolor{red}{red} cameras, i.e., train set, are a very small fraction of the camera set with little overlapping. On the other hand, \textcolor{blue}{blue} cameras, i.e., test set, cover various locations, distributed in various positions of the scene.}
  \vspace*{\fill}%
  \label{supp:cam}
\end{figure*}\vspace{-10pt}

%% file: Suppl_Writing/4_qual.tex
\section{Additional Qualitative Results}\label{supp:results}
In this section, we show additional qualitative comparisons in Fig.~\ref{qual:scan10_0710}, Fig.~\ref{qual:scan10_0758}, Fig.~\ref{qual:scan10_0781}, Fig.~\ref{qual:scan20_0710}, Fig.~\ref{qual:scan20_0758}, and Fig.~\ref{qual:scan20_0781} for ScanNet~\cite{Dai_2017_CVPR} dataset in two different settings and in Fig.~\ref{qual:truck}, Fig.~\ref{qual:francis}, Fig.~\ref{qual:lighthouse}, Fig.~\ref{qual:ignatius}, and Fig.~\ref{qual:family} for Tanks and Temples~\cite{knapitsch2017tanks} dataset.
\input{Suppl_Figures/_tex/Scan10/scan10_0710}
\input{Suppl_Figures/_tex/Scan10/scan10_0758}
\clearpage
\input{Suppl_Figures/_tex/Scan10/scan10_0781}
\input{Suppl_Figures/_tex/Scan20/scan20_0710}
\clearpage
\input{Suppl_Figures/_tex/Scan20/scan20_0758}
\input{Suppl_Figures/_tex/Scan20/scan20_0781}

\input{Suppl_Figures/_tex/TnT/TnT_truck}
\input{Suppl_Figures/_tex/TnT/TnT_francis}
\input{Suppl_Figures/_tex/TnT/TnT_lighthouse}
\input{Suppl_Figures/_tex/TnT/TnT_ignatius}
\input{Suppl_Figures/_tex/TnT/TnT_family}
\newpage

%% file: Suppl_Figures/_tex/Scan10/scan10_0710.tex
\begin{figure*}[h]
\centering
    \renewcommand{\thesubfigure}{}
     \subfigure[]
     {\includegraphics[width=0.195\textwidth]{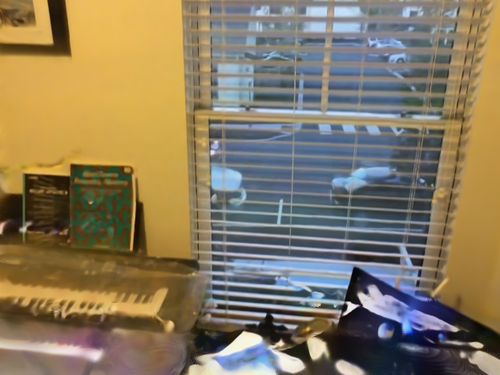}}
     \subfigure[]
     {\includegraphics[width=0.195\textwidth]{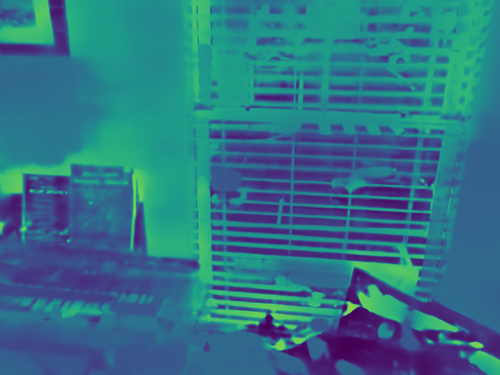}}
     \subfigure[]
     {\includegraphics[width=0.195\textwidth]{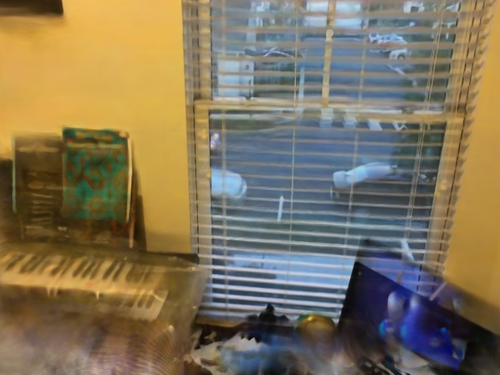}}
     \subfigure[]
     {\includegraphics[width=0.195\textwidth]{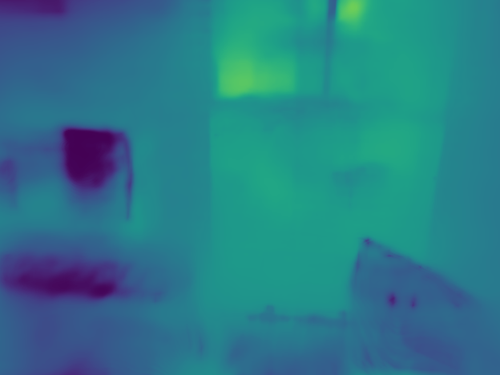}}
     \subfigure[]
     {\includegraphics[width=0.195\textwidth]{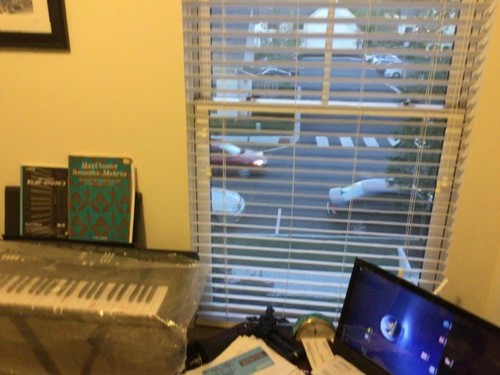}}
    \hfill\\\vspace{-20.5pt}    
     \subfigure[]
     {\includegraphics[width=0.195\textwidth]{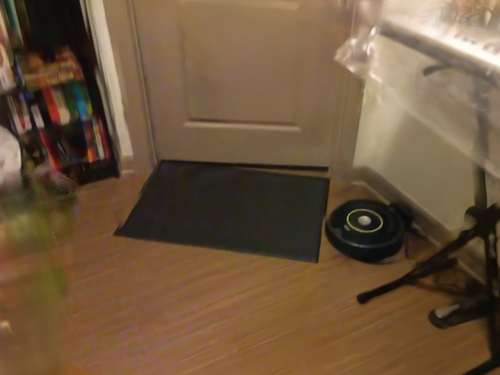}}
     \subfigure[]
     {\includegraphics[width=0.195\textwidth]{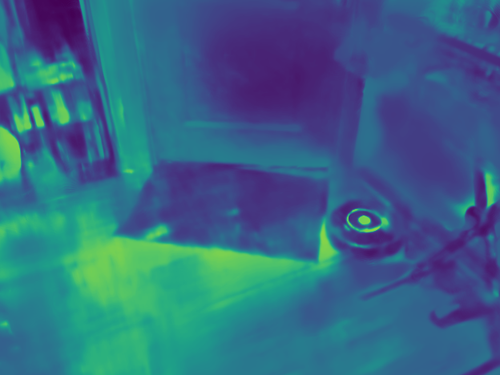}}
     \subfigure[]
     {\includegraphics[width=0.195\textwidth]{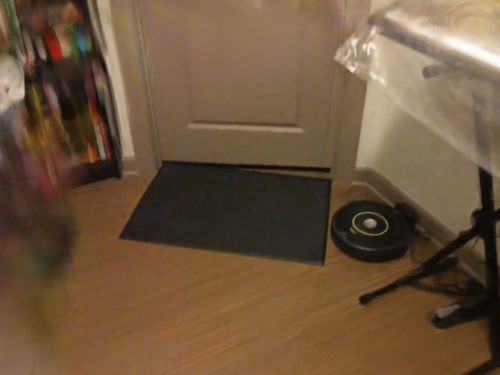}}
     \subfigure[]
     {\includegraphics[width=0.195\textwidth]{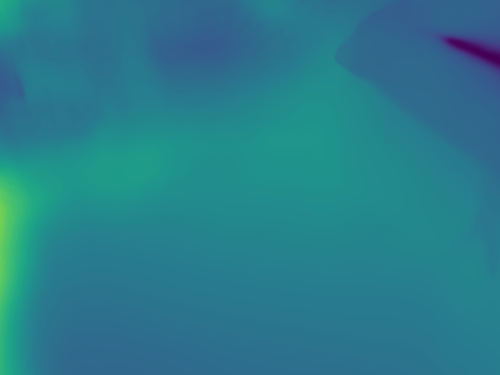}}
     \subfigure[]
     {\includegraphics[width=0.195\textwidth]{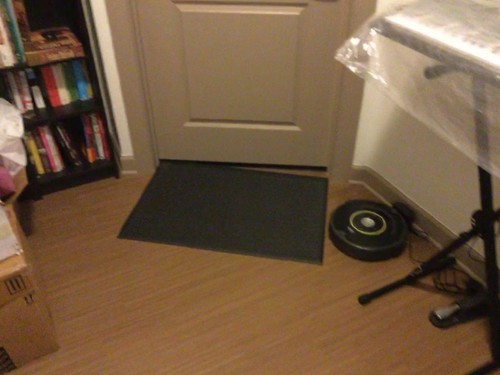}}
     \hfill\\\vspace{-20.5pt} 
     \subfigure[]
     {\includegraphics[width=0.195\textwidth]{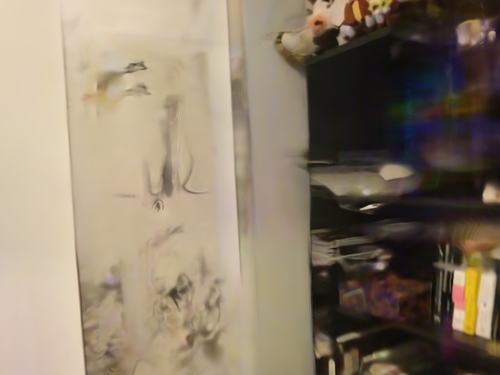}}
     \subfigure[]
     {\includegraphics[width=0.195\textwidth]{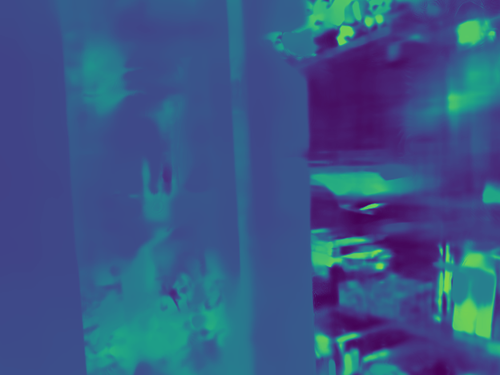}}
     \subfigure[]
     {\includegraphics[width=0.195\textwidth]{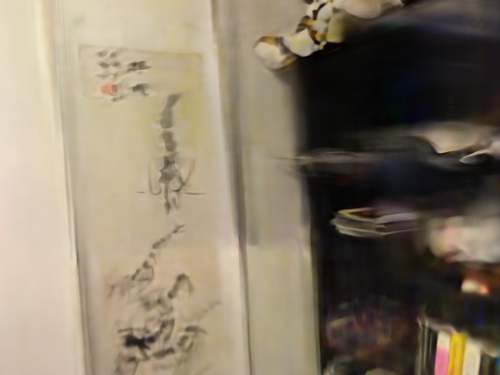}}
     \subfigure[]
     {\includegraphics[width=0.195\textwidth]{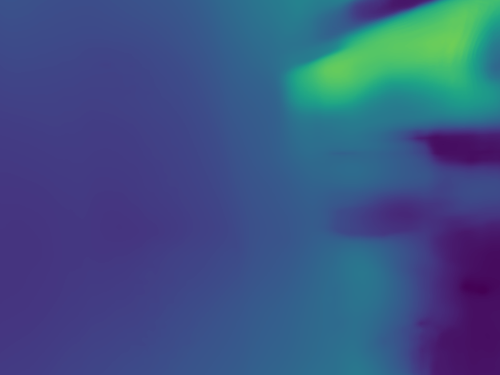}}
     \subfigure[]
     {\includegraphics[width=0.195\textwidth]{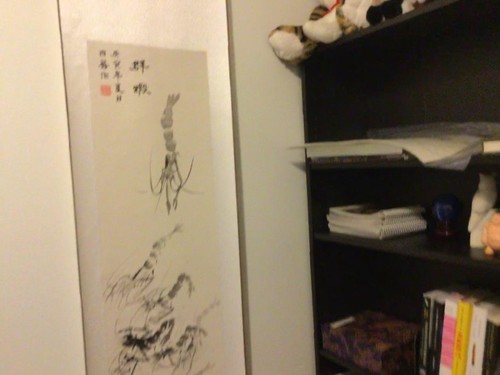}}
     \hfill\\\vspace{-20.5pt} 
     \subfigure[(a) Baseline~\cite{fridovich2023k}]
     {\includegraphics[width=0.195\textwidth]{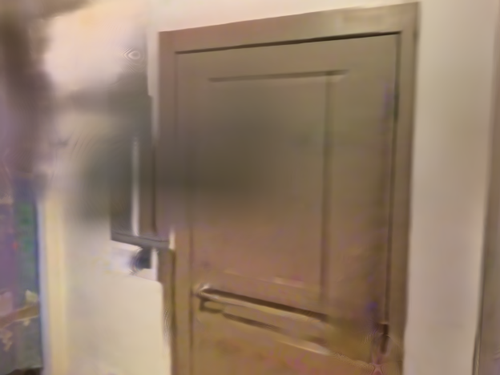}}
     \subfigure[(b) Baseline - Depth]
     {\includegraphics[width=0.195\textwidth]{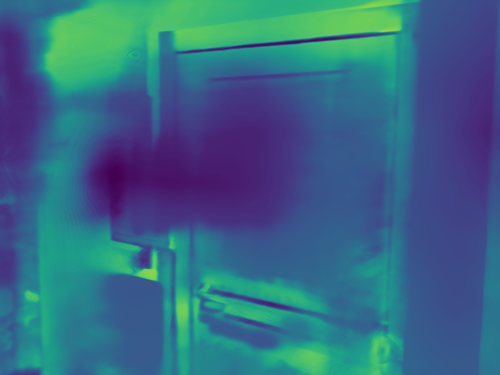}}
     \subfigure[(c) \ours]
     {\includegraphics[width=0.195\textwidth]{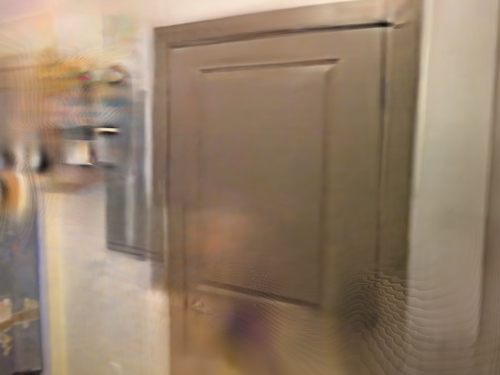}}
     \subfigure[(d) \ours - Depth]
     {\includegraphics[width=0.195\textwidth]{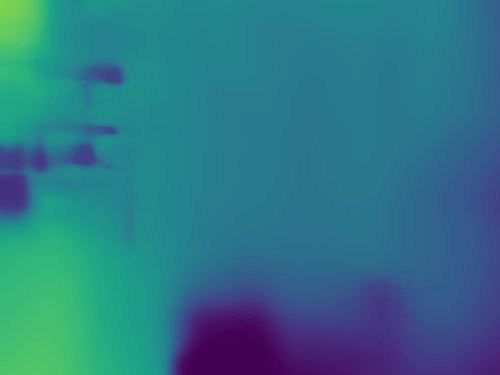}}
     \subfigure[(e) Ground truth]
     {\includegraphics[width=0.195\textwidth]{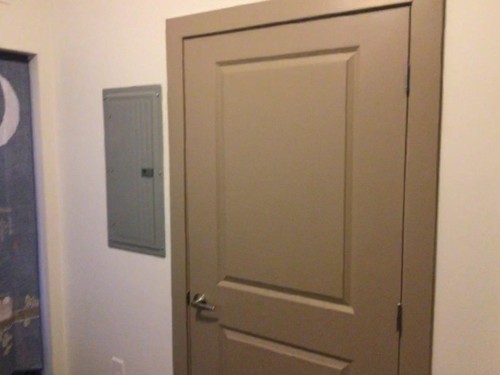}}
    \vspace{-5pt}
    \caption{\textbf{Qualitative results on Scan 0710 of ScanNet~\cite{Dai_2017_CVPR} with 9 - 10 input views}.}
    \label{qual:scan10_0710}
\end{figure*}

%% file: Suppl_Figures/_tex/Scan10/scan10_0758.tex
\begin{figure*}[h]
\centering
    \vspace{-5pt}
    \renewcommand{\thesubfigure}{}
     \subfigure[]
     {\includegraphics[width=0.195\textwidth]{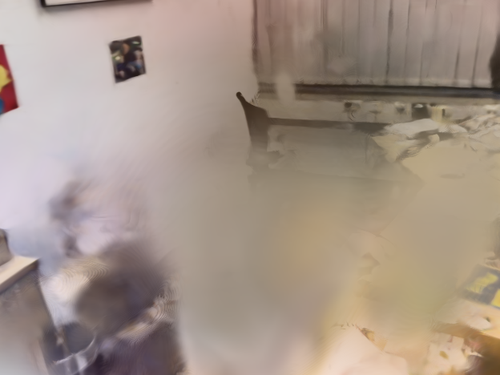}}
     \subfigure[]
     {\includegraphics[width=0.195\textwidth]{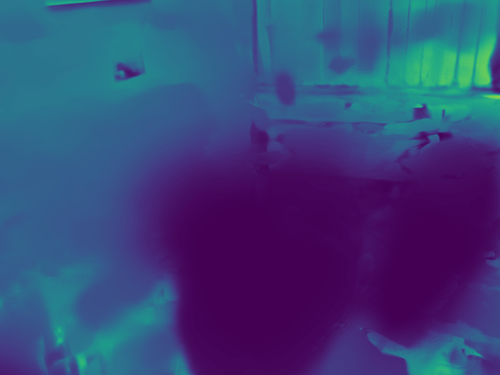}}
     \subfigure[]
     {\includegraphics[width=0.195\textwidth]{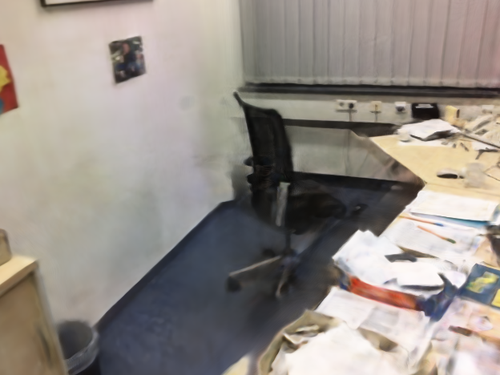}}
     \subfigure[]
     {\includegraphics[width=0.195\textwidth]{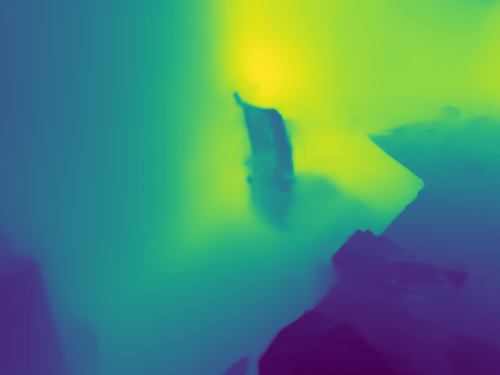}}
     \subfigure[]
     {\includegraphics[width=0.195\textwidth]{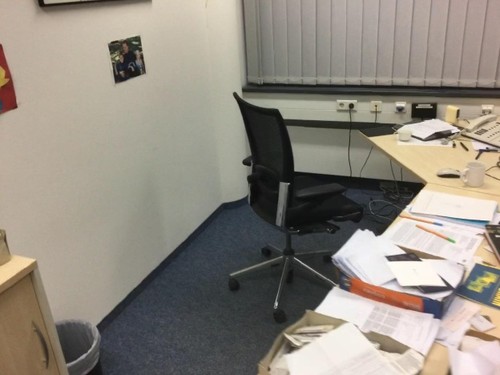}}
     \hfill\\\vspace{-20.5pt}
     \subfigure[]
     {\includegraphics[width=0.195\textwidth]{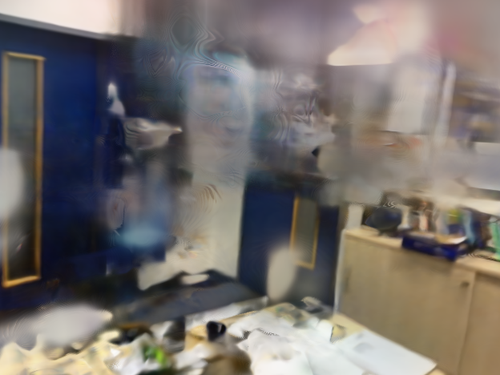}}
     \subfigure[]
     {\includegraphics[width=0.195\textwidth]{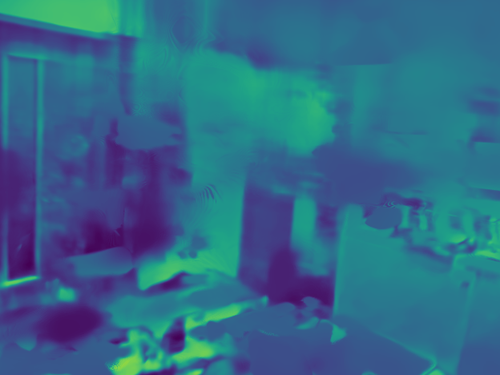}}
     \subfigure[]
     {\includegraphics[width=0.195\textwidth]{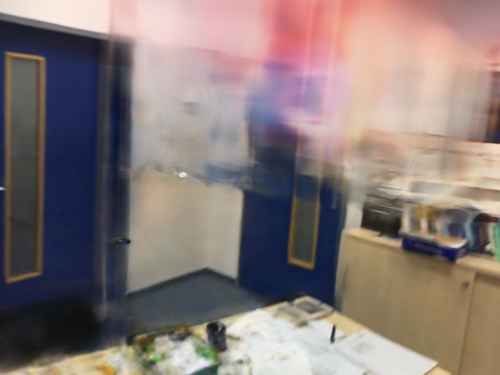}}
     \subfigure[]
     {\includegraphics[width=0.195\textwidth]{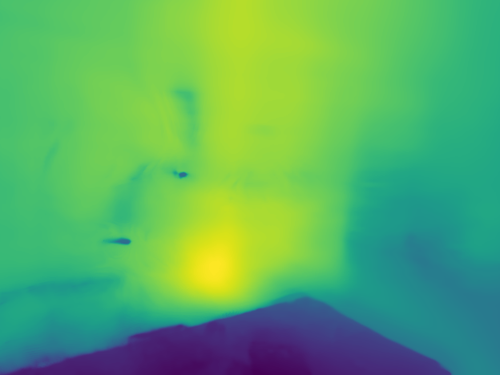}}
     \subfigure[]
     {\includegraphics[width=0.195\textwidth]{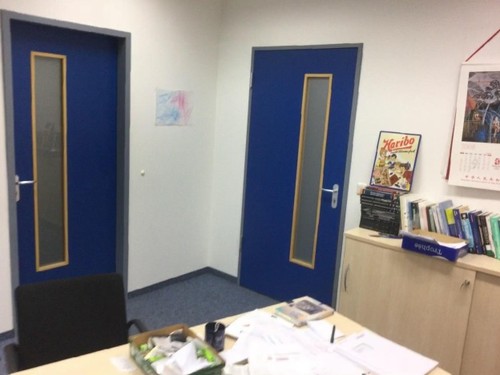}}
     \hfill\\\vspace{-20.5pt}
     \subfigure[]
     {\includegraphics[width=0.195\textwidth]{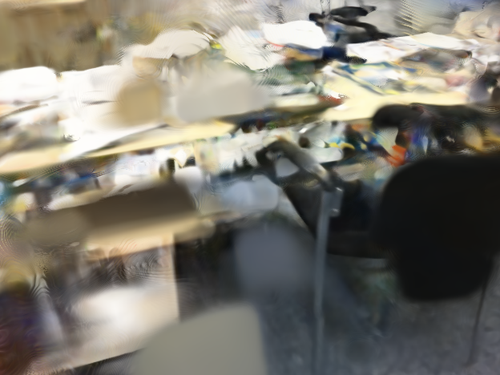}}
     \subfigure[]
     {\includegraphics[width=0.195\textwidth]{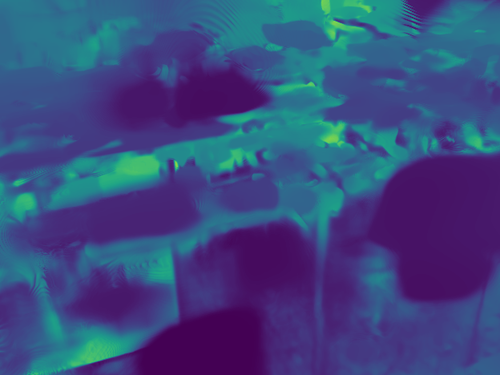}}
     \subfigure[]
     {\includegraphics[width=0.195\textwidth]{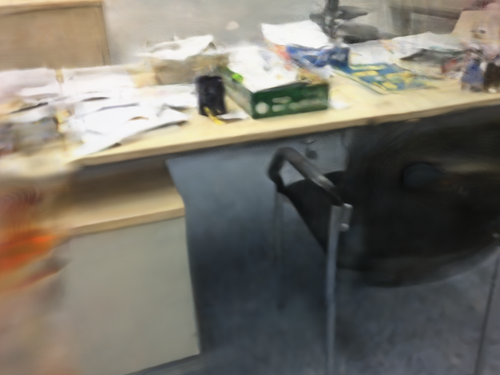}}
     \subfigure[]
     {\includegraphics[width=0.195\textwidth]{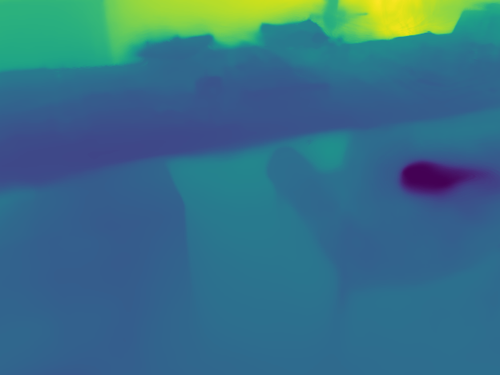}}
     \subfigure[]
     {\includegraphics[width=0.195\textwidth]{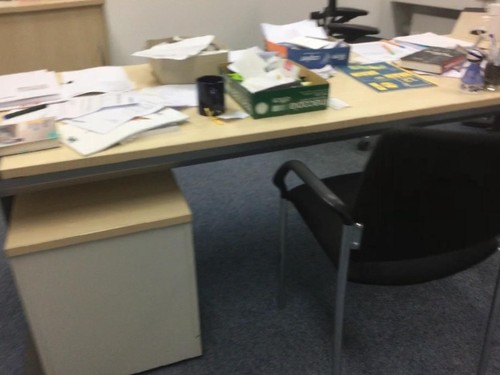}}
     \hfill\\\vspace{-20.5pt} 
     \subfigure[(a) Baseline~\cite{fridovich2023k}]
     {\includegraphics[width=0.195\textwidth]{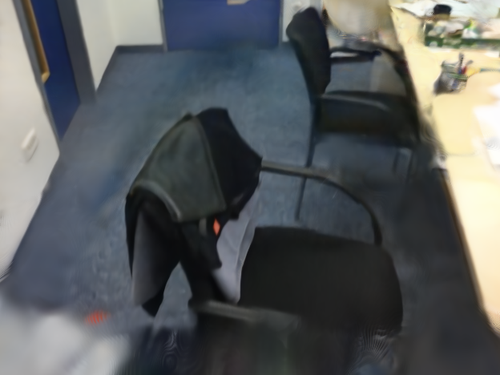}}
     \subfigure[(b) Baseline - Depth]
     {\includegraphics[width=0.195\textwidth]{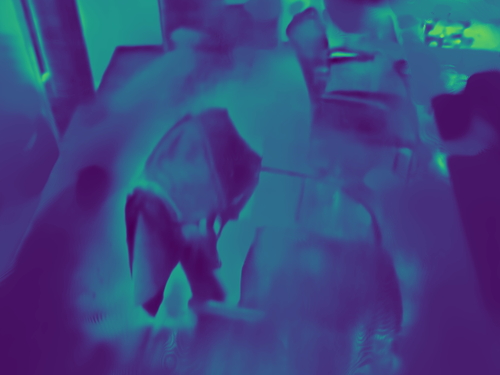}}
     \subfigure[(c) \ours]
     {\includegraphics[width=0.195\textwidth]{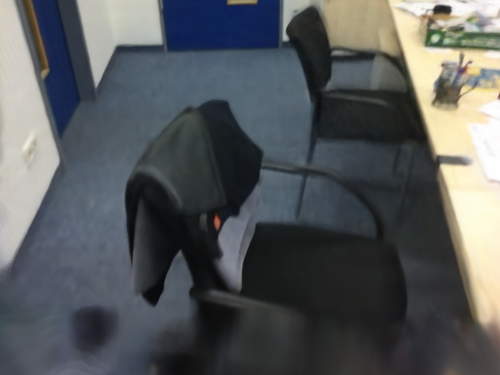}}
     \subfigure[(d) \ours - Depth]
     {\includegraphics[width=0.195\textwidth]{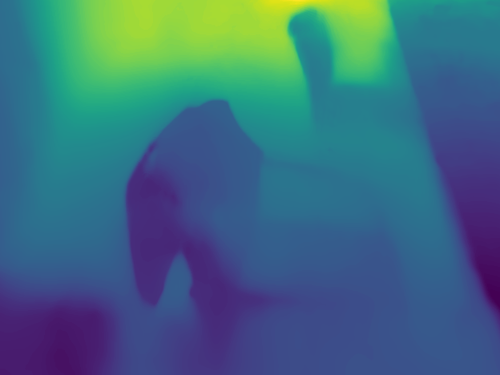}}
     \subfigure[(e) Ground truth]
    {\includegraphics[width=0.195\textwidth]{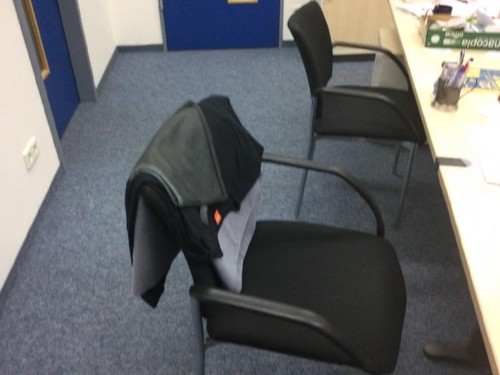}}
    \vspace{-5pt}
    \caption{\textbf{Qualitative results on Scan 0758 of ScanNet~\cite{Dai_2017_CVPR} with 9 - 10 input views}.}
    \label{qual:scan10_0758}
\end{figure*}

%% file: Suppl_Figures/_tex/Scan10/scan10_0781.tex
\begin{figure*}[]
\centering
    \renewcommand{\thesubfigure}{}
    \subfigure[]
    {\includegraphics[width=0.195\textwidth]{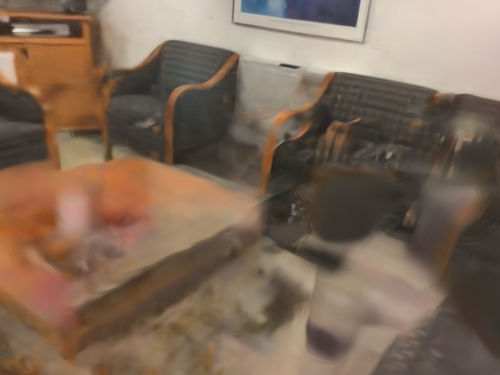}}
    \subfigure[]
    {\includegraphics[width=0.195\textwidth]{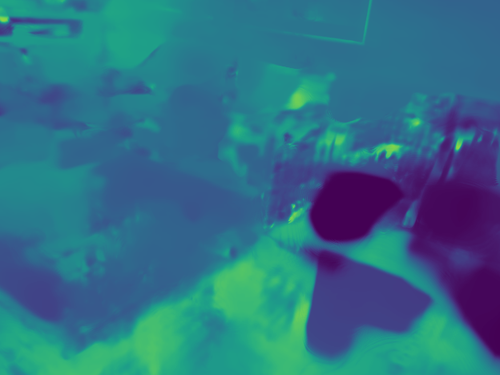}}
    \subfigure[]
    {\includegraphics[width=0.195\textwidth]{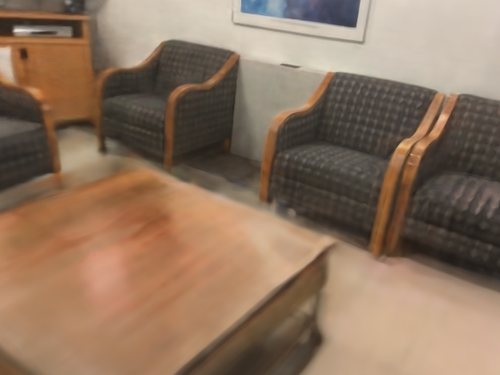}}
    \subfigure[]
    {\includegraphics[width=0.195\textwidth]{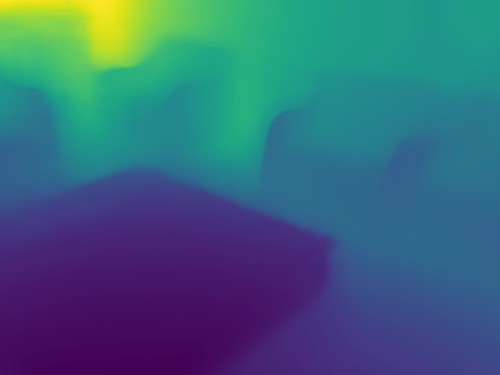}}
    \subfigure[]
    {\includegraphics[width=0.195\textwidth]{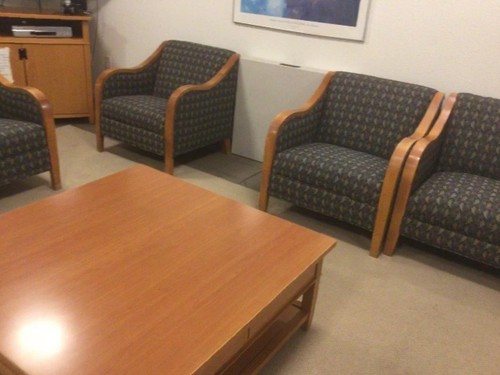}}
    \hfill\\\vspace{-20.5pt}    
    \subfigure[]
    {\includegraphics[width=0.195\textwidth]{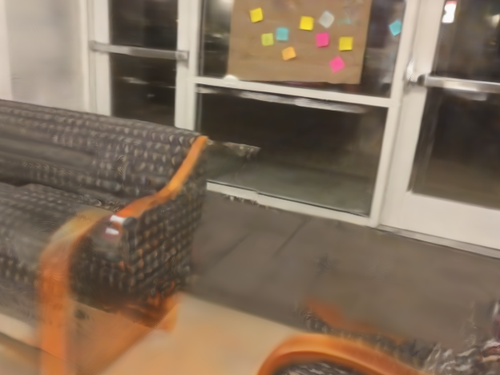}}
    \subfigure[]
    {\includegraphics[width=0.195\textwidth]{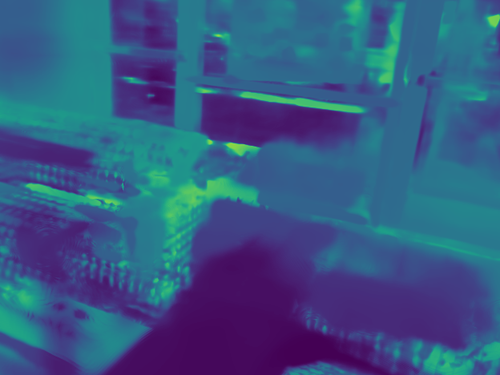}}
    \subfigure[]
    {\includegraphics[width=0.195\textwidth]{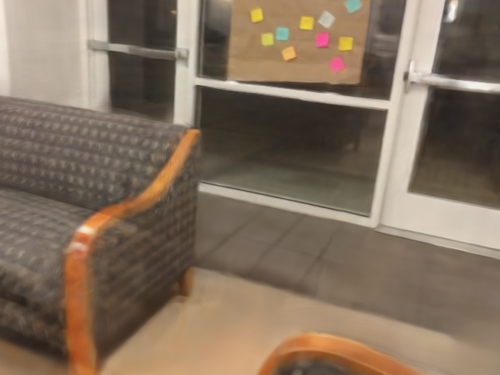}}
    \subfigure[]
    {\includegraphics[width=0.195\textwidth]{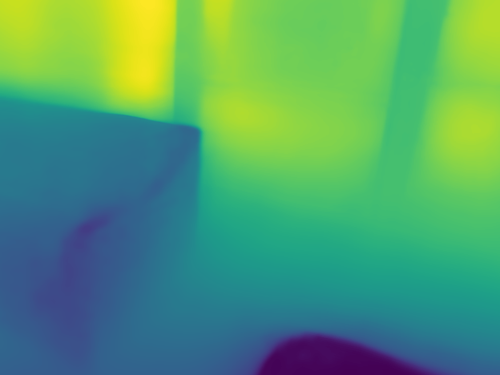}}
    \subfigure[]
    {\includegraphics[width=0.195\textwidth]{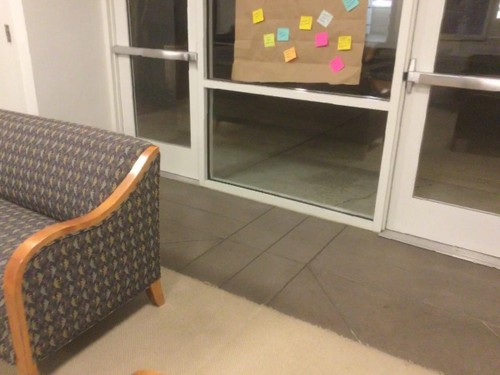}}
    \hfill\\\vspace{-20.5pt}
    \subfigure[]
    {\includegraphics[width=0.195\textwidth]{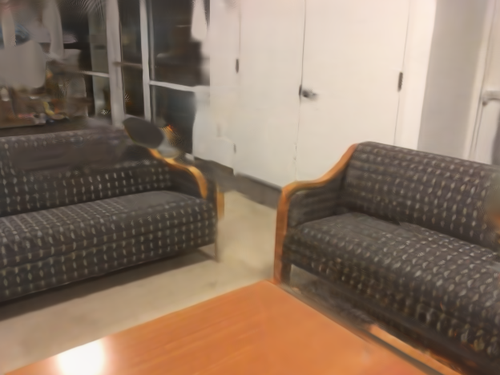}}
    \subfigure[]
    {\includegraphics[width=0.195\textwidth]{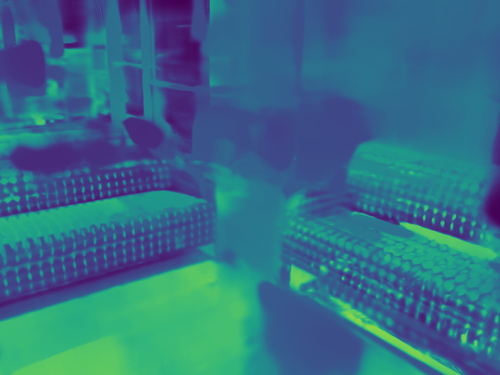}}
    \subfigure[]
    {\includegraphics[width=0.195\textwidth]{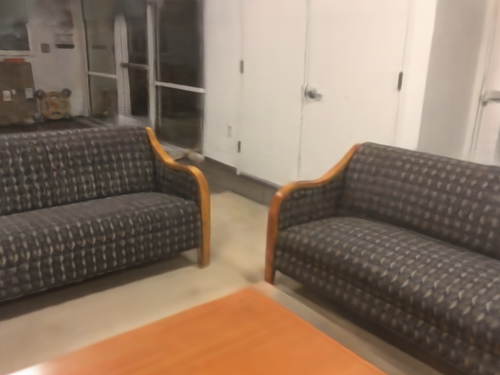}}
    \subfigure[]
    {\includegraphics[width=0.195\textwidth]{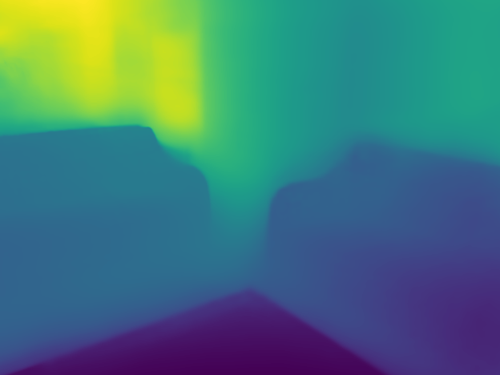}}
    \subfigure[]
    {\includegraphics[width=0.195\textwidth]{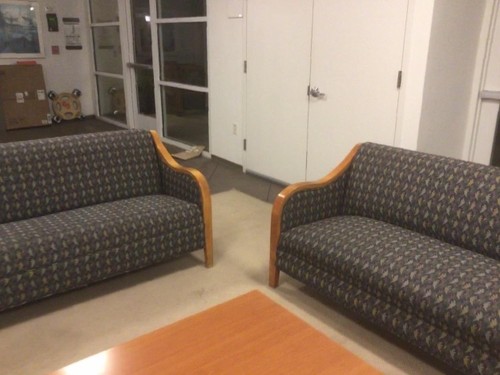}}
    \hfill\\\vspace{-20.5pt}
    \subfigure[(a) Baseline~\cite{fridovich2023k}]
    {\includegraphics[width=0.195\textwidth]{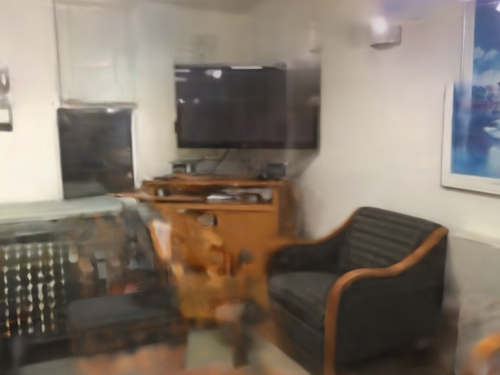}}
    \subfigure[(b) Baseline - Depth]
    {\includegraphics[width=0.195\textwidth]{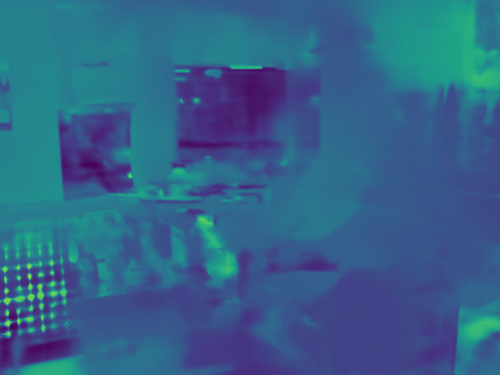}}
    \subfigure[(c) \ours]
    {\includegraphics[width=0.195\textwidth]{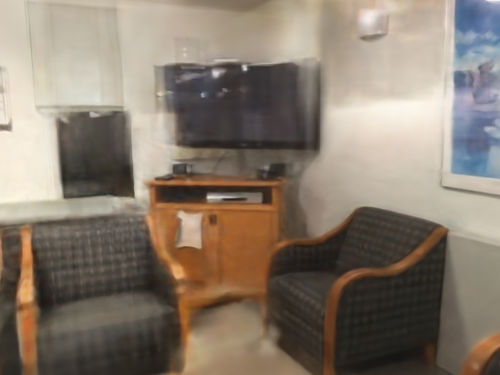}}
    \subfigure[(d) \ours - Depth]
    {\includegraphics[width=0.195\textwidth]{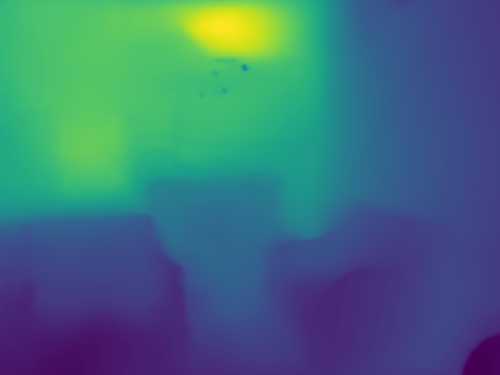}}
    \subfigure[(e) Ground truth]
    {\includegraphics[width=0.195\textwidth]{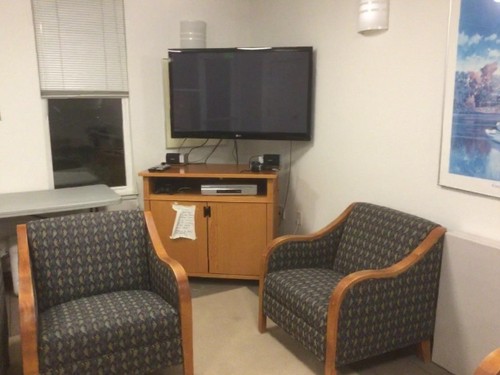}}
    \vspace{-5pt}
    \caption{\textbf{Qualitative results on Scan 0781 of ScanNet~\cite{Dai_2017_CVPR}  with 9 - 10 input views}.}
    \label{qual:scan10_0781}
\end{figure*}

%% file: Suppl_Figures/_tex/Scan20/scan20_0710.tex
\begin{figure*}[]
\centering
    \renewcommand{\thesubfigure}{}
    \subfigure[]
    {\includegraphics[width=0.195\textwidth]{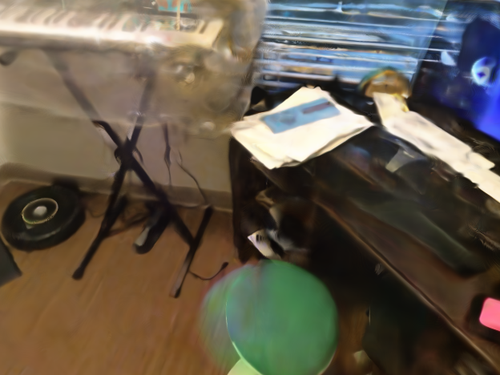}}
    \subfigure[]
    {\includegraphics[width=0.195\textwidth]{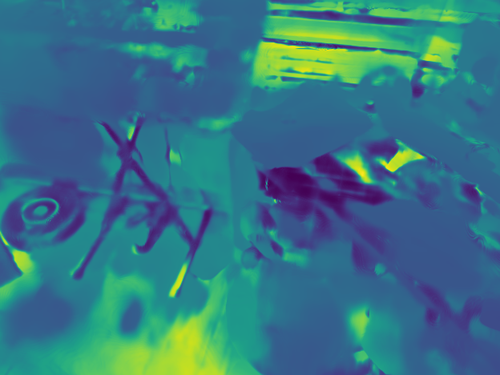}}
    \subfigure[]
    {\includegraphics[width=0.195\textwidth]{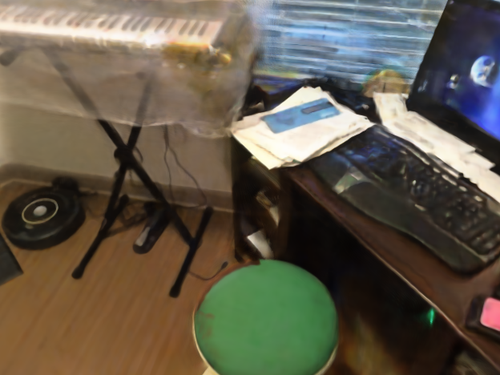}}
    \subfigure[]
    {\includegraphics[width=0.195\textwidth]{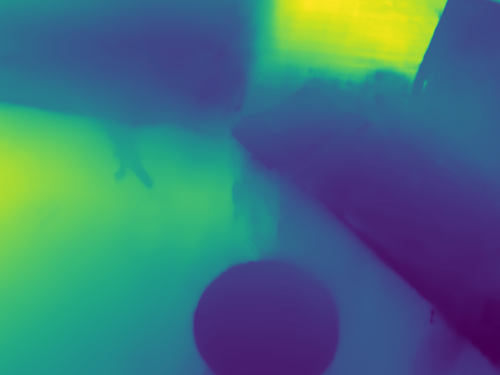}}
    \subfigure[]
    {\includegraphics[width=0.195\textwidth]{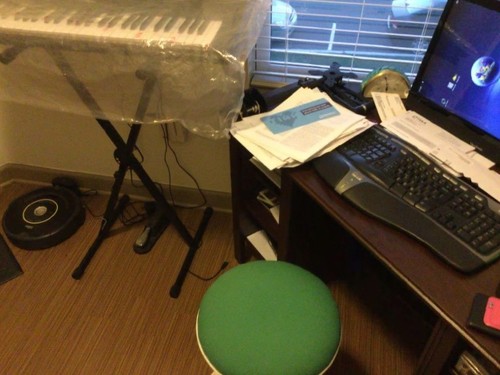}}
    \hfill\\\vspace{-20.5pt} 
    \subfigure[]
    {\includegraphics[width=0.195\textwidth]{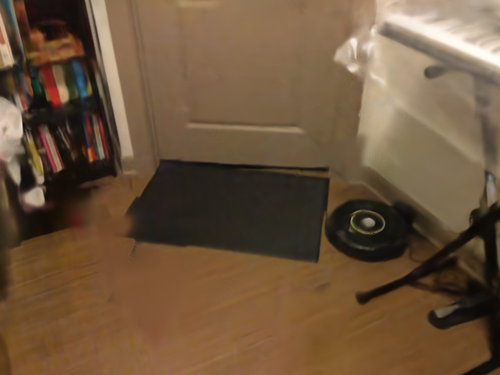}}
    \subfigure[]
    {\includegraphics[width=0.195\textwidth]{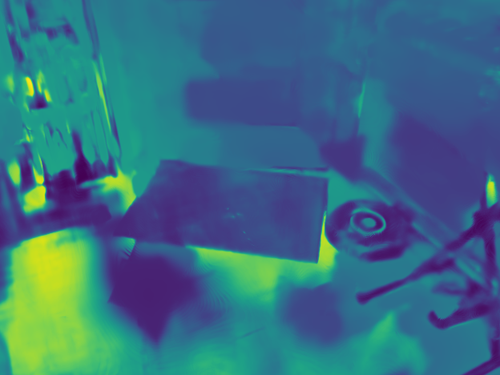}}
    \subfigure[]
    {\includegraphics[width=0.195\textwidth]{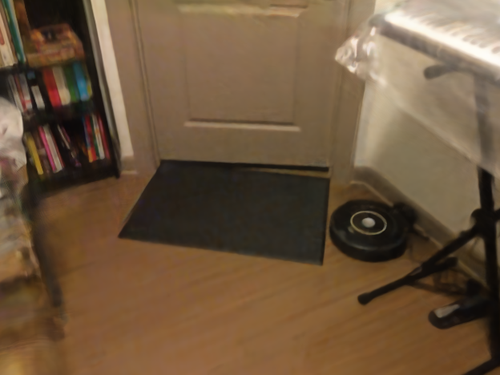}}
    \subfigure[]
    {\includegraphics[width=0.195\textwidth]{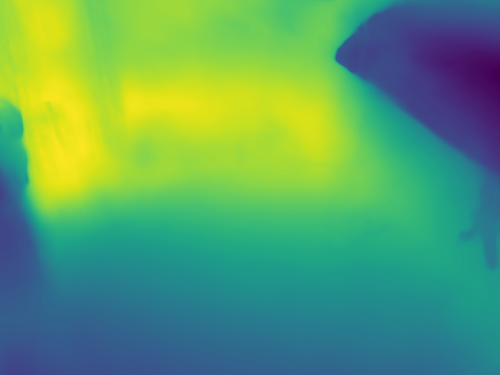}}
    \subfigure[]
    {\includegraphics[width=0.195\textwidth]{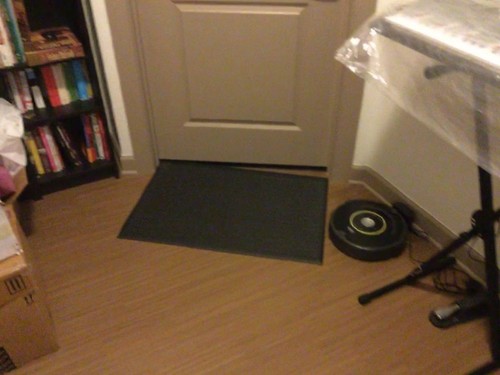}}
    \hfill\\\vspace{-20.5pt} 
    \subfigure[]
    {\includegraphics[width=0.195\textwidth]{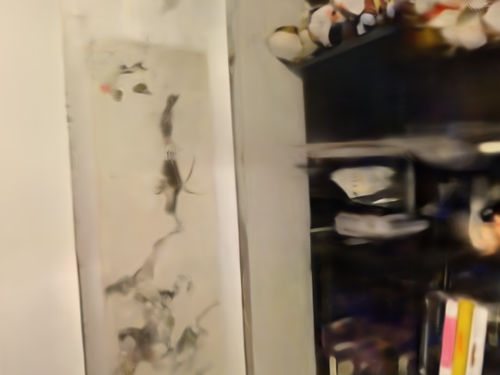}}
    \subfigure[]
    {\includegraphics[width=0.195\textwidth]{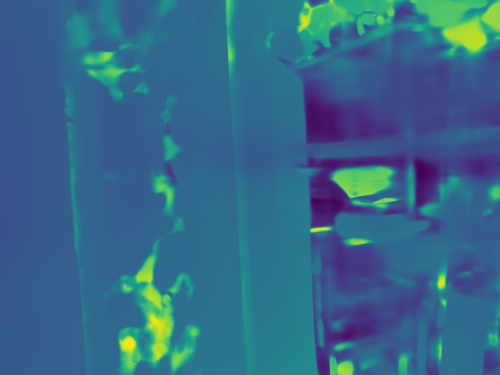}}
    \subfigure[]
    {\includegraphics[width=0.195\textwidth]{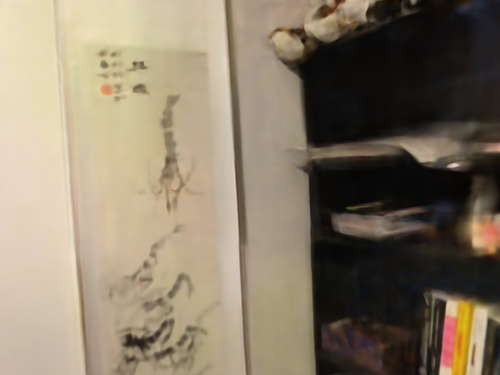}}
    \subfigure[]
    {\includegraphics[width=0.195\textwidth]{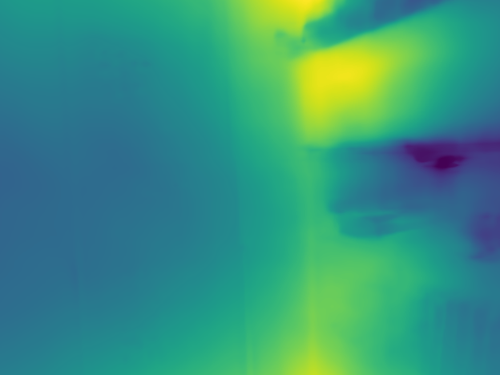}}
    \subfigure[]
    {\includegraphics[width=0.195\textwidth]{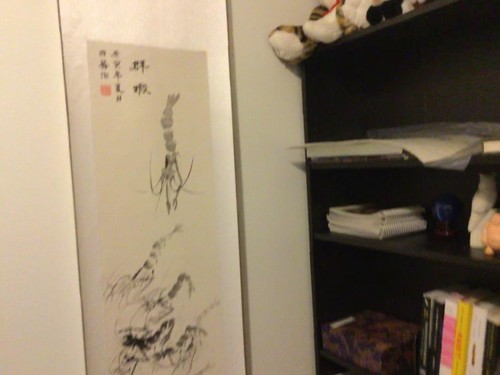}}
    \hfill\\\vspace{-20.5pt} 
    \subfigure[(a) Baseline~\cite{fridovich2023k}]
    {\includegraphics[width=0.195\textwidth]{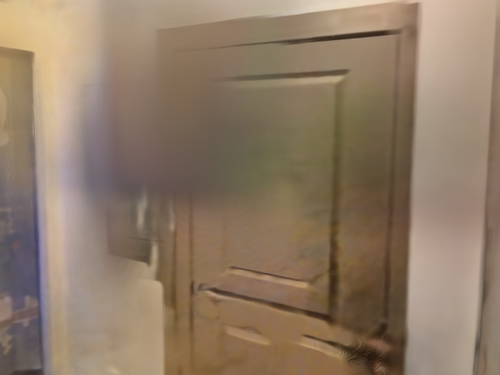}}
    \subfigure[(b) Baseline - Depth]
    {\includegraphics[width=0.195\textwidth]{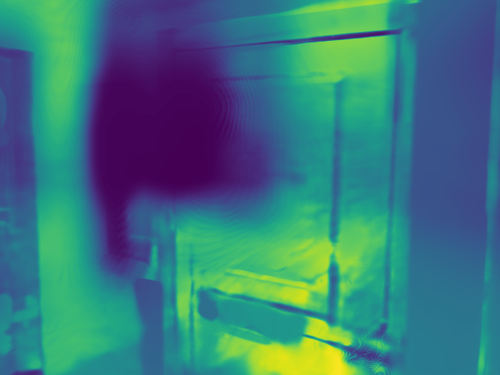}}
    \subfigure[(c) \ours]
    {\includegraphics[width=0.195\textwidth]{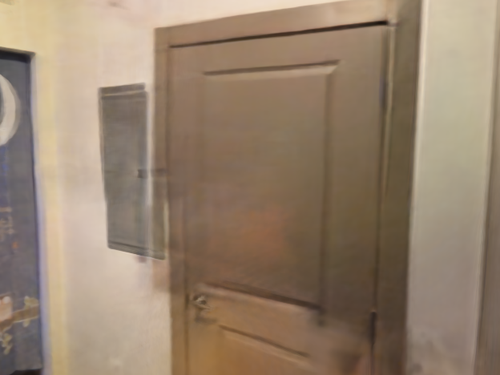}}
    \subfigure[(d) \ours - Depth]
    {\includegraphics[width=0.195\textwidth]{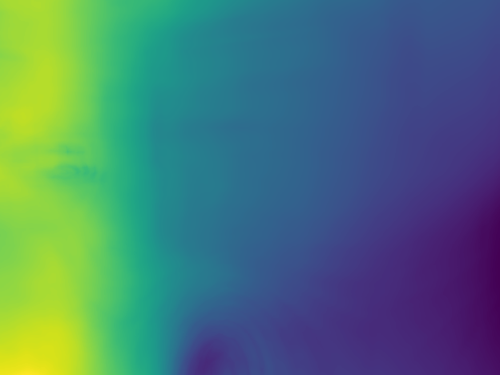}}
    \subfigure[(e) Ground truth]
    {\includegraphics[width=0.195\textwidth]{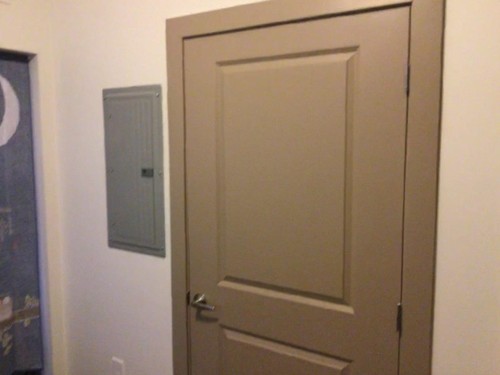}}
    \vspace{-5pt}
    \caption{\textbf{Qualitative results on Scan 0710 of ScanNet~\cite{Dai_2017_CVPR}  with 18 - 20 input views}.}
    \label{qual:scan20_0710}
\end{figure*}

%% file: Suppl_Figures/_tex/Scan20/scan20_0758.tex
\begin{figure*}[]
\centering
    \renewcommand{\thesubfigure}{}
    \subfigure[]
    {\includegraphics[width=0.195\textwidth]{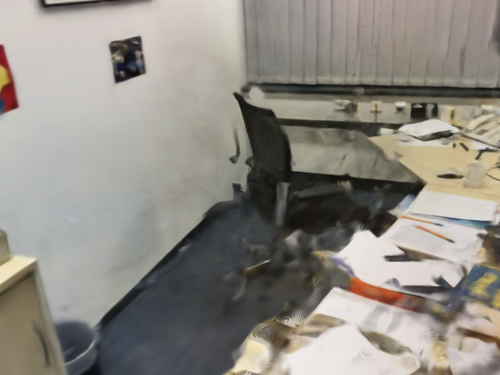}}
    \subfigure[]
    {\includegraphics[width=0.195\textwidth]{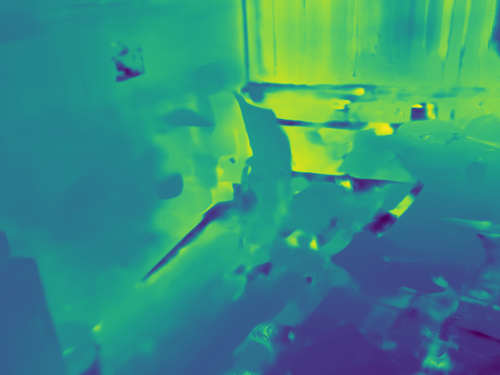}}
    \subfigure[]
    {\includegraphics[width=0.195\textwidth]{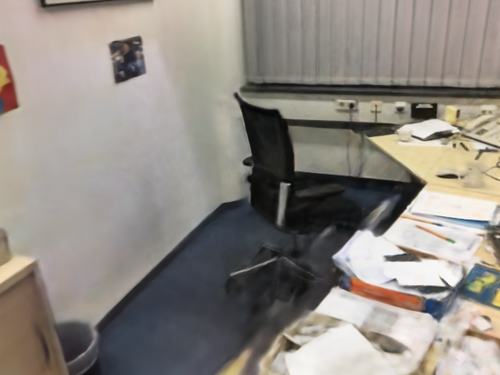}}
    \subfigure[]
    {\includegraphics[width=0.195\textwidth]{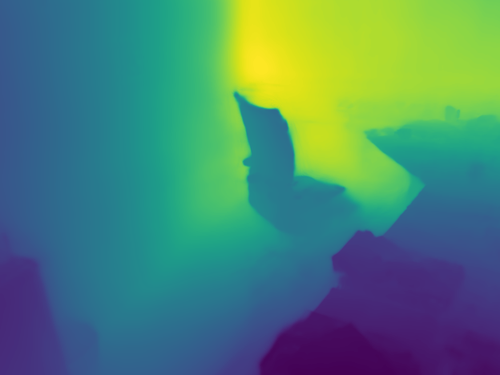}}
    \subfigure[]
    {\includegraphics[width=0.195\textwidth]{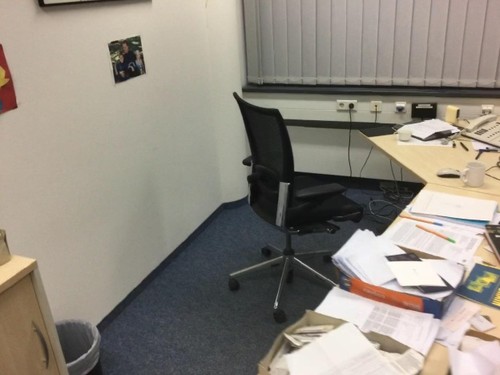}}
    \hfill\\\vspace{-20.5pt}
    \subfigure[]
    {\includegraphics[width=0.195\textwidth]{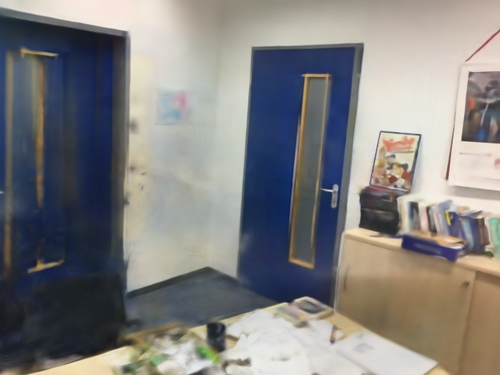}}
    \subfigure[]
    {\includegraphics[width=0.195\textwidth]{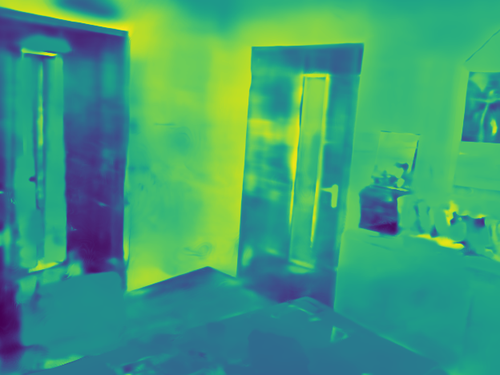}}
    \subfigure[]
    {\includegraphics[width=0.195\textwidth]{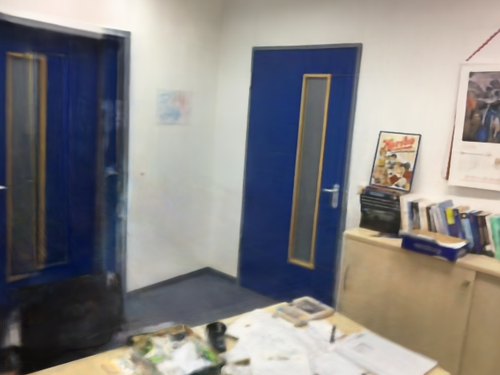}}
    \subfigure[]
    {\includegraphics[width=0.195\textwidth]{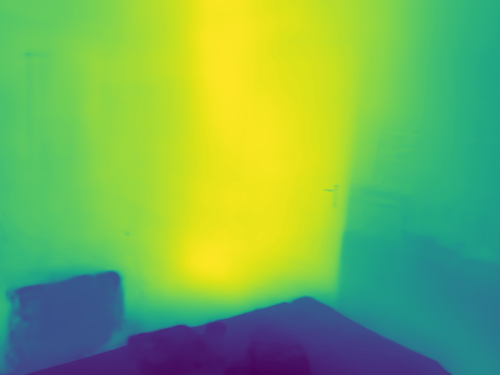}}
    \subfigure[]
    {\includegraphics[width=0.195\textwidth]{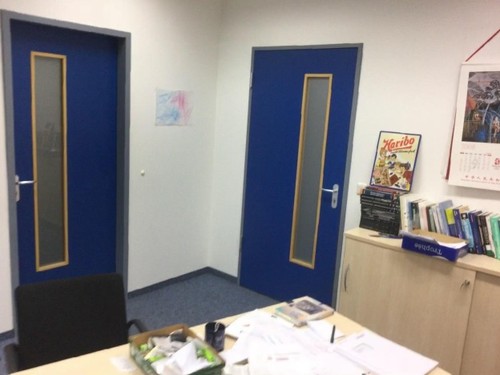}}
    \hfill\\\vspace{-20.5pt}
    \subfigure[]
    {\includegraphics[width=0.195\textwidth]{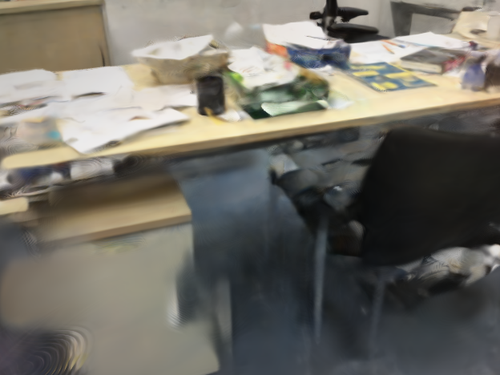}}
    \subfigure[]
    {\includegraphics[width=0.195\textwidth]{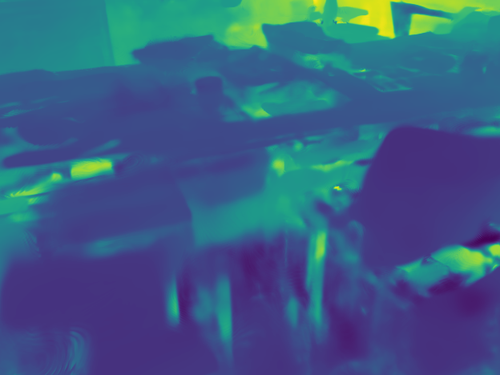}}
    \subfigure[]
    {\includegraphics[width=0.195\textwidth]{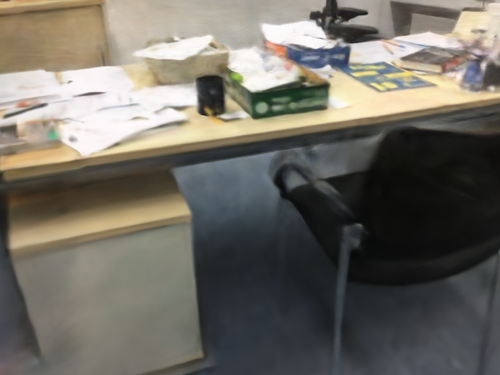}}
    \subfigure[]
    {\includegraphics[width=0.195\textwidth]{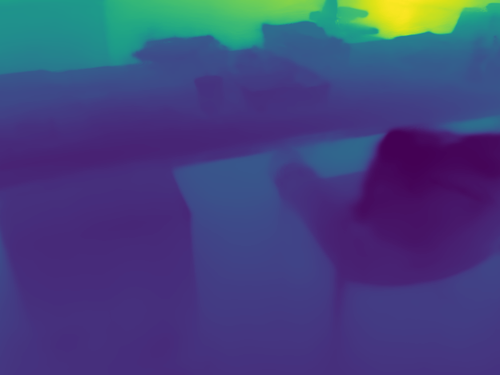}}
    \subfigure[]
    {\includegraphics[width=0.195\textwidth]{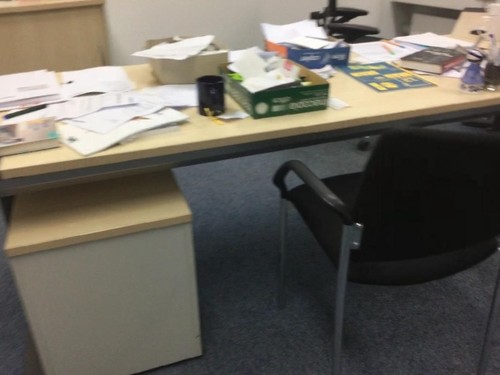}}
    \hfill\\\vspace{-20.5pt} 
    \subfigure[(a) Baseline~\cite{fridovich2023k}]
    {\includegraphics[width=0.195\textwidth]{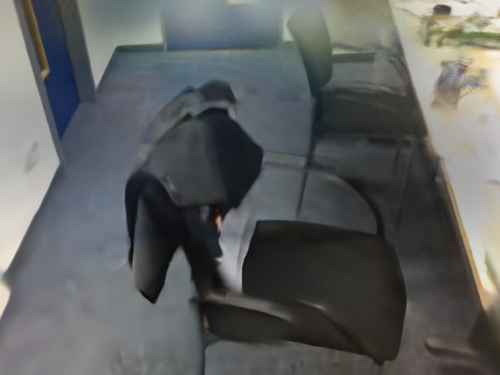}}
    \subfigure[(b) Baseline - Depth]
    {\includegraphics[width=0.195\textwidth]{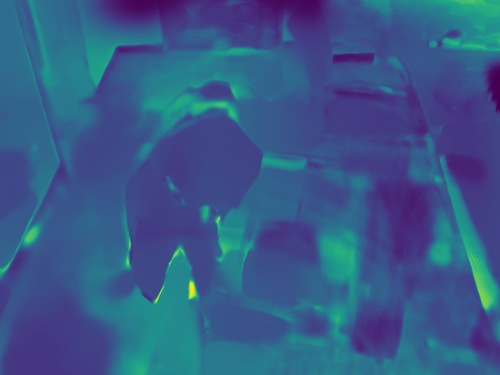}}
    \subfigure[(c) \ours]
    {\includegraphics[width=0.195\textwidth]{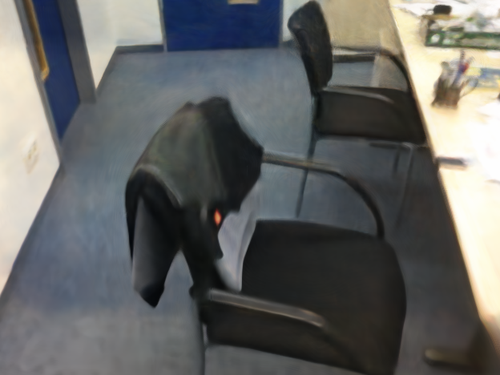}}
    \subfigure[(d) \ours - Depth]
    {\includegraphics[width=0.195\textwidth]{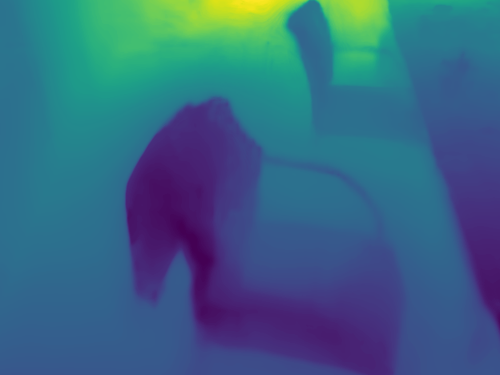}}
    \subfigure[(e) Ground truth]
    {\includegraphics[width=0.195\textwidth]{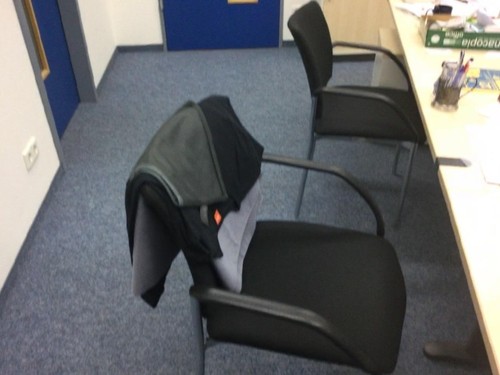}}\\
    \vspace{-5pt}
    \caption{\textbf{Qualitative results on Scan 0758 of ScanNet~\cite{Dai_2017_CVPR} with 18 - 20 input views}.}
    \label{qual:scan20_0758}
\end{figure*}

%% file: Suppl_Figures/_tex/Scan20/scan20_0781.tex
\begin{figure*}[]
\centering
    \renewcommand{\thesubfigure}{}
    \subfigure[]
    {\includegraphics[width=0.195\textwidth]{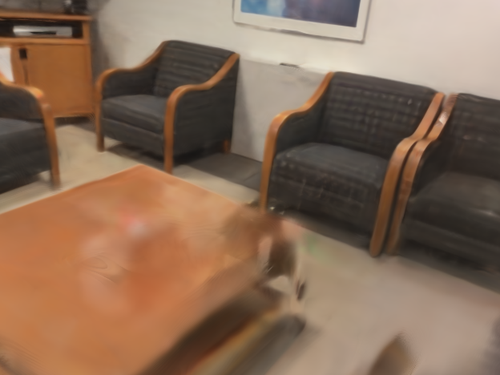}}
    \subfigure[]
    {\includegraphics[width=0.195\textwidth]{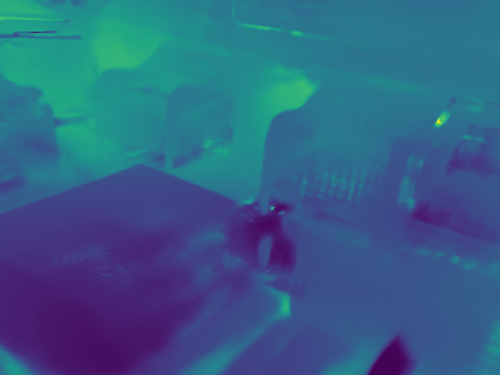}}
    \subfigure[]
    {\includegraphics[width=0.195\textwidth]{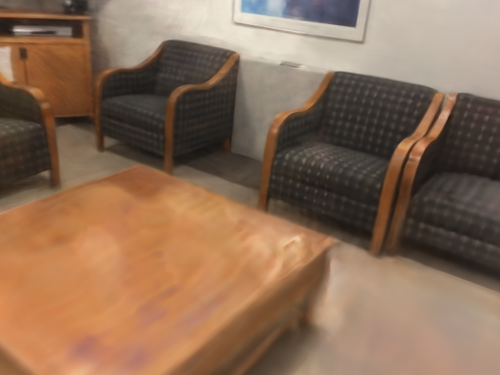}}
    \subfigure[]
    {\includegraphics[width=0.195\textwidth]{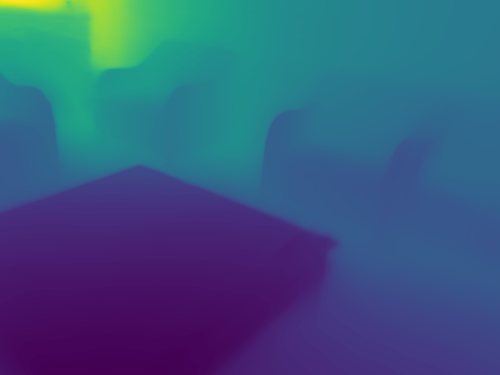}}
    \subfigure[]
    {\includegraphics[width=0.195\textwidth]{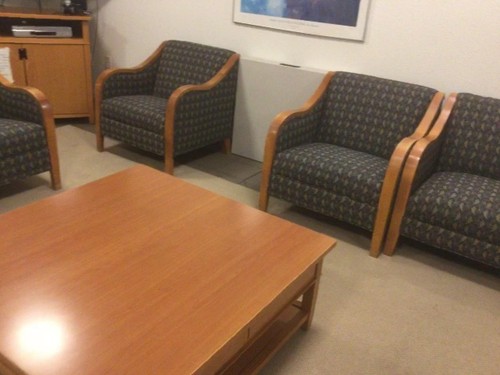}}
    \hfill\\\vspace{-20.5pt} 
    \subfigure[]
    {\includegraphics[width=0.195\textwidth]{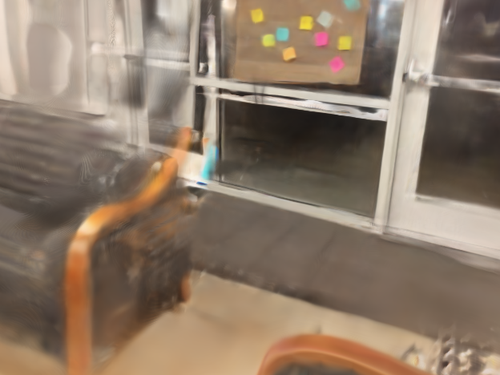}}
    \subfigure[]
    {\includegraphics[width=0.195\textwidth]{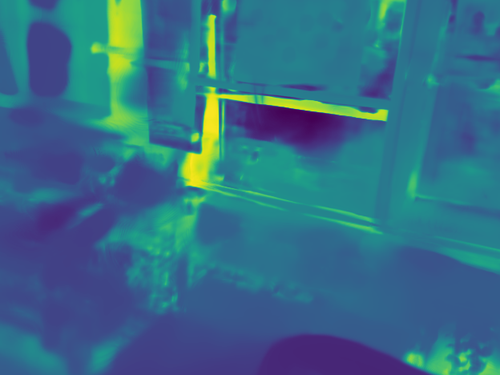}}
    \subfigure[]
    {\includegraphics[width=0.195\textwidth]{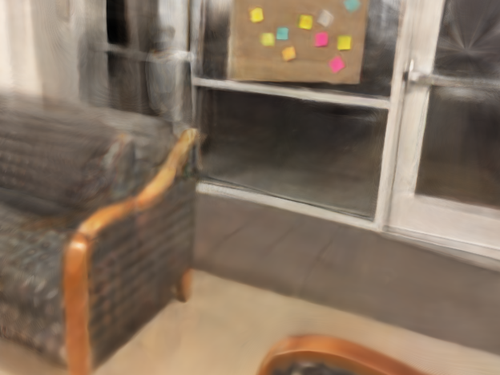}}
    \subfigure[]
    {\includegraphics[width=0.195\textwidth]{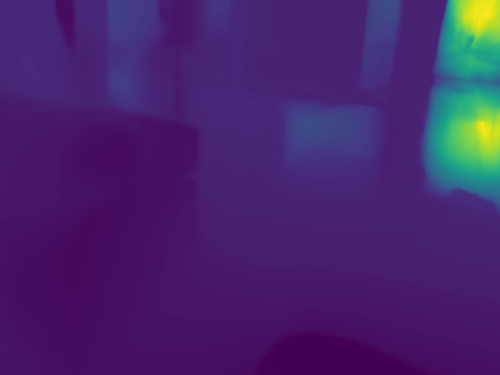}}
    \subfigure[]
    {\includegraphics[width=0.195\textwidth]{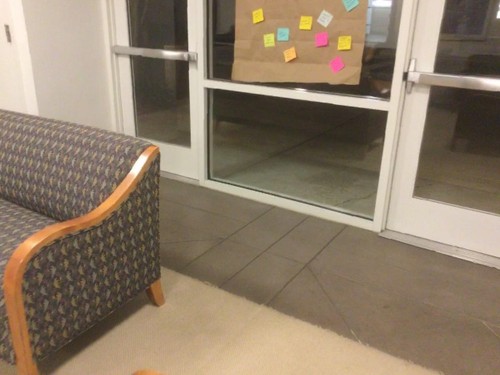}}
    \hfill\\\vspace{-20.5pt} 
    \subfigure[]
    {\includegraphics[width=0.195\textwidth]{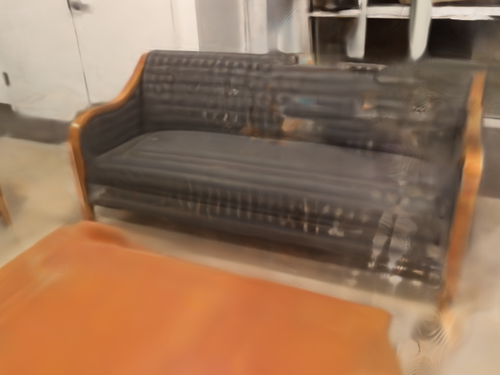}}
    \subfigure[]
    {\includegraphics[width=0.195\textwidth]{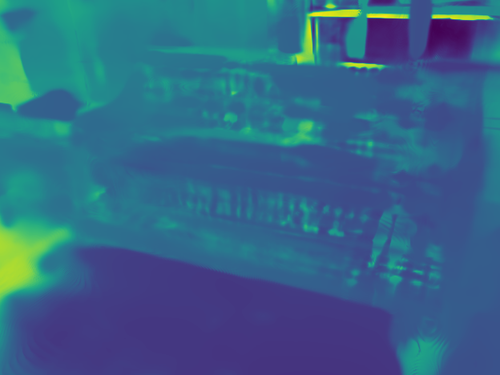}}
    \subfigure[]
    {\includegraphics[width=0.195\textwidth]{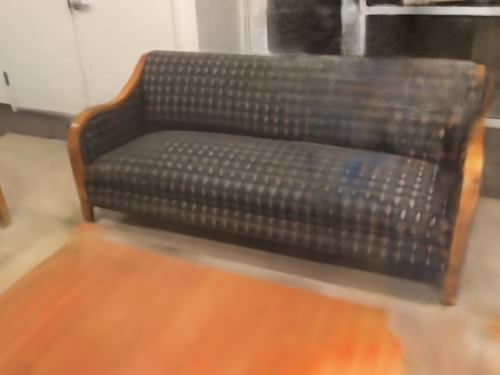}}
    \subfigure[]
    {\includegraphics[width=0.195\textwidth]{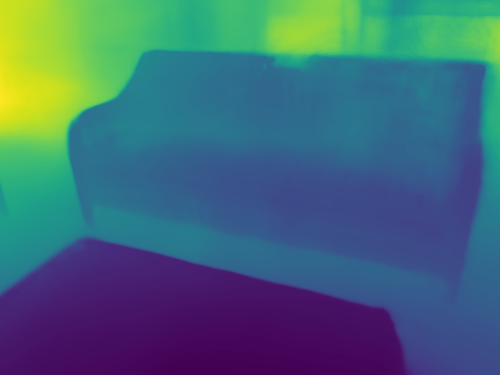}}
    \subfigure[]
    {\includegraphics[width=0.195\textwidth]{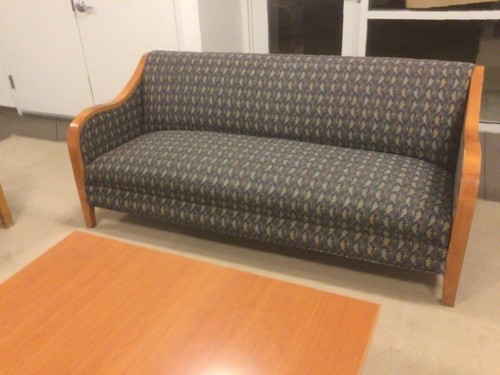}}
    \hfill\\\vspace{-20.5pt}
    \subfigure[(a) Baseline~\cite{fridovich2023k}]
    {\includegraphics[width=0.195\textwidth]{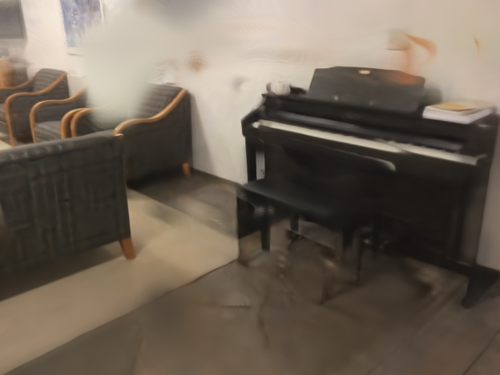}}
    \subfigure[(b) Baseline - Depth]
    {\includegraphics[width=0.195\textwidth]{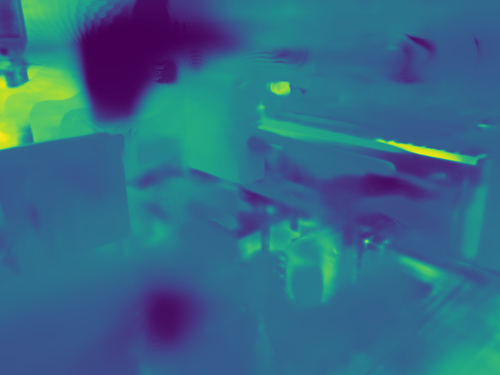}}
    \subfigure[(c) \ours]
    {\includegraphics[width=0.195\textwidth]{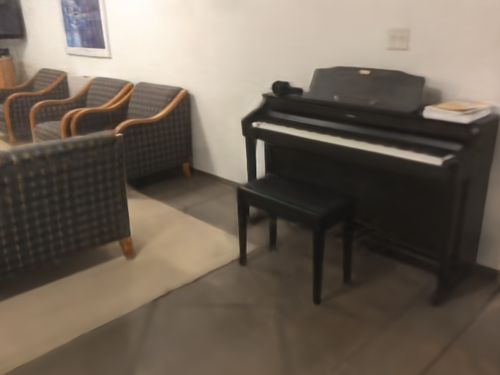}}
    \subfigure[(d) \ours - Depth]
    {\includegraphics[width=0.195\textwidth]{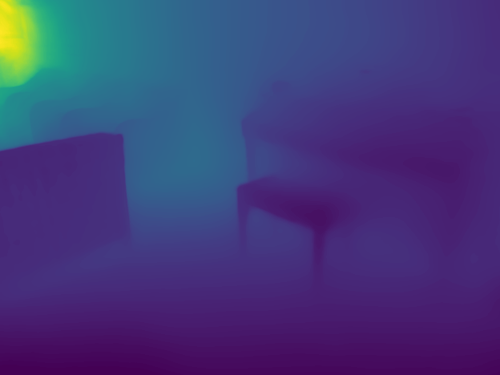}}
    \subfigure[(e) Ground truth]
    {\includegraphics[width=0.195\textwidth]{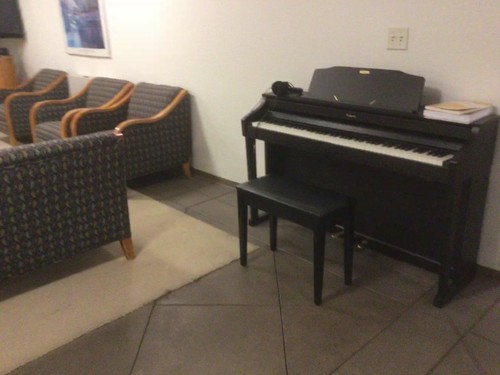}}
    \vspace{-5pt}
    \caption{\textbf{Qualitative results on Scan 0781 of ScanNet~\cite{Dai_2017_CVPR} with 18 - 20 input views}.}
    \label{qual:scan20_0781}
\end{figure*}

%% file: Suppl_Figures/_tex/TnT/TnT_truck.tex
\begin{figure*}[]
\centering
    \renewcommand{\thesubfigure}{}
    \subfigure[(a) Baseline~\cite{fridovich2023k}]
    {\includegraphics[width=0.193\textwidth]{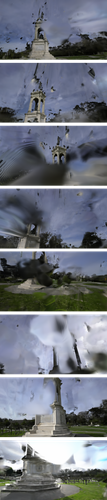}}
    \subfigure[(b) Baseline - Depth]
    {\includegraphics[width=0.193\textwidth]{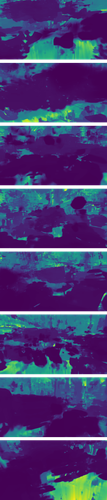}}
    \subfigure[(c) \ours]
    {\includegraphics[width=0.1915\textwidth]{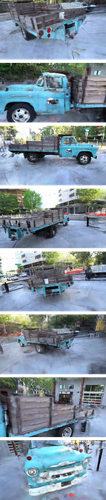}}
    \subfigure[(d) \ours - Depth]
    {\includegraphics[width=0.193\textwidth]{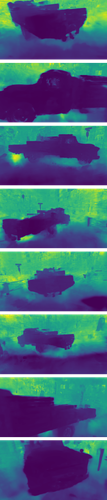}}
    \subfigure[(e) Ground truth]
    {\includegraphics[width=0.193\textwidth]{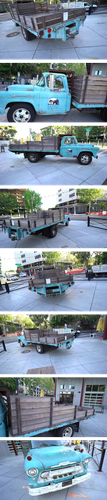}}\\
    \caption{\textbf{Qualitative results on truck scene of Tanks and Temples~\cite{knapitsch2017tanks} with 10 input views}.}
    \vspace*{\fill}%
    \label{qual:truck}
\end{figure*}
\clearpage%

%% file: Suppl_Figures/_tex/TnT/TnT_francis.tex
\begin{figure*}[]
\centering
    \renewcommand{\thesubfigure}{}
    \subfigure[(a) Baseline~\cite{fridovich2023k}]
    {\includegraphics[width=0.193\textwidth]{Suppl_Figures/tnt/francis/base.png}}
    \subfigure[(b) Baseline - Depth]
    {\includegraphics[width=0.193\textwidth]{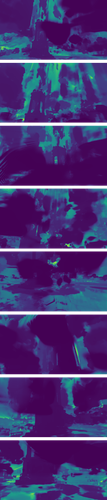}}
    \subfigure[(c) \ours]
    {\includegraphics[width=0.193\textwidth]{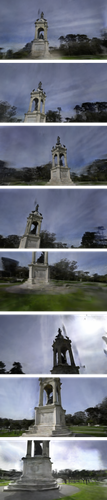}}
    \subfigure[(d) \ours - Depth]
    {\includegraphics[width=0.193\textwidth]{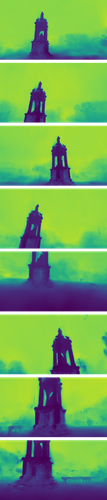}}
    \subfigure[(e) Ground truth]
    {\includegraphics[width=0.193\textwidth]{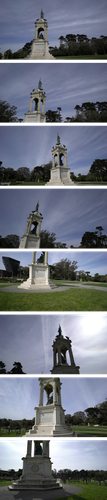}}\\
    \caption{\textbf{Qualitative results on francis scene of Tanks and Temples~\cite{knapitsch2017tanks} with 10 input views}.}
    \vspace*{\fill}%
    \label{qual:francis}
\end{figure*}
\clearpage%

%% file: Suppl_Figures/_tex/TnT/TnT_lighthouse.tex
\begin{figure*}[]
\vspace*{\fill}%
\centering
    \renewcommand{\thesubfigure}{}
    \subfigure[(a) Baseline~\cite{fridovich2023k}]
    {\includegraphics[width=0.195\textwidth]{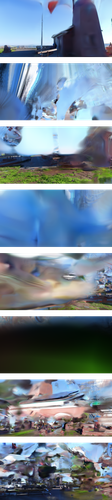}}
    \subfigure[(b) Baseline - Depth]
    {\includegraphics[width=0.195\textwidth]{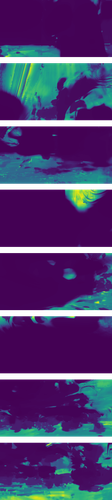}}
    \subfigure[(c) \ours]
    {\includegraphics[width=0.195\textwidth]{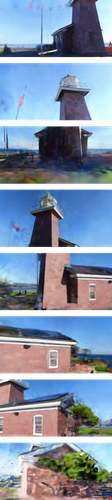}}
    \subfigure[(d) \ours - Depth]
    {\includegraphics[width=0.195\textwidth]{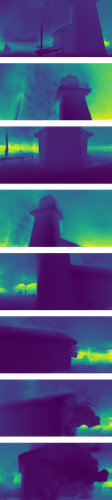}}
    \subfigure[(e) Ground truth]
    {\includegraphics[width=0.195\textwidth]{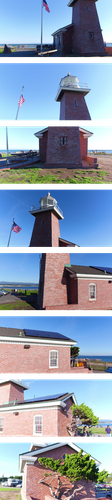}}\\
    \caption{\textbf{Qualitative results on lighthouse scene of Tanks and Temples~\cite{knapitsch2017tanks} with 10 input views}.}
    \vspace*{\fill}%
    \label{qual:lighthouse}
\end{figure*}
\clearpage%

%% file: Suppl_Figures/_tex/TnT/TnT_ignatius.tex
\begin{figure*}[]
\vspace*{\fill}%
\centering
    \renewcommand{\thesubfigure}{}
    \subfigure[]
    {\includegraphics[width=0.195\textwidth]{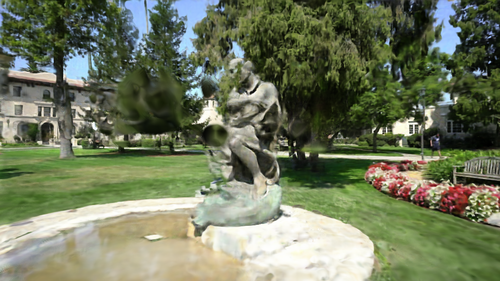}}
    \subfigure[]
    {\includegraphics[width=0.195\textwidth]{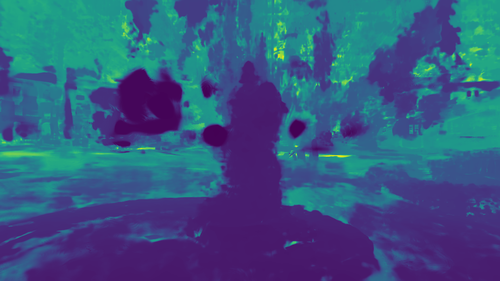}}
    \subfigure[]
    {\includegraphics[width=0.195\textwidth]{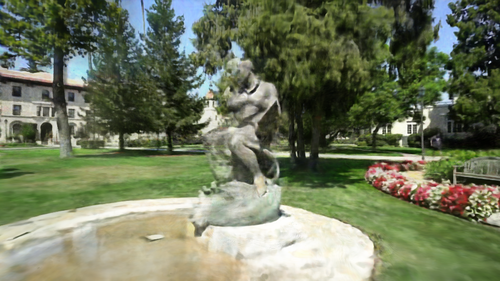}}
    \subfigure[]
    {\includegraphics[width=0.195\textwidth]{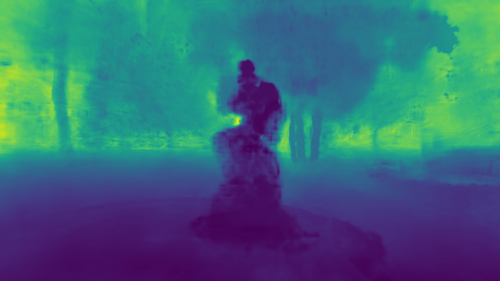}}
    \subfigure[]
    {\includegraphics[width=0.195\textwidth]{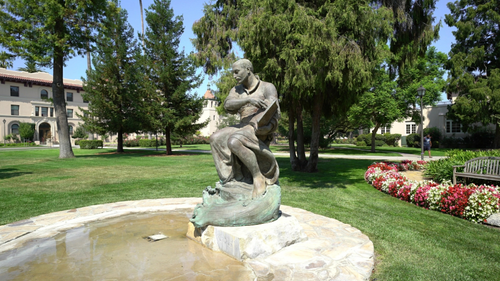}}
    \hfill\\\vspace{-20.5pt}    
    \subfigure[]
    {\includegraphics[width=0.195\textwidth]{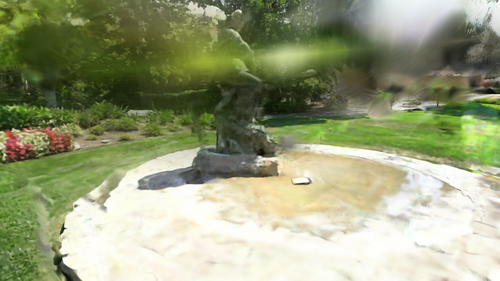}}
    \subfigure[]
    {\includegraphics[width=0.195\textwidth]{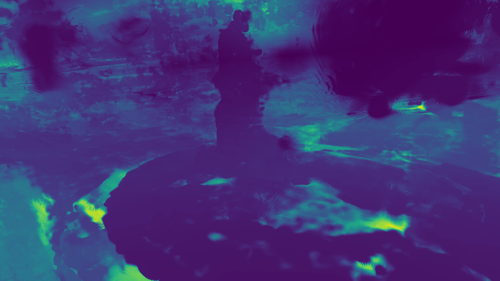}}
    \subfigure[]
    {\includegraphics[width=0.195\textwidth]{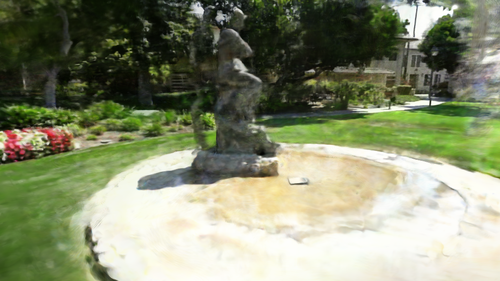}}
    \subfigure[]
    {\includegraphics[width=0.195\textwidth]{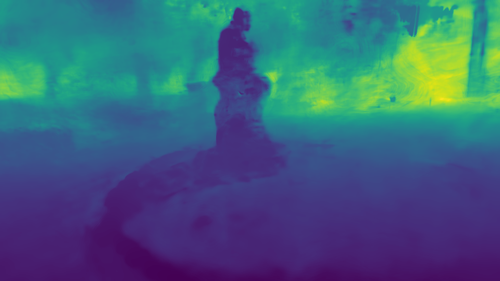}}
    \subfigure[]
    {\includegraphics[width=0.195\textwidth]{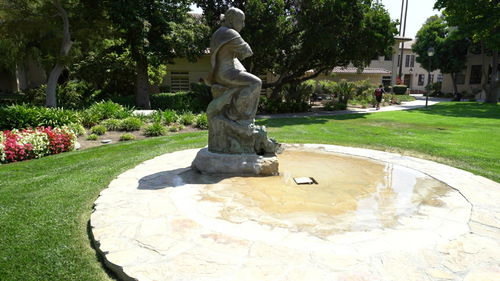}}
    \hfill\\\vspace{-20.5pt} 
    \subfigure[]
    {\includegraphics[width=0.195\textwidth]{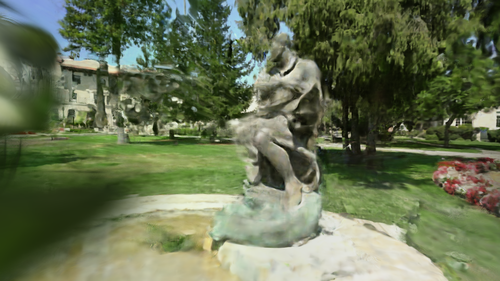}}
    \subfigure[]
    {\includegraphics[width=0.195\textwidth]{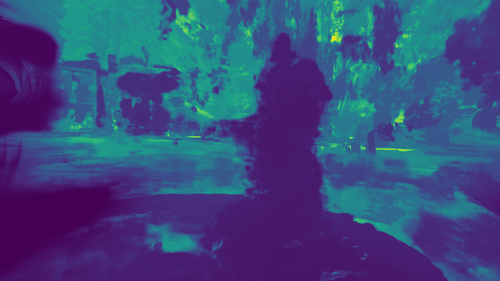}}
    \subfigure[]
    {\includegraphics[width=0.195\textwidth]{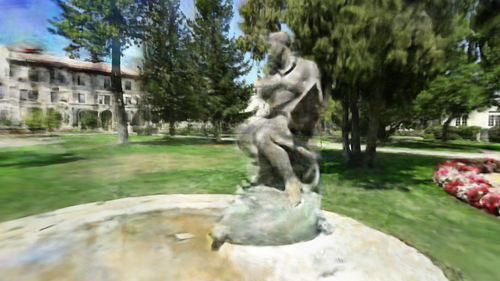}}
    \subfigure[]
    {\includegraphics[width=0.195\textwidth]{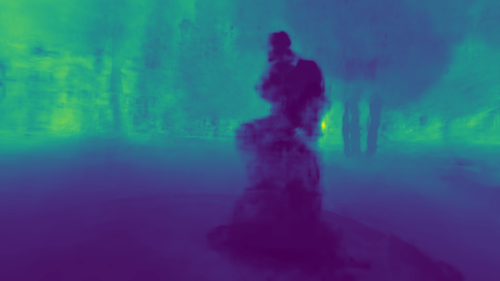}}
    \subfigure[]
    {\includegraphics[width=0.195\textwidth]{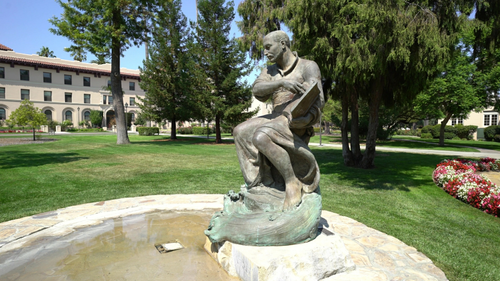}}
    \hfill\\\vspace{-20.5pt} 
    \subfigure[]
    {\includegraphics[width=0.195\textwidth]{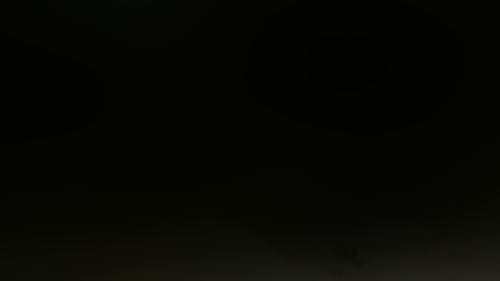}}
    \subfigure[]
    {\includegraphics[width=0.195\textwidth]{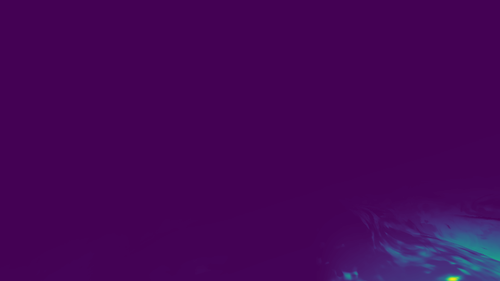}}
    \subfigure[]
    {\includegraphics[width=0.195\textwidth]{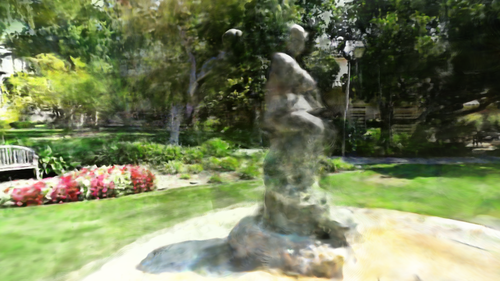}}
    \subfigure[]
    {\includegraphics[width=0.195\textwidth]{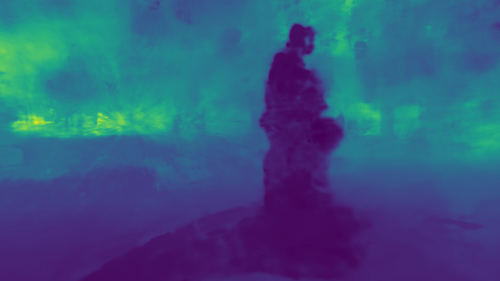}}
    \subfigure[]
    {\includegraphics[width=0.195\textwidth]{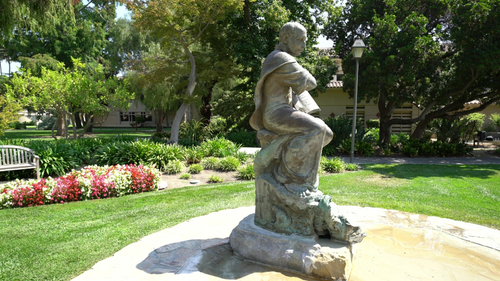}}
    \hfill\\\vspace{-20.5pt} 
    \subfigure[]
    {\includegraphics[width=0.195\textwidth]{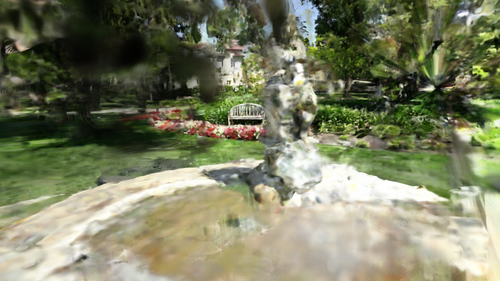}}
    \subfigure[]
    {\includegraphics[width=0.195\textwidth]{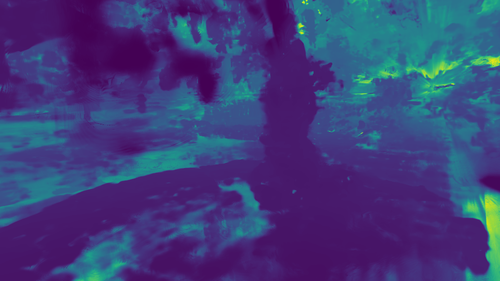}}
    \subfigure[]
    {\includegraphics[width=0.195\textwidth]{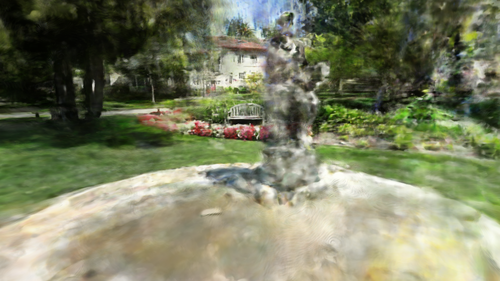}}
    \subfigure[]
    {\includegraphics[width=0.195\textwidth]{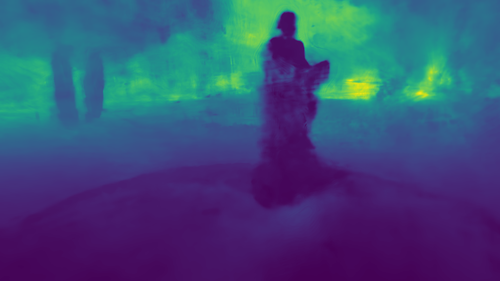}}
    \subfigure[]
    {\includegraphics[width=0.195\textwidth]{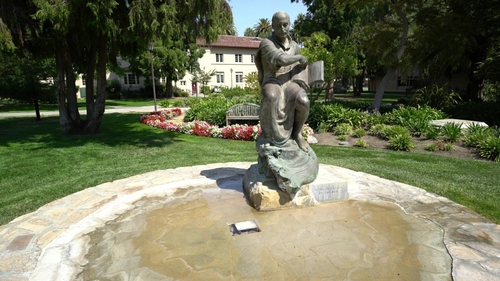}}
    \hfill\\\vspace{-20.5pt} 
    \subfigure[]
    {\includegraphics[width=0.195\textwidth]{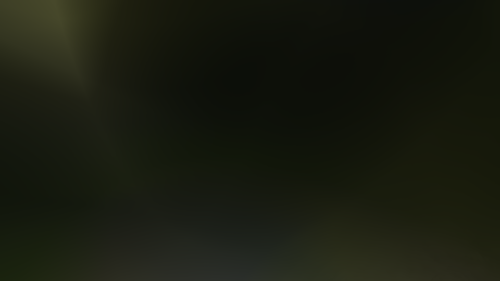}}
    \subfigure[]
    {\includegraphics[width=0.195\textwidth]{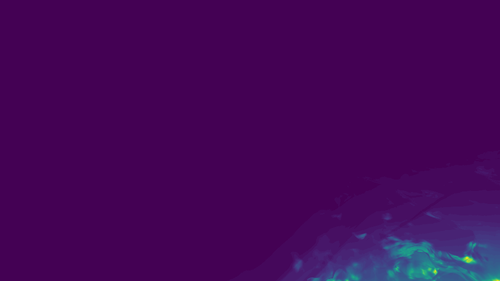}}
    \subfigure[]
    {\includegraphics[width=0.195\textwidth]{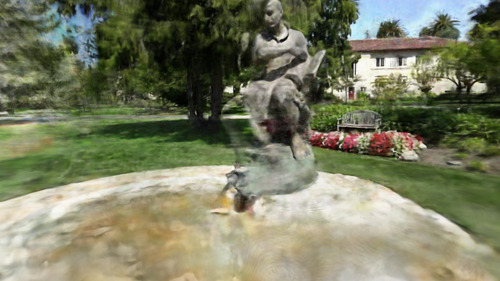}}
    \subfigure[]
    {\includegraphics[width=0.195\textwidth]{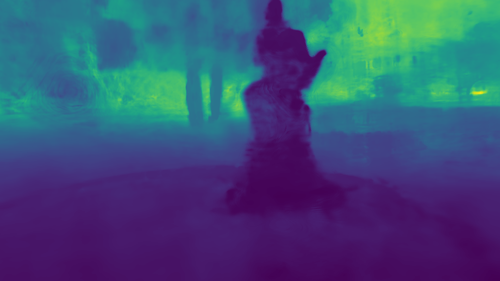}}
    \subfigure[]
    {\includegraphics[width=0.195\textwidth]{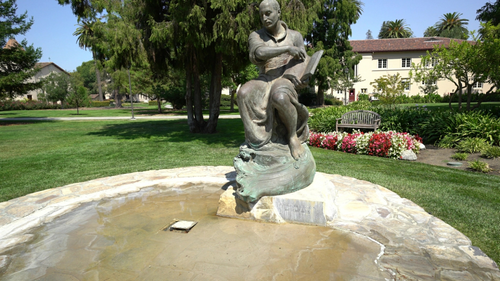}}
    \hfill\\\vspace{-20.5pt} 
    \subfigure[]
    {\includegraphics[width=0.195\textwidth]{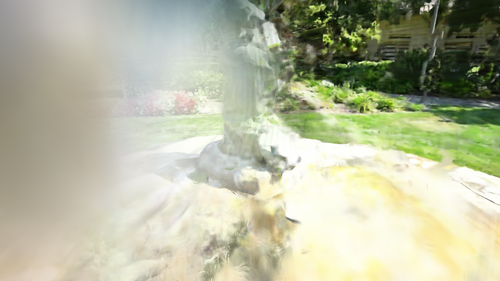}}
    \subfigure[]
    {\includegraphics[width=0.195\textwidth]{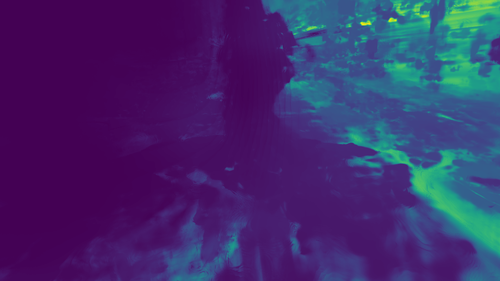}}
    \subfigure[]
    {\includegraphics[width=0.195\textwidth]{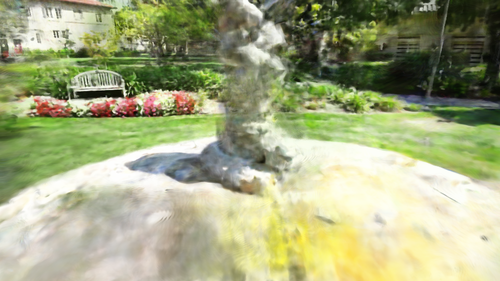}}
    \subfigure[]
    {\includegraphics[width=0.195\textwidth]{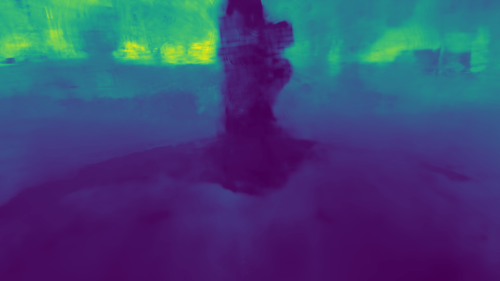}}
    \subfigure[]
    {\includegraphics[width=0.195\textwidth]{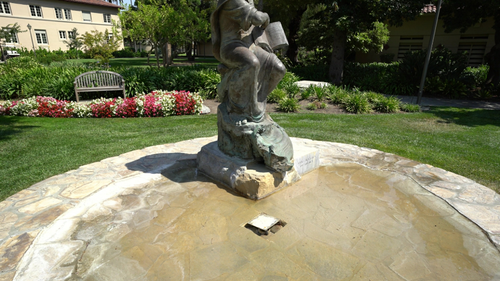}}
    \hfill\\\vspace{-20.5pt} 
    \subfigure[(a) Baseline~\cite{fridovich2023k}]
    {\includegraphics[width=0.195\textwidth]{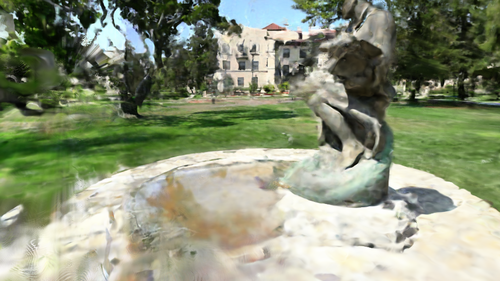}}
    \subfigure[(b) Baseline - Depth]
    {\includegraphics[width=0.195\textwidth]{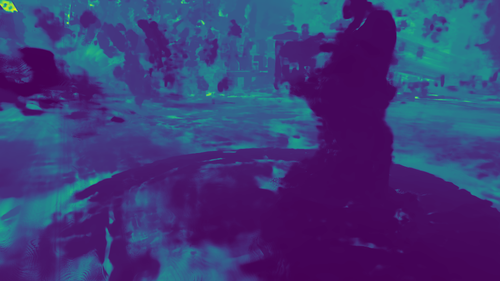}}
    \subfigure[(c) \ours]
    {\includegraphics[width=0.195\textwidth]{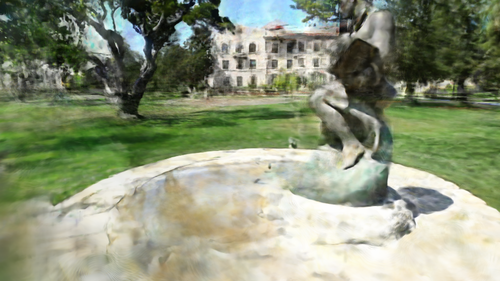}}
    \subfigure[(d) \ours - Depth]
    {\includegraphics[width=0.195\textwidth]{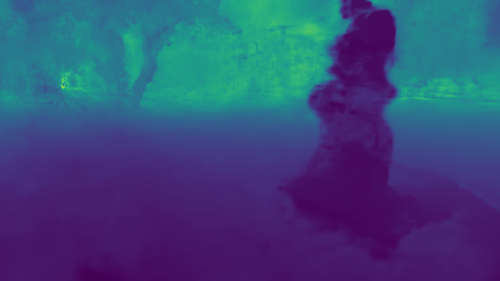}}
    \subfigure[(e) Ground truth]
    {\includegraphics[width=0.195\textwidth]{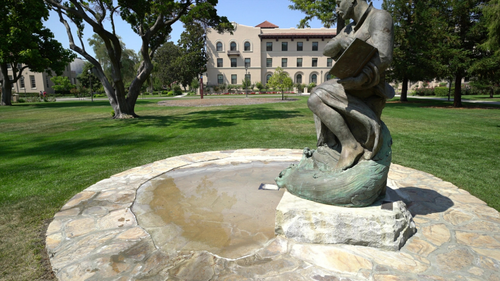}}\\
    \caption{\textbf{Qualitative results on ignatius scene of Tanks and Temples~\cite{knapitsch2017tanks}  with 10 input views}.}
    \vspace*{\fill}%
    \label{qual:ignatius}
\end{figure*}
\clearpage%

%% file: Suppl_Figures/_tex/TnT/TnT_family.tex
\begin{figure*}[]
\vspace*{\fill}%
\centering
    \renewcommand{\thesubfigure}{}
    \subfigure[(a) Baseline~\cite{fridovich2023k}]
    {\includegraphics[width=0.195\textwidth]{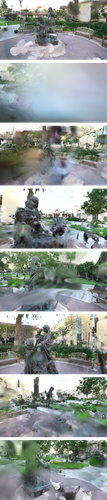}}
    \subfigure[(b) Baseline - Depth]
    {\includegraphics[width=0.195\textwidth]{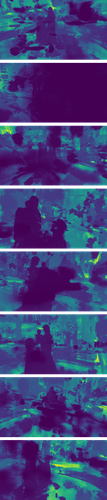}}
    \subfigure[(c) \ours]
    {\includegraphics[width=0.195\textwidth]{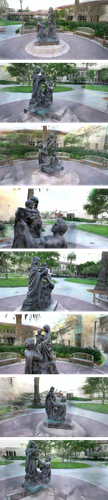}}
    \subfigure[(d) \ours - Depth]
    {\includegraphics[width=0.195\textwidth]{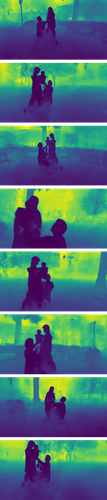}}
    \subfigure[(e) Ground truth]
    {\includegraphics[width=0.195\textwidth]{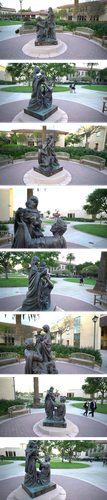}}\\
    \caption{\textbf{Qualitative results on family scene of Tanks and Temples~\cite{knapitsch2017tanks} with 10 input views}.}    
    \vspace*{\fill}%
    \label{qual:family}
\end{figure*}
\clearpage%

%% file: Suppl_Writing/5_others.tex
\newpage
\section{Limitations and Future Works}\label{supp:limitation}
While our method shows powerful performance quantitatively, its limitations can be noticed in its qualitative results above, where it struggles to reconstruct the fine-grained details present in ground truth images. Also, our usage of depth supervision from various viewpoints does not get rid of the artifacts completely: some artifacts that cloud the space between objects and the camera, are reduced yet still visible in rendering of unseen viewpoints. 

These limitations may be attributed to fundamental limitations in the few-shot NeRF setting~\cite{jain2021putting}, where fine-grained details are often occluded from one viewpoint to another due to an extreme lack of input images, preventing faithful geometric reconstruction of details. Also, since the seen viewpoints view a comparatively small portion of the entire scene, there inevitably occur artifacts in the unseen viewpoint as some depths cannot be perfectly determined from given input information. 

\section{Broader Impacts}\label{supp:impacts}
Our work achieves robust optimization and rendering of NeRF under sparse view scenarios, drastically reducing the number of viewpoints required for NeRF and bringing NeRF closer to real-life applications such as augmented reality, 3D reconstruction, and robotics. Our extension of few-shot NeRF to a real-world setting with the usage of monocular depth estimation networks also would enable NeRF optimization under various real-life lighting conditions and specular surfaces due to its increased robustness and generalization power. 